\documentclass[10pt,journal,compsoc]{IEEEtran}

\usepackage{
amsfonts,
cite,
pifont,
graphicx,
multirow,
booktabs,
amsmath,
amssymb,
hhline,
url,
hyperref}
\usepackage[ruled]{algorithm2e}
\usepackage[table,xcdraw]{xcolor}

\newcommand\shline{\specialrule{0.8pt}{0pt}{0pt}}

\usepackage{xcolor}
\definecolor{darkblue}{rgb}{0, 0, 0.6}
\definecolor{darkgreen}{rgb}{0, 0.7, 0}
\definecolor{darkred}{rgb}{0.8, 0, 0}

\let\oldcite\cite
\renewcommand{\cite}[1]{\textcolor{darkblue}{\oldcite{#1}}}

\newcommand{\x}{{\mathbf{x}}}

\newcommand{\ep}{{\boldsymbol{\epsilon}}}
\newcommand{\ephat}{{\hat{\boldsymbol{\epsilon}}}}
\newcommand{\xhat}{{\hat{\mathbf{x}}}}
\newcommand{\xbar}{{\bar{\mathbf{x}}}}
\newcommand{\Xhat}{{\hat{\mathbf{X}}}}
\newcommand{\Rbb}{{\mathbb{R}}}
\newcommand{\Ebb}{{\mathbb{E}}}
\newcommand{\y}{{\mathbf{y}}}

\newcommand{\A}{{\mathbf{A}}}

\newcommand{\F}{{\mathcal{F}}}
\newcommand{\Fbf}{{\mathbf{F}}}
\newcommand{\f}{{\mathbf{f}}}

\newcommand{\h}{{\mathbf{h}}}
\newcommand{\Loss}{{\mathcal{L}}}
\newcommand{\BigO}{{\mathcal{O}}}
\newcommand{\Th}{{\mathbf{\Theta}}}

\newcommand{\vs}{\textit{vs.}~}
\newcommand{\SM}{{\textcolor{blue}{\textit{\textbf{Supplementary Material~}}}}}

\newcommand{\best}[1]{\textbf{\textcolor{red}{#1}}}
\newcommand{\secondbest}[1]{\underline{\textcolor{blue}{#1}}}

\newcommand*\colorcheck{%
  \expandafter\newcommand\csname greencheck\endcsname{\textcolor{darkgreen}{\ding{52}}}%
}
\newcommand*\colorcross{%
  \expandafter\newcommand\csname redcross\endcsname{\textcolor{darkred}{\ding{56}}}%
}
\colorcheck
\colorcross

\begin{document}
\title{Invertible Diffusion Models for Compressed Sensing}

\author{Bin Chen, Zhenyu Zhang, Weiqi Li, Chen Zhao, Jiwen Yu, Shijie Zhao, Jie Chen, and Jian Zhang
\IEEEcompsocitemizethanks{
\IEEEcompsocthanksitem This work was partly supported by the National Natural Science Foundation of China (Grant No. 62372016) and Guangdong Provincial Key Laboratory of Ultra High Definition Immersive Media Technology (Grant No. 2024B1212010006).
\IEEEcompsocthanksitem Bin Chen, Zhenyu Zhang, Weiqi Li, Jiwen Yu, Jie Chen, and Jian Zhang are with the School of Electronic and Computer Engineering, Peking University, Shenzhen 518055, China.
\IEEEcompsocthanksitem Chen Zhao is with the King Abdullah University of Science and Technology, Thuwal 23955, Saudi Arabia.
\IEEEcompsocthanksitem Shijie Zhao is with ByteDance Inc, Shenzhen 518055, China.
\IEEEcompsocthanksitem Corresponding author: Chen Zhao.
\IEEEcompsocthanksitem E-mail: \{chenbin, liweiqi, yujiwen\}@stu.pku.edu.cn; \{zhenyuzhang, jiechen2019, zhangjian.sz\}@pku.edu.cn; chen.zhao@kaust.edu.sa; zhaoshijie.0526@bytedance.com}
\thanks{Manuscript received May 15, 2024; revised November 29, 2024, and January 22, 2025; accepted January 26, 2025.}}

\IEEEtitleabstractindextext{
\begin{abstract}
While deep neural networks (NN) significantly advance image compressed sensing (CS) by improving reconstruction quality, the necessity of training current CS NNs from scratch constrains their effectiveness and hampers rapid deployment. Although recent methods utilize pre-trained diffusion models for image reconstruction, they struggle with slow inference and restricted adaptability to CS. To tackle these challenges, this paper proposes \textbf{I}nvertible \textbf{D}iffusion \textbf{M}odels (\textbf{IDM}), a novel efficient, end-to-end diffusion-based CS method. IDM repurposes a large-scale diffusion sampling process as a reconstruction model, and fine-tunes it end-to-end to recover original images directly from CS measurements, moving beyond the traditional paradigm of one-step noise estimation learning. To enable such memory-intensive end-to-end fine-tuning, we propose a novel two-level invertible design to transform both (1) multi-step sampling process and (2) noise estimation U-Net in each step into invertible networks. As a result, most intermediate features are cleared during training to reduce up to 93.8\% GPU memory. In addition, we develop a set of lightweight modules to inject measurements into noise estimator to further facilitate reconstruction. Experiments demonstrate that IDM outperforms existing state-of-the-art CS networks by up to 2.64dB in PSNR. Compared to the recent diffusion-based approach DDNM, our IDM achieves up to 10.09dB PSNR gain and 14.54 times faster inference. Code is available at \url{https://github.com/Guaishou74851/IDM}.
\end{abstract}

}

\maketitle

\section{Introduction}
\IEEEPARstart{C}ompressed sensing (CS) \cite{donoho2006compressed,donoho2006most} is a novel signal acquisition paradigm that surpasses the Nyquist-Shannon theorem limit \cite{shannon1949communication}. It has inspired a range of imaging applications, including single-pixel imaging (SPI) \cite{duarte2008single}, magnetic resonance imaging (MRI) \cite{lustig2007sparse}, computational tomography (CT) \cite{chen2008prior,szczykutowicz2010dual}, and snapshot compressive imaging (SCI) \cite{yuan2021snapshot,fu2021coded,suo2023computational,cai2022mask,cai2022coarse,cai2022degradation}. In this paper, we focus on natural image CS reconstruction, aiming to recover the original image $\x \in \Rbb^{N}$ from its linear measurements $\y \in \Rbb^{M}$. These measurements are projected through a random sampling matrix $\A \in \Rbb^{M\times N}$ with a CS ratio $\gamma = M/N$, using $\y = \A \x$. A low $\gamma$ value, where $M \ll N$, is generally preferred for the appealing benefits such as lower energy consumption and shorter signal acquisition time. However, this also introduces the challenge of reconstructing $\x$ from limited information $\{\y, \A, \gamma\}$ due to the ill-posed nature of this inverse problem.

\begin{figure}[!t]
\centering
\includegraphics[width=\linewidth]{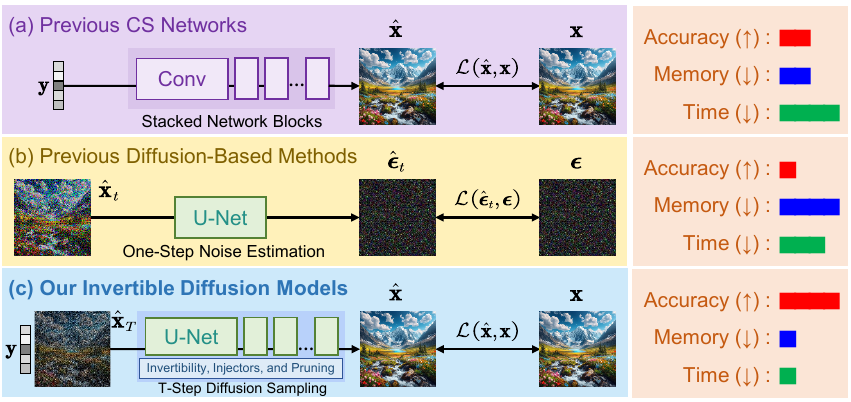}
\vspace{-20pt}
\caption{\textbf{Proposed IDM compared to previous methods.} \textbf{(a)} Conventional NN-based works \cite{chen2023deep} develop and train new CS NN architectures from scratch, limiting their ability to achieve higher performance within a short timeframe for rapid deployment. \textbf{(b)} Traditional diffusion-based image reconstruction methods \cite{saharia2023image} train a one-step noise estimation U-Net and use it as an off-the-shelf NN module for iterative sampling. This estimator lacks awareness of the entire recovery process from measurement to image, reducing its adaptability to CS. \textbf{(c)} Our invertible diffusion models (IDM) fine-tune a large-scale, pre-trained diffusion sampling process to directly predict original images from CS measurements end-to-end, significantly improving performance while reducing the required sampling steps (\textcolor{blue}{Contribution 1}). We further make the sampling process and noise estimation U-Net invertible, adding measurement injectors into our pruned noise estimation U-Net, resulting in a substantial performance boost while greatly reducing training GPU memory and runtime (\textcolor{blue}{Contributions 2 and 3}). Here, (a), (b), and (c) correspond to (12), (1), and (9) in Tab.~\ref{tab:abla}, respectively. Please refer to Sec.~\ref{sec:ablation_and_analysis} for more details.}
\label{fig:concept}
\end{figure}

In the field of natural image CS reconstruction, deep neural networks (NNs) demonstrate greater effectiveness than traditional optimization-based methods \cite{zhang2014group,dong2014compressive,zhao2016video} in terms of accuracy and efficiency. Early CS NNs \cite{mousavi2015deep,kulkarni2016reconnet} treat CS recovery as a de-aliasing problem, achieving fast, non-iterative reconstruction with a single forward pass through NN layers. Techniques like deep algorithm unrolling \cite{monga2021algorithm}, plug-and-play (PnP) \cite{zhang2022plug,kamilov2023plug}, and regularization by denoising (RED) \cite{romano2017little} effectively separate measurement consistency enforcement from image prior learning, balancing performance and interpretability. However, as shown in Fig.~\ref{fig:concept} (a), these methods typically require designing and training new NN architectures from scratch, a time-consuming process often resulting in suboptimal performance.

Recently, diffusion models \cite{nichol2021improved} have been widely leveraged for image reconstruction in many works \cite{zhu2023denoising,saharia2023image,wang2024exploiting,li2023self,lin2023diffbir,wu2024seesr} by utilizing pre-trained generative denoising priors. These approaches iteratively sample an image estimate $\xhat$ through steps from the posterior distribution $p(\x|\y)$, achieving impressive measurement-conditioned synthesis \cite{zhu2023denoising}. Similar to PnP methods, current diffusion-based approaches focus on learning a noise estimation network \cite{saharia2023image} or using pre-trained ones to address the image prior subproblems.

However, these methods often require extensive tuning of hyperparameters, such as step size, regularization coefficient, and noise level. Furthermore, as Fig.~\ref{fig:concept} (b) shows, their NN backbones are mainly trained for one-step noise estimation, not directly optimized for the full reconstruction process from CS measurements $\y$ to the target original image $\x$. Additionally, previous diffusion-based inverse problem solvers \cite{zhang2019deep,kawar2022denoising,song2023optimization,chung2023diffusion} can require numerous iterations, ranging from 10 to 1000, to achieve satisfactory results. Approaches using latent diffusion models, like stable diffusion (SD) \cite{rombach2022high,stabilityai}, typically involve multi-stage training \cite{xia2023diffir} or frequent transitions between image and latent spaces via deep VAE encoders and decoders during sampling, reducing the efficiency of CS imaging systems.

To address these challenges, in this paper, we propose \textbf{I}nvertible \textbf{D}iffusion \textbf{M}odels (\textbf{IDM}) for image CS reconstruction, as shown in Fig.~\ref{fig:concept} (c). Moving beyond the limitations of one-step noise estimation learning in existing methods, our IDM establishes a novel end-to-end framework, training directly to align a large-scale, pre-trained diffusion sampling process with the ideal reconstruction mapping $\y \mapsto \x$. This alignment ensures that all diffusion parameters and weights of noise estimation network are specifically optimized for image CS reconstruction, achieving significant performance gains and eliminating the need for a great deal of sampling steps. However, training such large diffusion models end-to-end demands high GPU memory, making it nearly infeasible on standard GPUs. To address this, we leverage the memory efficiency of invertible NNs \cite{zhao2023re2tal} and propose a novel two-level invertible design. This design introduces auxiliary connections into (1) the overall diffusion sampling framework across multiple steps and (2) noise estimation U-Net within each step, transforming both into invertible networks. This allows IDM to clear most intermediate features during the forward pass and recompute them in back-propagation, significantly reducing GPU memory consumption.

Compared with previous end-to-end CS NNs that require training from scratch, IDM leverages pre-trained diffusion models, adapting them to CS with minimal fine-tuning. Additionally, we introduce lightweight NN modules, called injectors, that directly integrate the information of measurement $\y$ into the deep features of the noise estimation U-Net at various spatial scales to enhance reconstruction. Our method harnesses the power of pre-trained diffusion models to improve CS performance, establishing a new path for the customized application of large-scale diffusion models in image reconstruction. In summary, our contributions are:

(1) We propose IDM, an efficient, end-to-end diffusion-based CS method. Unlike previous one-step diffusion noise estimation, IDM fine-tunes a large-scale sampling process for direct recovery from CS measurements end-to-end, improving performance by up to 4.34dB in PSNR with a 98\% reduction in step number and over 25 times faster inference.

(2) We introduce a novel two-level invertible design for both the diffusion sampling framework and noise estimation U-Net, reducing fine-tuning memory by up to 93.8\%.

(3) We develop a set of lightweight injector modules that integrate CS measurements and sampling matrix into the noise estimation U-Net to enhance CS reconstruction. These injectors achieve over 2dB PSNR improvements with only 0.02M extra parameters, compared to over 100M of U-Net.

(4) We evaluate our IDM, as shown in Fig.~\ref{fig:concept} (c), across four typical tasks: natural image CS, inpainting, accelerated MRI, and sparse-view CT. IDM achieves a new state-of-the-art (SOTA), surpassing existing CS NNs by up to 2.64dB in PSNR, with up to 10.09dB PSNR gain and 14.54 times faster inference compared to diffusion-based method DDNM \cite{wang2023zero}.

\begin{figure*}[!t]
\centering
\includegraphics[width=\linewidth]{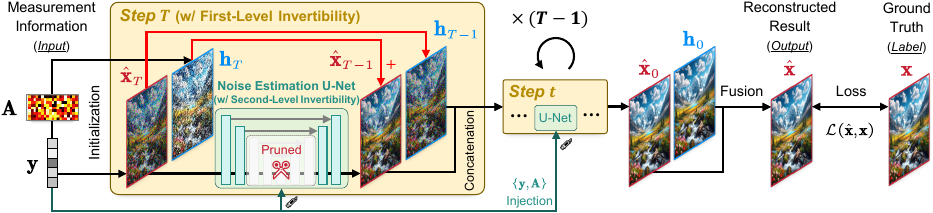}
\vspace{-20pt}
\caption{\textbf{Illustration of our proposed IDM framework.} It receives an initial image estimate $\xhat_T$ and learns $T$ diffusion sampling steps for end-to-end recovery. Auxiliary connections (shown as \textcolor{red}{red arrows}) enable invertibility and facilitate the reuse of powerful large-scale pre-trained SD models.}
\label{fig:sampling}
\end{figure*}

\section{Related Work}
\subsection{Deep End-to-End Learned Image CS Networks}
\label{sec:background_e2e}
Foundational NN-based CS research \cite{mousavi2015deep,iliadis2018deep} pioneered the use of fully connected layers to decode measurements into images. The introduction of convolutions \cite{kulkarni2016reconnet} and self-attention layers \cite{vaswani2017attention} led to more effective architectures like residual blocks \cite{he2016deep} and Transformers \cite{ye2023csformer}, which capture both local and long-range dependencies in images. This popularized expressive NN designs such as hierarchical and non-local architectures \cite{shi2019scalable,cui2021image}. Recent works \cite{chen2020deep,chen2024practical} leverage measurement information $\{\y,\A,\gamma\}$ to build physics-informed CS NNs. Among these methods, deep unrolling \cite{zhang2023physics}, which reinterprets truncated optimization inferences as iterative NNs, has established a new paradigm in this field. In comparison, our work introduces powerful diffusion priors, avoids resource-intensive training from scratch, and enhances the efficiency of CS recovery learning.

\subsection{Diffusion-Based Image Reconstruction}
Denoising diffusion models, particularly the denoising diffusion probabilistic models (DDPM) \cite{ho2020denoising}, have emerged as effective generative priors for inverse imaging problems. DDPM employs a $T$-step noising process $\x_t =\sqrt{\alpha_t}\x_{t-1}+\sqrt{1-\alpha_t}\ep$, which can often be equivalently expressed as $\x_t =\sqrt{\bar{\alpha}_t}\x_{0}+\sqrt{1-\bar{\alpha}_t}\ep$ by utilizing the variance-preserving stochastic differential equation (VP-SDE) formulation \cite{song2021score}. Here, $\x_t$ represents the sampled, scaled, and noisy image at step $t$, $\ep\sim \mathcal{N}(\mathbf{0}_N,\mathbf{I}_N)$ is random Gaussian noise, $\bar{\alpha}_t=\prod_{i=0}^{t}{\alpha_i}$ determines the scaling factors with $\alpha_t=1-\beta_t$ and noise schedule $\beta_t\in[0,1]$~$(\beta_0 =0)$, while $\x_{0}$ is sampled from the clean image distribution $p(\x)$. Using a pre-trained NN $\ep_\Th$ that regresses $\min_{\Th}\lVert \ep - \ep_\Th(\sqrt{\bar{\alpha}_t}\x_{0}+\sqrt{1-\bar{\alpha}_t}\ep, t) \rVert_2^2$ with learnable parameter set $\Th$, and given a starting point $\xhat_T$, the denoising diffusion implicit model (DDIM) \cite{song2021denoising} enables accelerated generation of images via a deterministic sampling strategy $\xhat_{t-1}=\sqrt{\bar{\alpha}_{t-1}}\xhat_{0|t}+\sqrt{1-\bar{\alpha}_{t-1}}\ephat_t$, where $\xhat_{0|t}={\left(\xhat_{t}-\sqrt{1-\bar{\alpha}_{t}}\ephat_t\right)}/{\sqrt{\bar{\alpha}_{t}}}$ denotes the current denoised image, and $\ephat_t=\ep_\Th(\xhat_{t},t)$ is the estimation of added noise.

Recent research \cite{saharia2022palette,saharia2023image} regards image reconstruction as the task of measurement-conditioned image generation. Zero-shot diffusion solvers for inverse problems \cite{graikos2022diffusion} have leveraged the frameworks of singular value decomposition (SVD) \cite{kawar2022denoising}, manifold constraints \cite{chung2022improving}, posterior sampling \cite{chung2023diffusion}, range-nullspace decomposition (RND) \cite{wang2023zero}, and pseudoinverse \cite{song2022pseudoinverse} for recovery of different degradation operators. To our knowledge, no diffusion models are specifically designed for natural image CS. While many methods can be applied to CS, their noise estimation U-Nets are pre-trained for single-step noise estimation (see Fig.~\ref{fig:concept} (b)). Our work overcomes this limitation by aligning diffusion sampling with CS targets via end-to-end learning, improving performance and reducing the required number of steps.

\subsection{Invertible Neural Networks for Vision Tasks}
Invertible NNs, mathematically invertible functions, are a type of NN that can reconstruct the input from the output. Initially, they were utilized for image generation \cite{kingma2018glow,ho2019flow++}. Later research found that, unlike non-invertible NNs which cache all intermediate activations for gradient computation, invertible NNs allow most features to be freed up and recomputed as needed, reducing memory requirements \cite{gomez2017reversible}. Various invertible architectures have since been proposed, including convolutional, recurrent, and graph NNs \cite{song2019mintnet,mackay2018reversible,li2021training}, as well as Transformers \cite{mangalam2022reversible,zhao2023re2tal}. Their applications extend to image reconstruction \cite{liu2021invertible,huang2022winnet} and image editing \cite{wallace2023edict}, with on-the-fly optimization in the latent space of autoencoders \cite{wallace2023end}. The memory-efficient nature of invertible NNs is especially advantageous for training large diffusion models end-to-end. In this work, we propose a novel two-level invertible design for end-to-end fine-tuning of a large-scale pre-trained diffusion sampling process at minimal memory cost, making it feasible with limited GPU resources.

\section{Method}
\subsection{Preliminary}
\textbf{Denoising Diffusion Null Space Model (DDNM), and Its Limitations.} DDNM \cite{wang2023zero} is a state-of-the-art training-free image recovery approach that employs a pre-trained noise estimator $\ep_{\Th}$ as generative prior and solves the CS problem by iterating through the following three DDIM substeps:
\begin{align}
& \xhat_{0|t}=\left[\xhat_{t}-\sqrt{1-\bar{\alpha}_{t}}\ep_\Th(\xhat_{t},t)\right]/{\sqrt{\bar{\alpha}_{t}}}, \label{eq:ddnm_denoising} \\
& \xbar_{0|t} = \A^{\dagger}\y + (\mathbf{I}_N-\A^{\dagger}\A) \xhat_{0|t}, \label{eq:ddnm_rnd} \\
& \xhat_{t-1}=\sqrt{\bar{\alpha}_{t-1}}\xbar_{0|t}+\sqrt{1-\bar{\alpha}_{t-1}}\ep_\Th(\xhat_{t},t). \label{eq:ddnm_sampling}
\end{align}
In Eq.~(\ref{eq:ddnm_denoising}), the denoised image is predicted, while Eq.~(\ref{eq:ddnm_rnd}) applies the RND theory \cite{schwab2019deep,chen2020deep} for ensuring measurement consistency $\A\xbar_{0|t}=\A\A^\dagger \y +\A(\mathbf{I}_N-\A^{\dagger}\A) \xhat_{0|t}\equiv\y$. Finally, Eq.~(\ref{eq:ddnm_sampling}) performs a deterministic DDIM sampling step, where $\A^{\dagger}\in\Rbb^{N\times M}$ is the pseudo-inverse of CS sampling matrix $\A$. However, the performance of DDNM is limited by a task shift from pre-training $\ep_{\Th}$ on noise estimation in DDPM to its application in CS. The deep features within $\ep_{\Th}$ are not necessarily optimized for mapping $\y$ to $\x$. Similar challenges exist in other diffusion-based methods, including zero-shot solvers \cite{chung2023diffusion} and conditional models \cite{saharia2022palette,saharia2023image}.

\subsection{Learn Diffusion Sampling End-to-End for CS}
\textbf{Diffusion-Based End-to-End CS Learning Framework.} We hypothesize that directly fitting the steps of DDNM to the recovery mapping $\y \mapsto \x$ end-to-end can mitigate task shift issues and improve final performance. To achieve this, we construct a DDNM-based CS reconstruction framework, repurposing the sampling process of DDNM as a $T$-layer NN $\F=\F_1\circ\F_2\circ \cdots \F_T$, where each layer $\xhat_{t-1}=\F_t(\xhat_t; \y, \A)$ denotes a single sampling step encompassing Eqs.~(\ref{eq:ddnm_denoising})-(\ref{eq:ddnm_sampling}), with $\circ$ representing the composition of sequential layers. Our IDM framework fine-tunes both the diffusion parameters $\{\alpha_t\}$ and the pre-trained weights in $\ep_{\Th}$. It optimizes deep features to minimize the difference between estimation $\xhat=\xhat_0=\F(\xhat_T; \y, \A)$ and the ground truth $\x$ using an $L_1$ loss $\Loss(\xhat,\x)=\frac{1}{N}\lVert \xhat - \x \rVert_1$ and standard back-propagation \cite{rumelhart1986learning}. This approach enables comprehensive adaptability to the full CS recovery process, not limited to noise estimation, enhancing performance, and introducing new gains orthogonal to developments in solvers and text prompts \cite{chung2023prompt}.

\begin{figure}[!t]
\centering
\includegraphics[width=\linewidth]{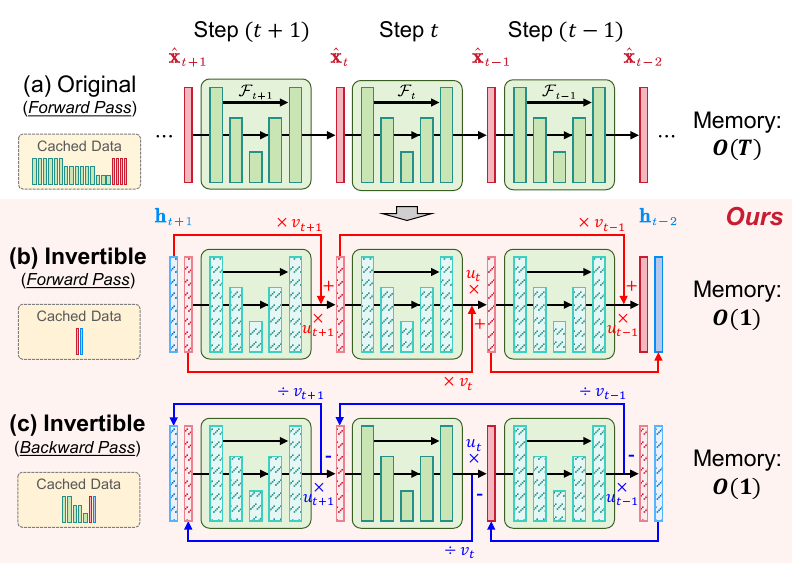}
\vspace{-20pt}
\caption{\textbf{Illustration of our wiring technique, exemplified with three diffusion sampling steps and a three-scale noise estimation U-Net.} Here, light-colored rectangles with diagonal dashed lines represent images/features that are cleared, while dark-colored, empty rectangles indicate images/features that must be preserved. \textbf{(a)} The original non-invertible forward pass caches all inputs, features, and outputs, causing memory usage to increase linearly with the step number. \textbf{(b)} We add connections to construct invertible layers, reducing memory usage to a constant level. \textbf{(c)} During back-propagation, the necessary intermediate inputs/features are sequentially recomputed and cleared to obtain gradients from the last to the first step. Our wired sampling framework is equivalent to the original setup in (a) when $u_t = 1$ and $v_t = 0$.}
\label{fig:wiring}
\end{figure}

\textbf{Initialization Strategy of $\xhat_T$ for CS Reconstruction.} Unlike typical diffusion models that sample from random noise or employ NNs to estimate an initial result \cite{whang2022deblurring}, we propose a simple yet effective initialization of image by calculating an expectation $\xhat_T=\Ebb_{\ep}\left[\sqrt{\bar{\alpha}_T}\xhat_0^{\prime}+\sqrt{1-\bar{\alpha}_T}\ep\right]=\sqrt{\bar{\alpha}_T}\A^{\dagger}\y$ using back-projection $\xhat_0^{\prime}=\left(\A^{\dagger}\y\right)\in \arg\min_{\x}\lVert \A\x -\y \rVert_2^2$. This strategy enhances reconstruction quality at low computational cost, outperforming noise-initialized counterparts.

\begin{figure*}[!t]
\centering
\includegraphics[width=1.0\linewidth]{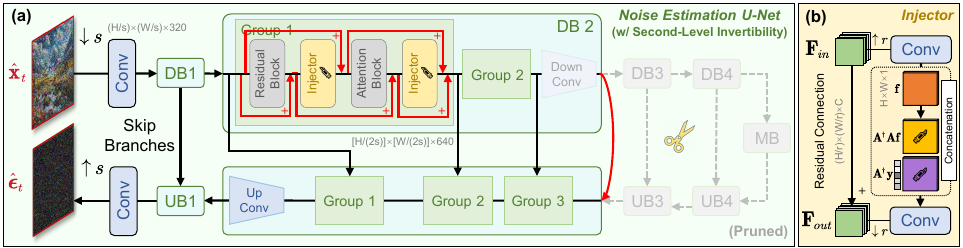}
\vspace{-20pt}
\caption{\textbf{Illustration of our modified noise estimation U-Net $\ep_\Th$ for image CS reconstruction tasks, based on the SD v1.5 models \cite{rombach2022high,stabilityai}.} {\textbf{(a)}} Injectors are added behind each residual and attention block and grouped within downsampling, upsampling, and middle blocks (marked as DB, UB, and MB) for invertibility. Our method is orthogonal to and compatible with network pruning \cite{kim2023bk} for enhanced efficiency. {\textbf{(b)}} Each injector learns to fuse measurement physics $\{\y,\A\}$ into deep features using convolutions and residual connections.}
\label{fig:unet}
\end{figure*}

\subsection{Two-Level Invertible Design for Memory Efficiency}
\label{sec:wiring}
Our end-to-end training scheme improves performance but can challenge the memory capacity of GPUs, especially with increased step number $T$ and a large noise estimator $\ep_\Th$ due to the substantial footprint of intermediate features cached for back-propagation. For instance, it is infeasible to directly train a diffusion sampling process with $T=3$ and the SD v1.5 U-Net backbone end-to-end on 4 NVIDIA A100 GPUs with 80GB memory. To reduce memory consumption, we leverage the memory efficiency of invertible NNs and propose a novel two-level invertible design, making both (1) the multi-step sampling framework and (2) the noise estimation NN invertible through a new ``wiring''\footnote{Throughout this paper, ``wiring'' refers to the process of integrating auxiliary connections into a pre-given NN architecture.} technique.

\textbf{First-Level Invertibility for Multi-Step Diffusion Sampling Framework.} Inspired by \cite{zhao2023re2tal,zhao2024dr2net}, we propose enabling invertibility by introducing new connections into the sampling steps, as shown in Fig.~\ref{fig:wiring}. Each connection transmits the input $\xhat_t$ from one step $\F_t$ to the output $\xhat_{t-2}$ of the next step $\F_{t-1}$. We use two learnable weighting scalars $u_t$ and $v_t$, such that $u_t + v_t = 1$, to scale the output of each step and the transmitted data, respectively, and fuse them to obtain $\xhat_{t-1}$. An auxiliary variable $\h_t$ is introduced to transform each layer into an invertible one, allowing recovery of its input $\{\xhat_t, \h_t\}$ from a given output $\{\xhat_{t-1}, \h_{t-1}\}$ through a second pathway. Mathematically, the forward and inverse computations of each wired layer are formulated as:
\begin{align}
&\text{Forward:}&&\left\{
\begin{aligned}
&\xhat_{t-1} = u_t \F_t(\xhat_t; \y, \A) + v_t \h_t, \\
&\h_{t-1} = \xhat_t,
\end{aligned}
\right.\label{eq:forward}
\\
&\text{Inverse:}&&\left\{
\begin{aligned}
&\xhat_t = \h_{t-1}, \\
&\h_t = {\left[\xhat_{t-1} - u_t \F_t(\xhat_t; \y, \A)\right]} / {v_t}.
\end{aligned}
\right.\label{eq:inverse}
\end{align}
To maintain the dimensional consistency in the framework’s input and output after wiring, we introduce two learnable scalars, $w_T$ and $w_0$, at the beginning and end, computing the input and output as $\h_T = w_T \xhat_T$ and $\xhat = \xhat_0 + w_0 \h_0$, respectively. As illustrated in Figs.~\ref{fig:wiring} (b) and (c), in the forward pass, we cache only the final output of the wired layers, freeing up all intermediate features. After computing the loss function $\Loss$, we execute standard back-propagation layer-by-layer, from $\{\xhat_0, \h_0\}$ back to $\{\xhat_T, \h_T\}$. For layer $t$, we jointly recalculate its inputs, necessary features, and parameter gradients based on the outputs $\xhat_{t-1}$ and $\h_{t-1}$, as well as their respective partial derivatives $\left(\partial \Loss / \partial \xhat_{t-1}\right)$ and $\left(\partial \Loss / \partial \h_{t-1}\right)$ with respect to the loss function $\Loss$. We implement efficient memory management by clearing all cached intermediate features and outputs as we move through each layer from $t$ back to $(t+1)$\footnote{A code snippet is provided in Sec.~B.3 in \SM.}.

Consequently, the memory of cached images and features in the framework is reduced from linear complexity $\BigO(T)$ to constant complexity $\BigO(1)$ when the sampling substeps and the architecture of noise estimation NN $\ep_\Th$ are pre-determined. The developed wiring technique offers two advantages. Firstly, it can be applied to arbitrary diffusion models and solvers without the need to design new NNs. Secondly, we reuse the pre-trained weights of $\ep_\Th$ for fine-tuning, with appropriate settings of $u_t$ and $v_t$. Our approach thereby leads to considerable savings in memory, time, and computation for improving reconstruction performance.

\textbf{Second-Level Invertibility for Noise Estimator in Each Step.} A U-Net architecture \cite{ronneberger2015u} is generally utilized as noise estimator $\ep_\Th$ in diffusion solvers. It includes multiple spatial scales with main and skip branches in both its down- and up-sampling blocks. These blocks contain many complex, equidimensional transformations \cite{rombach2022high} like residual connections and self-attention layers that are memory-intensive. To further improve memory efficiency, we extend our wiring technique to U-Net blocks in each step. As depicted in Fig.~\ref{fig:unet} (a), we group and wire consecutive transformation blocks within each down-/up-sampling block to make them invertible. During training, we clear all input and intermediate activations from memory for these grouped blocks, while preserving only the features for the first/last convolutions and skip branches for back-propagation. This second-level invertible design further reduces memory usage, making IDM training feasible on standard consumer-grade GPUs.

\subsection{Inject Measurement Physics into Noise Estimator}
Effective utilization of physics information $\{\y,\A\}$ is critical for CS reconstruction. Using it solely to initialize $\xhat_T$ and in the RND-based step in Eq.~(\ref{eq:ddnm_rnd}) limits reconstruction quality in complex scenarios (as shown in Fig.~\ref{fig:comp_abla_CS} (5) \vs (9)), as it is challenging for the NN layers in $\ep_\Th$ to learn the physics. Therefore, it is necessary to inject the measurement physics directly into the layers. Simply appending the measurement at the input layer of $\ep_\Th$ in each step, as in \cite{saharia2023image}, can yield unsatisfactory results, as the information of sampling matrix $\A$ and measurement $\y$ is difficult to retain in deep features.

To address this, we propose a series of injectors inspired by the success of physics-informed NNs (see Sec.~\ref{sec:background_e2e}). As illustrated in Fig.~\ref{fig:unet}, each injector is positioned behind every grouped block within $\ep_\Th$. For an intermediate deep feature $\Fbf_{in}\in\Rbb^{(H/r)\times (W/r)\times C}$ in U-Net, our injector fuses measurement physics information $\{\y,\A\}$ and $\Fbf_{in}$ as follows:
\begin{align}
&\f=\text{Conv}_1\left((\Fbf_{in})_{\uparrow r}\right)~\in\Rbb^{H\times W\times 1}~(N=H\times W), \\
&\Fbf_{out}=\Fbf_{in}+\left(\text{Conv}_2([\f,\A^\dagger \A\f, \A^\dagger \y])\right)_{\downarrow r}.\label{eq:injector}
\end{align}
Here, $\text{Conv}_1$ and $\text{Conv}_2$ denote two $3\times 3$ convolutions that transition data between feature and image domains, $(\cdot)_{\uparrow/\downarrow r}$ is a Pixel-Shuffle/-Unshuffle layer with scaling ratio $r$, and $[\f,\A^\dagger \A\f, \A^\dagger \y]\in\Rbb^{H\times W\times 3}$ is a channel-wise concatenation of the intermediate image space data $\f$, its interaction with the CS sampling matrix $\A^\dagger \A\f$, and the back-projected measurement $\A^\dagger \y$. These lightweight injectors are jointly wired and learned within each block group in $\ep_\Th$, establishing direct pathways that fuse the measurement information and features, effectively enhancing final recovery quality.

\begin{table*}[!t]
\caption{\textbf{Comparison of average PSNR (dB, $\uparrow$) (±std) across various deep end-to-end learned CS methods} on the luminance component of test images. Throughout this paper, the best and second-best results of each case are highlighted in \best{bold red} and \secondbest{underlined blue}, respectively.}
\vspace{-10pt}
\label{tab:compare_sota_e2e}
\centering
\resizebox{1.0\textwidth}{!}{
\begin{tabular}{lc|ccc|ccc}
\shline
\rowcolor[HTML]{EFEFEF} 
\multicolumn{1}{l|}{\cellcolor[HTML]{EFEFEF}} &
  Test Set &
  \multicolumn{3}{c|}{\cellcolor[HTML]{EFEFEF}Set11} &
  \multicolumn{3}{c}{\cellcolor[HTML]{EFEFEF}CBSD68} \\ \hhline{>{\arrayrulecolor[HTML]{EFEFEF}}->{\arrayrulecolor{black}}|-------} 
\rowcolor[HTML]{EFEFEF} 
\multicolumn{1}{l|}{\multirow{-2}{*}{\cellcolor[HTML]{EFEFEF}Method}} & CS Ratio $\gamma$ & 10\% & 30\% & 50\% & 10\% & 30\% & 50\% \\ \hline \hline
\multicolumn{2}{l|}{ReconNet (CVPR 2016) \cite{kulkarni2016reconnet}}        & 24.08 (±2.29) & 29.46 (±2.32) & 32.76 (±2.19) & 23.92 (±3.32) & 27.97 (±3.74) & 30.79 (±3.77) \\
\multicolumn{2}{l|}{ISTA-Net$^+$ (CVPR 2018) \cite{zhang2018ista}}    & 26.49 (±3.15) & 33.70 (±3.14) & 38.07 (±3.03) & 25.14 (±3.97) & 30.24 (±4.66) & 33.94 (±4.91) \\
\multicolumn{2}{l|}{CSNet$^+$ (TIP 2019) \cite{shi2019image}}       & 28.34 (±3.19) & 34.30 (±3.03) & 38.52 (±3.04) & 27.04 (±3.84) & 31.60 (±4.18) & 35.27 (±4.24) \\
\multicolumn{2}{l|}{SCSNet (CVPR 2019) \cite{shi2019scalable}} & 28.52 (±3.15) & 34.64 (±3.02) & 39.01 (±3.10) & 27.14 (±3.88) & 31.72 (±4.26) & 35.62 (±4.32) \\
\multicolumn{2}{l|}{OPINE-Net$^+$ (JSTSP 2020) \cite{zhang2020optimization}} & 29.81 (±3.24) & 35.99 (±2.87) & 40.19 (±2.90) & 27.66 (±4.25) & 32.38 (±4.65) & 36.21 (±4.74) \\
\multicolumn{2}{l|}{DPA-Net (TIP 2020) \cite{sun2020dual}}         & 27.66 (±3.37) & 33.60 (±2.98) & - & 25.33 (±4.12) & 29.58 (±4.31) & - \\
\multicolumn{2}{l|}{MAC-Net (ECCV 2020) \cite{chen2020learning}}         & 27.92 (±3.05) & 33.87 (±3.01) & 37.76 (±3.07) & 25.70 (±4.16) & 30.10 (±4.74) & 33.37 (±4.05) \\
\multicolumn{2}{l|}{RK-CCSNet (ECCV 2020) \cite{zheng2020sequential}} & - & - & 38.01 (±3.22) & - & - & 34.69 (±4.37) \\
\multicolumn{2}{l|}{ISTA-Net$^{++}$ (ICME 2021) \cite{you2021ista}} & 28.34 (±3.34) & 34.86 (±2.93) & 38.73 (±2.92) & 26.25 (±4.25) & 31.10 (±4.66) & 34.85 (±4.68) \\
\multicolumn{2}{l|}{COAST (TIP 2021) \cite{you2021coast}} & 30.02 (±3.43) & 36.33 (±2.80) & 40.33 (±2.88) & 27.76 (±4.35) & 32.56 (±4.69) & 36.34 (±4.74) \\
\multicolumn{2}{l|}{AMP-Net (TIP 2021) \cite{zhang2021amp}} & 29.40 (±3.05) & 36.03 (±2.65) & 40.34 (±2.85) & 27.71 (±4.06) & 32.72 (±4.46) & 36.72 (±4.38) \\
\multicolumn{2}{l|}{MADUN (ACM MM 2021) \cite{song2021memory}}           & 29.89 (±3.24) & 36.90 (±2.76) & 40.75 (±3.05) & 28.04 (±4.27) & 33.07 (±4.74) & 36.99 (±4.72) \\
\multicolumn{2}{l|}{DGUNet$^+$ (CVPR 2022) \cite{mou2022deep}}          & 30.92 (±3.23) & - & 41.24 (±3.42) & 27.89 (±4.26) & - & 36.86 (±4.69) \\
\multicolumn{2}{l|}{MR-CCSNet$^+$ (CVPR 2022) \cite{fan2022global}} & - & - & 39.27 (±3.31) & - & - & 35.45 (±4.49) \\
\multicolumn{2}{l|}{FSOINet (ICASSP 2022) \cite{chen2022fsoinet}} & 30.44 (±3.27) & 37.00 (±2.72) & 41.08 (±3.16) & 28.12 (±4.35) & 33.15 (±4.74) & 37.18 (±4.76) \\
\multicolumn{2}{l|}{CASNet (TIP 2022) \cite{chen2022content}} & 30.31 (±3.49) & 36.91 (±2.83) & 40.93 (±3.30) & 28.16 (±4.34) & 33.05 (±4.65) & 36.99 (±4.74) \\
\multicolumn{2}{l|}{TransCS (TIP 2022) \cite{shen2022transcs}} & 29.54 (±2.92) & 35.62 (±2.56) & 40.50 (±3.00) & 27.76 (±4.18) & 32.43 (±4.48) & 36.65 (±4.71) \\
\multicolumn{2}{l|}{OCTUF$^+$ (CVPR 2023) \cite{song2023optimization}} & 30.70 (±3.33) & 37.35 (±2.83) & 41.36 (±3.23) & 28.18 (±4.40) & 33.23 (±4.77) & 37.28 (±4.84) \\
\multicolumn{2}{l|}{CSformer (TIP 2023) \cite{ye2023csformer}} & 30.66 (±2.96) & - & 41.04 (±3.05) & 28.11 (±4.17) & - & 36.75 (±4.47) \\
\multicolumn{2}{l|}{PRL-PGD$^+$ (IJCV 2023) \cite{chen2023deep}} & \secondbest{31.70} (±3.00) & \secondbest{37.89} (±2.91) & \secondbest{41.78} (±3.27) & \secondbest{28.50} (±4.51) & \secondbest{33.41} (±4.79) & \secondbest{37.37} (±4.80) \\
\rowcolor[HTML]{FFEEED} \multicolumn{2}{l|}{\textbf{IDM (Ours, $T=3$)}} & \best{32.92} (±3.34) &  \best{38.85} (±3.28) & \best{42.48} (±3.28) & \best{28.69} (±5.19) & \best{34.67} (±5.57) & \best{39.57} (±5.85) \\ \shline
\end{tabular}}
\\ ~\vspace{3pt} \\
\resizebox{1.0\textwidth}{!}{
\begin{tabular}{lc|ccc|ccc}
\shline
\rowcolor[HTML]{EFEFEF} 
\multicolumn{1}{l|}{\cellcolor[HTML]{EFEFEF}} &
  Test Set &
  \multicolumn{3}{c|}{\cellcolor[HTML]{EFEFEF}Urban100} &
  \multicolumn{3}{c}{\cellcolor[HTML]{EFEFEF}DIV2K} \\ \hhline{>{\arrayrulecolor[HTML]{EFEFEF}}->{\arrayrulecolor{black}}|-------} 
\rowcolor[HTML]{EFEFEF} 
\multicolumn{1}{l|}{\multirow{-2}{*}{\cellcolor[HTML]{EFEFEF}Method}} & CS Ratio $\gamma$ & 10\% & 30\% & 50\% & 10\% & 30\% & 50\% \\ \hline \hline
\multicolumn{2}{l|}{ReconNet (CVPR 2016) \cite{kulkarni2016reconnet}} & 20.71 (±3.69) & 25.15 (±4.38) & 28.15 (±4.61) & 24.41 (±3.91) & 29.09 (±4.32) & 32.15 (±4.32) \\
\multicolumn{2}{l|}{ISTA-Net$^+$ (CVPR 2018) \cite{zhang2018ista}}    & 22.81 (±4.85) & 29.83 (±6.27) & 34.33 (±6.54) & 26.30 (±4.83) & 32.65 (±5.37) & 36.88 (±5.41) \\
\multicolumn{2}{l|}{CSNet$^+$ (TIP 2019) \cite{shi2019image}}       & 23.96 (±4.24) & 29.12 (±5.10) & 32.76 (±5.38) & 28.23 (±4.82) & 33.63 (±5.22) & 37.59 (±5.22) \\
\multicolumn{2}{l|}{SCSNet (CVPR 2019) \cite{shi2019scalable}}          & 24.22 (±4.36) & 29.41 (±5.19) & 33.31 (±5.37) & 28.41 (±4.84) & 33.85 (±5.31) & 37.97 (±5.18) \\
\multicolumn{2}{l|}{OPINE-Net$^+$ (JSTSP 2020) \cite{zhang2020optimization}} & 25.90 (±5.31) & 31.97 (±5.97) & 36.28 (±5.95) & 29.26 (±5.07) & 35.03 (±5.42) & 39.27 (±5.34) \\
\multicolumn{2}{l|}{DPA-Net (TIP 2020) \cite{sun2020dual}}         & 24.00 (±4.98) & 29.04 (±5.24) & - & 27.09 (±4.92) & 32.37 (±5.35) & - \\
\multicolumn{2}{l|}{MAC-Net (ECCV 2020) \cite{chen2020learning}} & 23.71 (±5.18) & 29.03 (±5.67) & 33.10 (±5.87) & 26.72 (±4.83) & 32.23 (±5.19) & 35.40 (±4.94) \\
\multicolumn{2}{l|}{RK-CCSNet (ECCV 2020) \cite{zheng2020sequential}} & - & - & 33.39 (±5.63) & - & - & 37.32 (±5.29) \\
\multicolumn{2}{l|}{ISTA-Net$^{++}$ (ICME 2021) \cite{you2021ista}} & 24.95 (±5.60) & 31.50 (±5.80) & 35.58 (±5.59) & 27.82 (±5.04) & 33.74 (±5.15) & 37.78 (±4.94)\\
\multicolumn{2}{l|}{COAST (TIP 2021) \cite{you2021coast}}           & 26.17 (±5.53) & 32.48 (±6.00) & 36.56 (±5.95) & 29.46 (±5.18) & 35.32 (±5.43) & 39.43 (±5.30) \\
\multicolumn{2}{l|}{AMP-Net (TIP 2021) \cite{zhang2021amp}}         & 25.32 (±5.01) & 31.63 (±5.75) & 35.91 (±5.71) & 29.08 (±5.09) & 35.41 (±5.30) & 39.46 (±5.15) \\
\multicolumn{2}{l|}{MADUN (ACM MM 2021) \cite{song2021memory}} & 26.23 (±5.17) & 33.00 (±5.89) & 36.69 (±5.81) & 29.62 (±4.92) & 36.04 (±5.47) & 40.06 (±5.33) \\
\multicolumn{2}{l|}{DGUNet$^+$ (CVPR 2022) \cite{mou2022deep}} & 27.42 (±5.28) & - & 37.13 (±5.46) & 30.25 (±5.29) & - & 40.33 (±5.42) \\
\multicolumn{2}{l|}{MR-CCSNet$^+$ (CVPR 2022) \cite{fan2022global}} & - & - & 34.40 (±5.63) & - & - & 38.16 (±5.40) \\
\multicolumn{2}{l|}{FSOINet (ICASSP 2022) \cite{chen2022fsoinet}} & 26.87 (±5.52) & 33.29 (±5.83) & 37.25 (±5.80) & 30.01 (±5.23) & 36.03 (±5.52) & 40.38 (±5.41) \\
\multicolumn{2}{l|}{CASNet (TIP 2022) \cite{chen2022content}} & 26.85 (±5.52) & 32.85 (±5.94) & 36.94 (±5.99) & 30.01 (±5.30) & 35.90 (±5.56) & 40.14 (±5.45) \\
\multicolumn{2}{l|}{TransCS (TIP 2022) \cite{shen2022transcs}} & 25.82 (±4.87) & 31.18 (±5.53) & 36.64 (±5.77) & 29.09 (±4.79) & 34.76 (±5.28) & 39.75 (±5.40) \\
\multicolumn{2}{l|}{OCTUF$^+$ (CVPR 2023) \cite{song2023optimization}} & 27.28 (±5.70) & 33.87 (±5.82) & 37.82 (±5.75) & 30.17 (±5.27) & 36.25 (±5.55) & 40.61 (±5.45) \\
\multicolumn{2}{l|}{CSformer (TIP 2023) \cite{ye2023csformer}} & 27.13 (±4.91) & - & 37.04 (±5.03) & 29.96 (±4.84) & - & 40.04 (±4.95) \\
\multicolumn{2}{l|}{PRL-PGD$^+$ (IJCV 2023) \cite{chen2023deep}} & \secondbest{28.77} (±5.76) & \secondbest{34.73} (±5.73) & \secondbest{38.57} (±5.66) & \secondbest{30.75} (±5.40) & \secondbest{36.64} (±5.60) & \secondbest{40.97} (±5.48) \\
\rowcolor[HTML]{FFEEED} \multicolumn{2}{l|}{\textbf{IDM (Ours, $T=3$)}} & \best{31.41} (±5.89) & \best{36.76} (±5.53) & \best{40.33} (±5.62) & \best{31.07} (±5.91) & \best{36.98} (±5.64) & \best{41.15} (±5.40) \\ \shline
\end{tabular}}
\end{table*}

\subsection{Discussion}
Our proposed method reinterprets the diffusion framework DDNM for image CS reconstruction, moving beyond the existing paradigm of posterior sampling \cite{chung2023diffusion,rout2023solving}. In this section, we discuss the relationship between our IDM method and previous diffusion-based approaches that employ posterior sampling, as well as the connections and differences between IDM and algorithm unrolling \cite{zhang2023physics} techniques.

\subsubsection{Relationship with Previous Diffusion-Based Methods}
Traditional diffusion models based on DDPM \cite{ho2020denoising} often perform image generation and reconstruction by modeling the reverse diffusion process through posterior sampling. These models solve stochastic differential equations (SDEs) or ordinary differential equations (ODEs) and gradually denoise a randomly initialized noise to obtain a sample from the desired data distribution, typically requiring a large number of iterative steps (e.g., hundreds or thousands). In contrast, IDM repurposes the diffusion sampling process as an end-to-end deterministic mapping tailored specifically for image CS reconstruction. By fine-tuning a pre-trained diffusion model, we directly learn the mapping from measurement $\y$ to the original image $\x$ using a limited number of sampling steps (e.g., $T=3$). To be specific, this approach diverges from previous diffusion methods in four key aspects:

Firstly, our IDM method eliminates the randomness in diffusion models by initializing the sampling process with a deterministic estimate, $\xhat_T = \sqrt{\bar{\alpha}_T} \A^\dagger \y$. This deterministic initialization improves performance, as demonstrated in our ablation studies (see Tab. \ref{tab:abla}, variant (8) vs. (9)).

Secondly, our approach fine-tunes all parameters of the diffusion model, including the noise estimation network $\ep_\Th$, directly on the CS reconstruction task using an $L_1$ loss. This end-to-end training aligns the model’s capacity with CS-specific requirements, optimizing both diffusion sampling process and noise estimator for accurate reconstruction.

Thirdly, by adapting the sampling process of DDNM to CS and leveraging powerful pre-trained SD \cite{stabilityai} models, our method achieves high-quality reconstruction with significantly fewer steps (e.g., $T=3$) compared to hundreds or thousands of steps commonly used in previous methods.

Fourthly, due to its deterministic nature and limited diffusion sampling steps, our IDM may not strictly align with the interpretation of solving SDEs or ODEs via posterior sampling. Instead, it functions as an iterative reconstruction model optimized specifically for image CS. Although our method may not fully adhere to the theoretical framework of posterior sampling, we focus on practical gains in performance and efficiency for CS reconstruction. The substantial performance improvements achieved by our method justify this approach, as shown by our experimental results.

\begin{table*}[!t]
\caption{\textbf{Comparison of average PSNR (dB, $\uparrow$) and SSIM ($\uparrow$) (±std) across eight diffusion-based methods} on three-channel RGB test images.}
\vspace{-10pt}
\label{tab:compare_sota_diff_quality}
\centering
\resizebox{1.0\textwidth}{!}{
\begin{tabular}{lc|ccc}
\shline
\rowcolor[HTML]{EFEFEF} 
\multicolumn{1}{c|}{\cellcolor[HTML]{EFEFEF}} &
Test Set &
\multicolumn{3}{c}{\cellcolor[HTML]{EFEFEF}Urban100 (100 RGB images of size $256\times 256$)} \\ \hhline{>{\arrayrulecolor[HTML]{EFEFEF}}->{\arrayrulecolor{black}}|---->{\arrayrulecolor[HTML]{EFEFEF}}>{\arrayrulecolor{black}}} 
\rowcolor[HTML]{EFEFEF} \multicolumn{1}{l|}{\multirow{-2}{*}{\cellcolor[HTML]{EFEFEF}Method}} &
CS Ratio $\gamma$ & 10\% & 30\% & 50\% \\ \hline \hline
\multicolumn{2}{l|}{DDRM (NeurIPS 2022) \cite{kawar2022denoising}} & 19.16 (±2.86)/0.4348 (±0.1185) & \secondbest{28.91} (±3.96)/\secondbest{0.8518} (±0.0717) & \secondbest{33.61} (±3.78)/\secondbest{0.9365} (±0.0342) \\
\multicolumn{2}{l|}{$\Pi$GDM (ICLR 2023) \cite{song2023pseudoinverse}} & 20.09 (±3.67)/\secondbest{0.5089} (±0.1818) & 26.70 (±4.60)/0.7925 (±0.1164) & 29.75 (±4.01)/0.8724 (±0.0904) \\
\multicolumn{2}{l|}{DPS (ICLR 2023) \cite{chung2023diffusion}} & 17.12 (±4.26)/0.3270 (±0.2110) & 18.47 (±4.48)/0.3891 (±0.2229) & 19.21 (±4.57)/0.4308 (±0.2277) \\
\multicolumn{2}{l|}{DDNM (ICLR 2023) \cite{wang2023zero}} & \secondbest{20.76} (±4.51)/0.4682 (±0.1727) & 28.76 (±4.98)/0.8284 (±0.0948) & 32.86 (±4.87)/0.9164 (±0.0539) \\
\multicolumn{2}{l|}{GDP (CVPR 2023) \cite{fei2023generative}} & 20.74 (±4.73)/0.5075 (±0.1686) & 24.81 (±4.31)/0.7086 (±0.1338) & 26.12 (±3.62)/0.7711 (±0.1170) \\
\multicolumn{2}{l|}{PSLD (NeurIPS 2023) \cite{rout2023solving}} & 19.43 (±5.04)/0.4054 (±0.2311) & 22.42 (±4.15)/0.6399 (±0.1859) & 22.78 (±3.90)/0.6882 (±0.1656) \\
\multicolumn{2}{l|}{SR3 (TPAMI 2023) \cite{saharia2023image}} & 18.90 (±3.05)/0.5023 (±0.1467) & 21.37 (±3.08)/0.6428 (±0.1200) & 23.12 (±2.99)/0.7233 (±0.0981) \\
\rowcolor[HTML]{FFEEED} \multicolumn{2}{l|}{\textbf{IDM (Ours, $T=3$)}} &  \best{30.85} (±5.32)/\best{0.8970} (±0.0855) & \best{35.78} (±5.05)/\best{0.9570} (±0.0432) & \best{39.16} (±5.11)/\best{0.9771} (±0.0244) \\ \shline
\end{tabular}}
\\ ~\vspace{3pt} \\
\resizebox{1.0\textwidth}{!}{
\begin{tabular}{lc|ccc}
\shline
\rowcolor[HTML]{EFEFEF} 
\multicolumn{1}{c|}{\cellcolor[HTML]{EFEFEF}} &
Test Set &
\multicolumn{3}{c}{\cellcolor[HTML]{EFEFEF}DIV2K (100 RGB images of size $256\times 256$)} \\ \hhline{>{\arrayrulecolor[HTML]{EFEFEF}}->{\arrayrulecolor{black}}|---->{\arrayrulecolor[HTML]{EFEFEF}}>{\arrayrulecolor{black}}} 
\rowcolor[HTML]{EFEFEF} \multicolumn{1}{l|}{\multirow{-2}{*}{\cellcolor[HTML]{EFEFEF}Method}} &
CS Ratio $\gamma$ & 10\% & 30\% & 50\% \\ \hline \hline
\multicolumn{2}{l|}{DDRM (NeurIPS 2022) \cite{kawar2022denoising}} & 20.91 (±3.53)/0.4323 (±0.1232) & \secondbest{28.91} (±3.93)/\secondbest{0.7852} (±0.0815) & \secondbest{33.68} (±3.72)/\secondbest{0.9057} (±0.0391) \\
\multicolumn{2}{l|}{$\Pi$GDM (ICLR 2023) \cite{song2023pseudoinverse}} & 22.46 (±3.71)/0.5551 (±0.1515) & 28.01 (±3.74)/0.7714 (±0.1058) & 30.79 (±3.47)/0.8497 (±0.0797) \\
\multicolumn{2}{l|}{DPS (ICLR 2023) \cite{chung2023diffusion}} & 19.47 (±4.36)/0.4222 (±0.1827) & 20.78 (±4.35)/0.4652 (±0.1809) & 21.37 (±4.30)/0.4884 (±0.1785) \\
\multicolumn{2}{l|}{DDNM (ICLR 2023) \cite{wang2023zero}} & 22.18 (±4.30)/0.4383 (±0.1435) & 28.50 (±4.90)/0.7422 (±0.1110) & 32.74 (±4.88)/0.8726 (±0.0629) \\
\multicolumn{2}{l|}{GDP (CVPR 2023) \cite{fei2023generative}} & \secondbest{25.17} (±4.44)/\secondbest{0.6334} (±0.1274) & 27.75 (±2.85)/0.7664 (±0.0887) & 28.47 (±1.53)/0.8029 (±0.0682) \\
\multicolumn{2}{l|}{PSLD (NeurIPS 2023) \cite{rout2023solving}} & 21.30 (±4.35)/0.4427 (±0.1683) & 23.87 (±4.06)/0.6473 (±0.1501) & 24.41 (±3.83)/0.7124 (±0.1200) \\
\multicolumn{2}{l|}{SR3 (TPAMI 2023) \cite{saharia2023image}} & 20.10 (±2.93)/0.4794 (±0.1415) & 22.07 (±3.23)/0.5689 (±0.1282) &  23.63 (±2.77)/0.6518 (±0.1106) \\
\rowcolor[HTML]{FFEEED} \multicolumn{2}{l|}{\textbf{IDM (Ours, $T=3$)}} & \best{31.22} (±5.43)/\best{0.8581} (±0.0888) & \best{36.83} (±5.17)/\best{0.9482} (±0.0382) & \best{40.81} (±4.96)/\best{0.9755} (±0.0200) \\ \shline
\end{tabular}}
\end{table*}

\begin{table*}[!t]
\caption{\textbf{Comparison of FID ($\downarrow$) and LPIPS ($\downarrow$) across eight diffusion-based methods} on three-channel RGB benchmark images.}
\vspace{-10pt}
\label{tab:compare_sota_diff_fid_lpips}
\centering
\resizebox{1.0\textwidth}{!}{
\begin{tabular}{lc|ccc|ccc}
\shline
\rowcolor[HTML]{EFEFEF}
\multicolumn{1}{l|}{\cellcolor[HTML]{EFEFEF}} &
  Test Set &
  \multicolumn{3}{c|}{\cellcolor[HTML]{EFEFEF}Urban100 (100 RGB images of size $256\times 256$)} &
  \multicolumn{3}{c}{\cellcolor[HTML]{EFEFEF}DIV2K (100 RGB images of size $256\times 256$)} \\ \hhline{>{\arrayrulecolor[HTML]{EFEFEF}}->{\arrayrulecolor{black}}|------->{\arrayrulecolor{black}}} 
\rowcolor[HTML]{EFEFEF} 
\multicolumn{1}{l|}{\multirow{-2}{*}{\cellcolor[HTML]{EFEFEF}Method}} &
  CS Ratio $\gamma$ &
  10\% &
  30\% &
  50\% &
  10\% &
  30\% &
  50\%  \\ \hline \hline
\multicolumn{2}{l|}{DDRM (NeurIPS 2022) \cite{graikos2022diffusion}} & \secondbest{184.95}/0.4844 & \secondbest{50.71}/\secondbest{0.1495} & \secondbest{21.76}/\secondbest{0.0690} & 264.46/0.5428 & \secondbest{93.66}/\secondbest{0.2484} & \secondbest{39.78}/\secondbest{0.1211}  \\
\multicolumn{2}{l|}{$\Pi$GDM (ICLR 2023) \cite{song2023pseudoinverse}} &  209.43/0.4873 & 108.90/0.2650 & 76.93/0.1928 & 254.26/0.5322 & 165.39/0.3511 &  134.47/0.2809  \\
\multicolumn{2}{l|}{DPS (ICLR 2023) \cite{chung2023diffusion}} & 313.62/0.6876 & 281.52/0.6238 & 271.42/0.5838 & 317.67/0.6797 & 300.14/0.6355 & 299.35/0.6090  \\
\multicolumn{2}{l|}{DDNM (ICLR 2023) \cite{wang2023zero}} & 212.84/0.5208 & 56.60/0.2001 & 23.74/0.1067 & 269.37/0.5638 & 111.64/0.3143 & 52.42/0.1819  \\
\multicolumn{2}{l|}{GDP (CVPR 2023) \cite{fei2023generative}} & 211.52/0.4636 & 118.38/0.3005 & 88.00/0.2383 & \secondbest{202.66}/\secondbest{0.4142} & 116.52/0.2894 & 95.81/0.2448 \\
\multicolumn{2}{l|}{PSLD (NeurIPS 2023) \cite{rout2023solving}} &  263.62/0.5749 & 157.69/0.3805 & 128.81/0.3277 & 258.68/0.5768 & 191.54/0.4259 & 164.17/0.3676 \\
\multicolumn{2}{l|}{SR3 (TPAMI 2023) \cite{saharia2023image}} &  210.77/\secondbest{0.4493} &  144.52/0.3406 &  103.67/0.2724 & 255.78/0.5024 &  193.18/0.4116 &  160.33/0.3548 \\
\rowcolor[HTML]{FFEEED} \multicolumn{2}{l|}{\textbf{IDM (Ours, $T=3$)}} &  \best{42.72}/\best{0.1253} & \best{15.46}/\best{0.0566} & \best{6.47}/\best{0.0296} & \best{75.44}/\best{0.1997} & \best{27.29}/\best{0.0886} & \best{10.69}/\best{0.0432} \\ \shline
\end{tabular}}
\end{table*}

\begin{table*}[!t]
\centering
\caption{\textbf{Comparison of average inference time and storage across diffusion-based methods} on an NVIDIA A100 GPU for a $256\times 256$ RGB image.}
\vspace{-10pt}
\label{tab:compare_sota_diff_complexity}
\resizebox{1.0\textwidth}{!}{
\begin{tabular}{l|ccccccc>{\columncolor[HTML]{FFEEED}}c}
\shline
\rowcolor[HTML]{EFEFEF} Method & DDRM \cite{kawar2022denoising} & $\Pi$GDM \cite{song2023pseudoinverse} & DPS \cite{chung2023diffusion} & DDNM \cite{wang2023zero} & GDP \cite{fei2023generative} & PSLD \cite{rout2023solving} & SR3 \cite{saharia2023image} & \textbf{IDM (Ours)} \\ \hline \hline
Inference Time (s, $\downarrow$) & 9.28 & 18.25 & 19.02 & \secondbest{9.16} & 21.35 & 233.10 & 35.62 & \best{0.63} \\
Model Size (GB, $\downarrow$) & \secondbest{2.1} & \secondbest{2.1} & \secondbest{2.1} & \secondbest{2.1} & \secondbest{2.1} & 4.3 & \best{0.4} & \best{0.4}\\ \shline
\end{tabular}}
\vspace{-10pt}
\end{table*}

\subsubsection{Relationship with Deep Algorithm Unrolling}
Deep algorithm-unrolling NNs \cite{monga2021algorithm,zhang2023physics} transforms iterative optimization algorithms into trainable NN architectures by unrolling a fixed number of iterations into NN layers. Each layer corresponds to one iteration of the original algorithm, and the entire NN is trained end-to-end to enhance performance on specific tasks such as image reconstruction. Our method shares similarities with deep unrolling, as we interpret the iterative steps of the diffusion sampling process as layers of a reconstruction model, which are then fine-tuned end-to-end for the CS task. However, our method differs from algorithm unrolling in the following four aspects:

Firstly, unlike traditional algorithm-unrolled NNs that are trained from scratch with randomly initialized weights, IDM leverages the large-scale pre-trained diffusion model SD v1.5. This enables us to capitalize on the rich representations learned from extensive datasets, providing a strong foundation that enhances reconstruction performance.

Secondly, traditional unrolling approaches are based on specific optimization algorithms (e.g., ISTA, ADMM) and inherit their convergence properties and interpretability. In contrast, our method is rooted in the diffusion framework, inherently different from conventional optimization algorithms. This shift allows us to explore new architectures and training strategies that are not constrained by the limitations of traditional optimization-based methods.

Thirdly, to address the computational challenges of training large-scale diffusion models end-to-end, we introduce a novel two-level invertible design that reduces memory consumption and enables model's efficient training on standard GPUs. This innovation is specific to our approach and is not commonly found in traditional deep unrolling methods.

Fourthly, by reusing and fine-tuning the pre-trained SD models, our method can be quickly adapted to different inverse problems like image CS, accelerated MRI, and sparse-view CT with minimal training effort. This scalability is advantageous compared to unrolled NNs, which often require extensive retraining or redesign when applied to new tasks.

Overall, while IDM shares the iterative nature of deep unrolling, the use of diffusion models and their benefits distinguish our method. We bridge the gap between diffusion-based generative modeling and inverse problem-solving in a new way, showing that diffusion models can be effectively adapted and fine-tuned for high-quality CS reconstruction.

\begin{figure}[!t]
\setlength{\tabcolsep}{0.5pt}
\centering
\resizebox{\linewidth}{!}{
\scriptsize
\begin{tabular}{ccccc>{\columncolor[HTML]{FFEEED}}c}
Ground Truth & CSNet$^+$ & OPINE-Net$^+$ & TransCS & PRL-PGD$^+$ & \textbf{IDM (Ours)}\\
\includegraphics[width=0.09\textwidth]{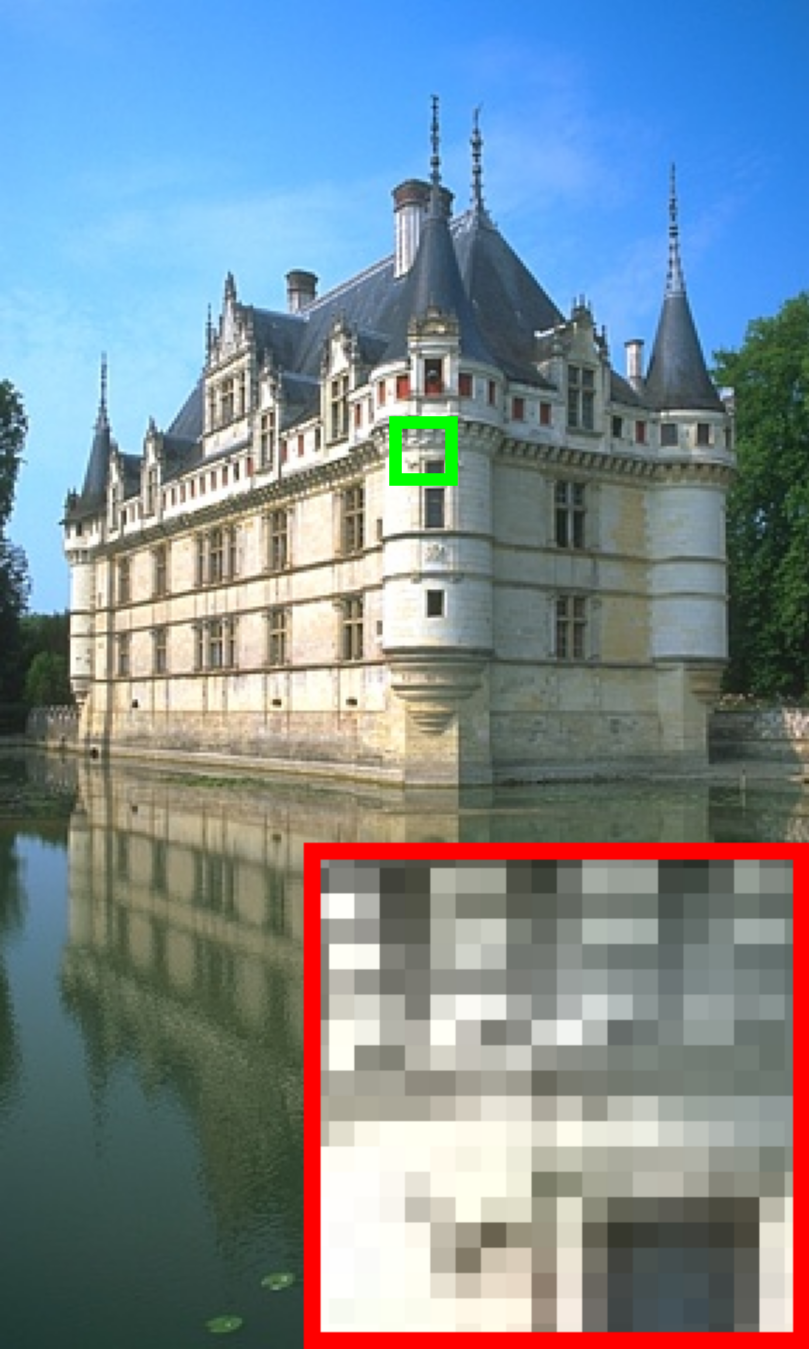}
&\includegraphics[width=0.09\textwidth]{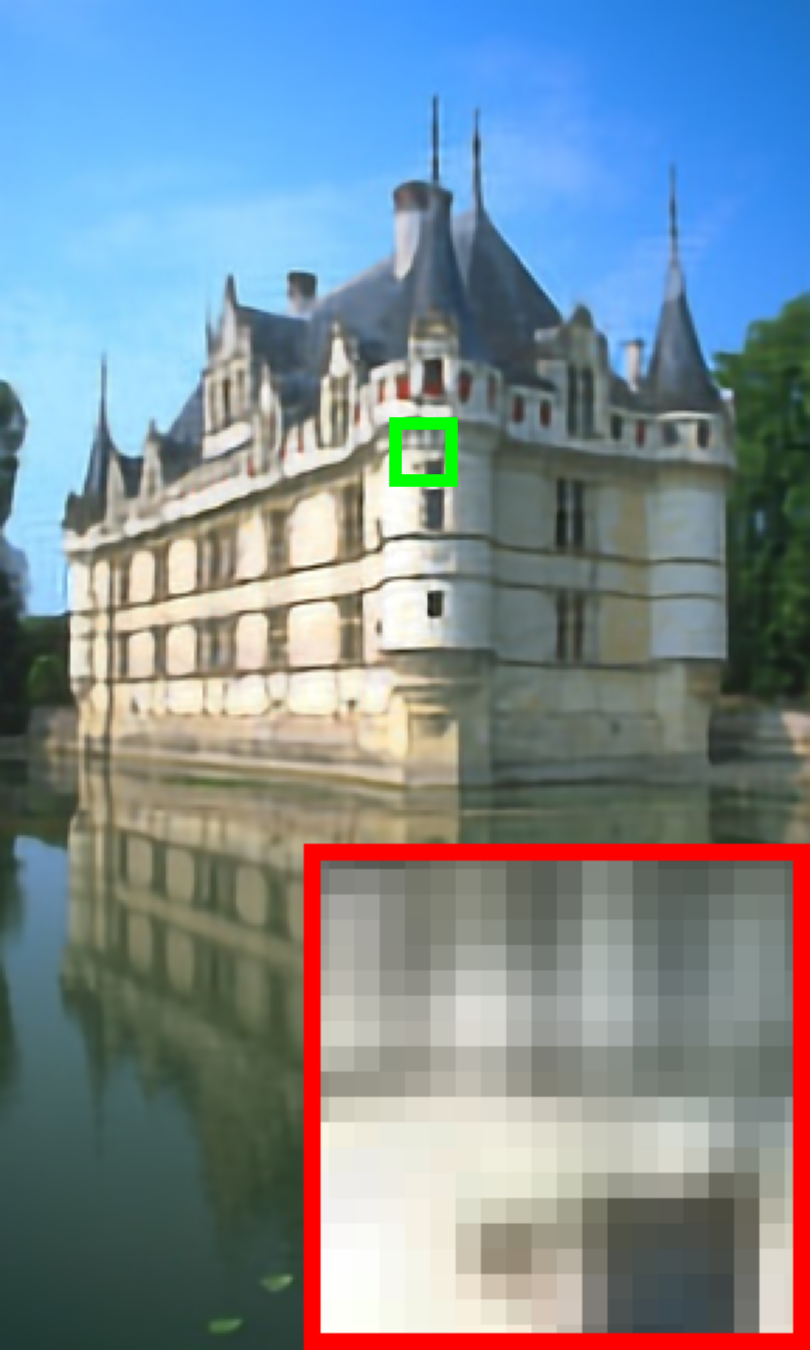}
&\includegraphics[width=0.09\textwidth]{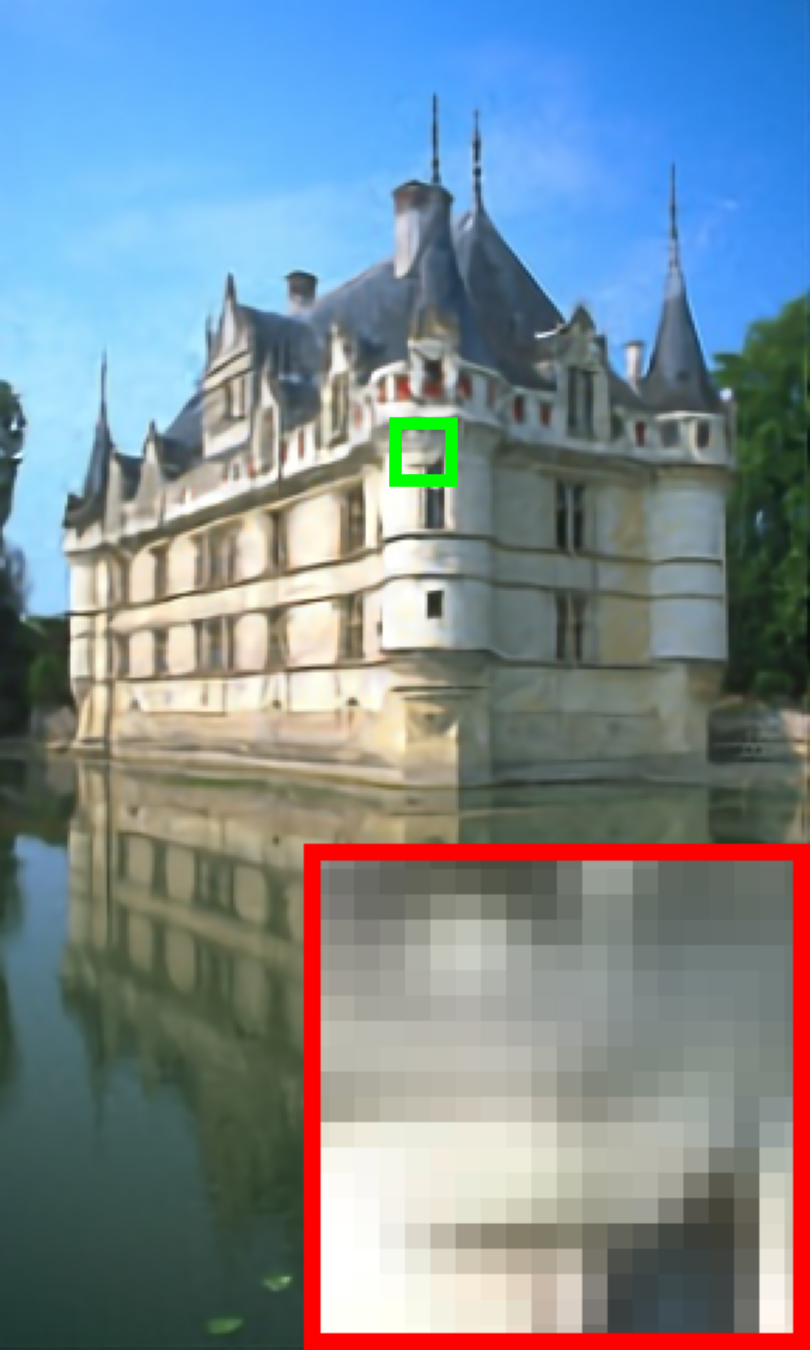}
&\includegraphics[width=0.09\textwidth]{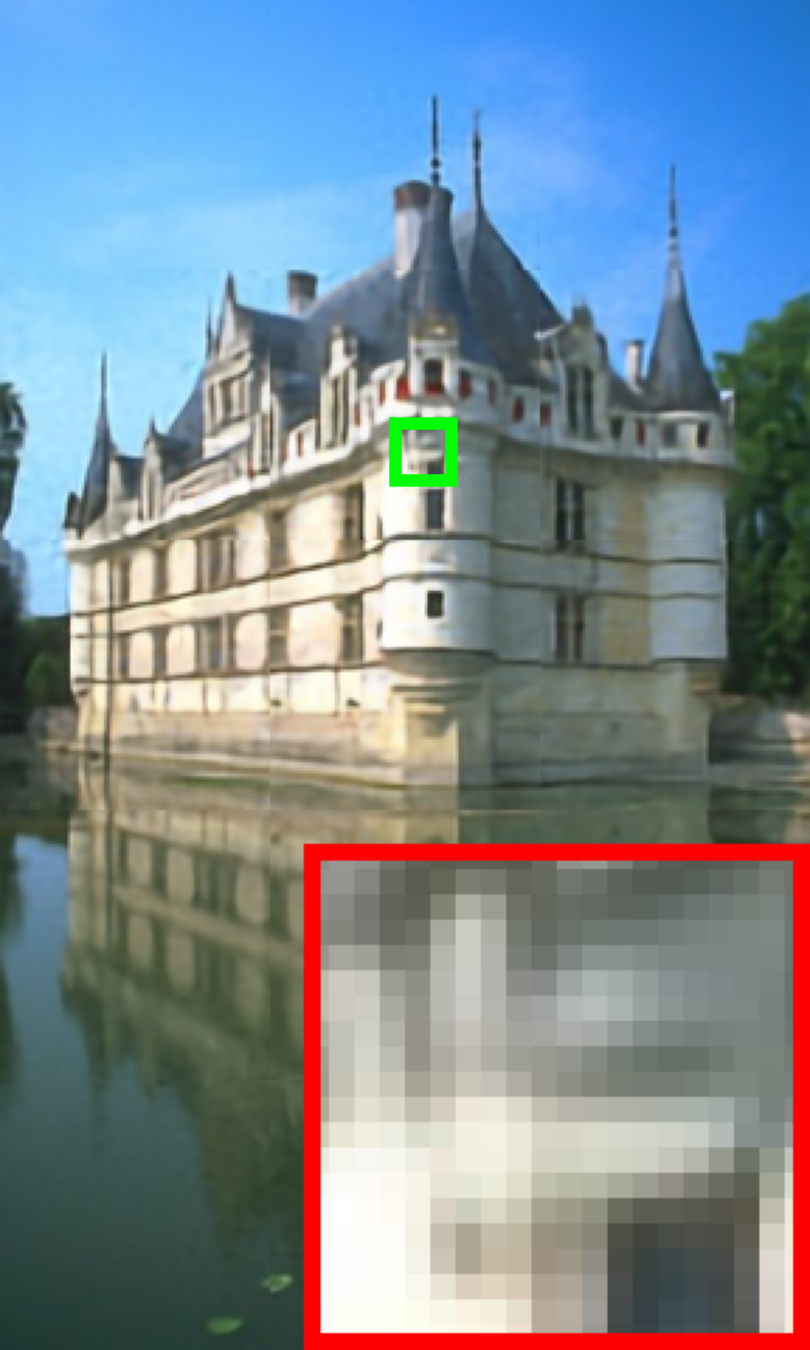}
&\includegraphics[width=0.09\textwidth]{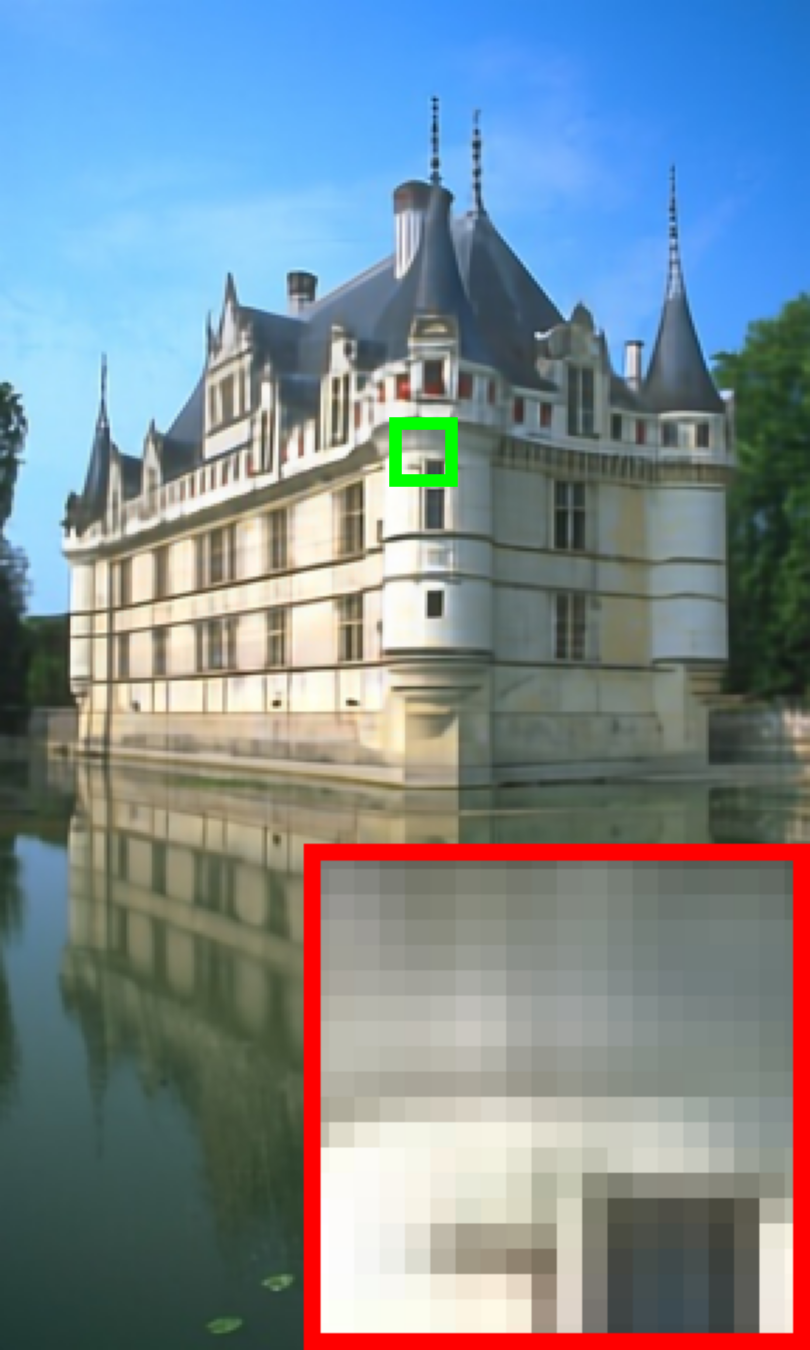}
&\includegraphics[width=0.09\textwidth]{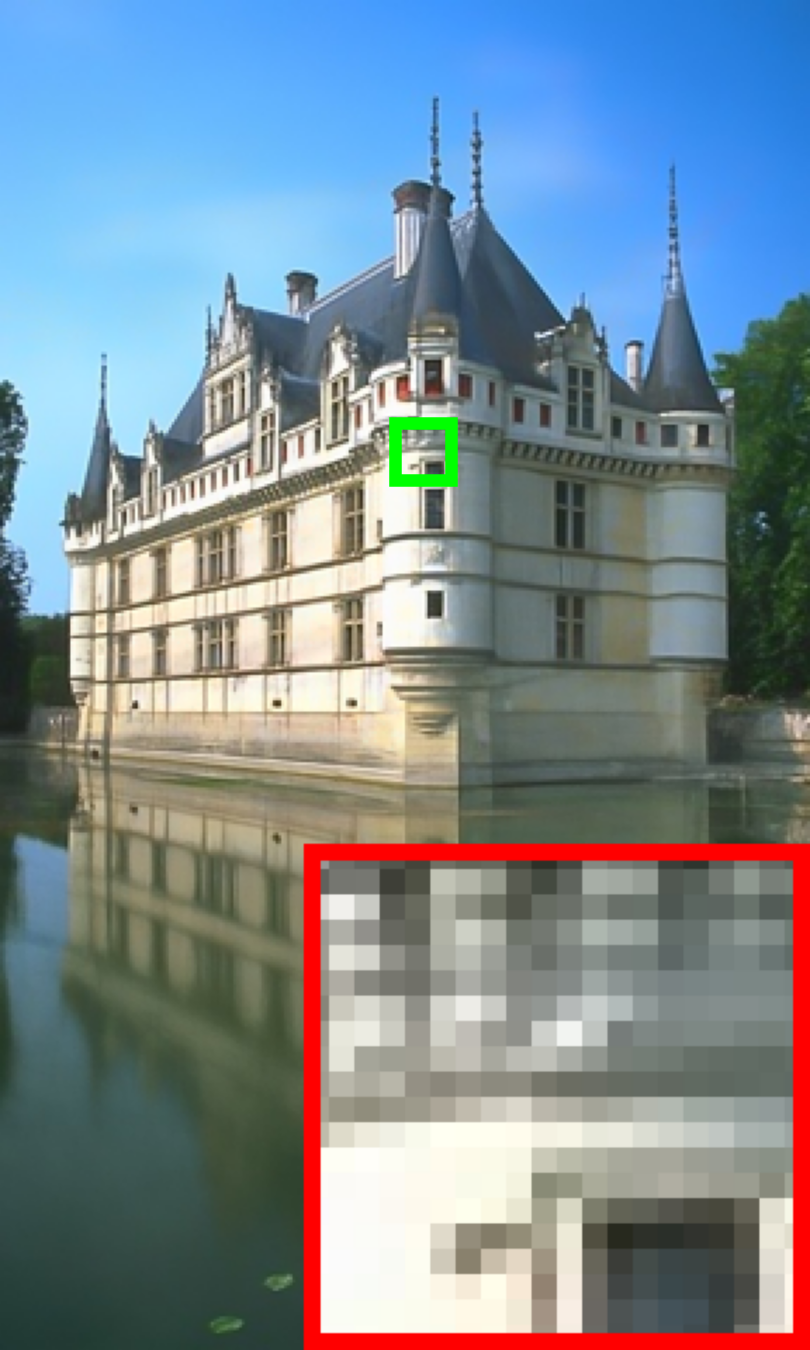}\\
PSNR/SSIM & 26.09/0.8369 & 26.56/0.8544 & 26.82/0.8594 & \secondbest{28.11}/\secondbest{0.8855} & \best{31.02}/\best{0.8859}
\end{tabular}}
\vspace{-10pt}
\caption{\textbf{Comparison of CS recovery results among various end-to-end learned methods} on ``test\_03" image from CBSD68 at $\gamma =10\%$.}
\label{fig:comp_cbsd68}
\end{figure}

\begin{figure}[!t]
\setlength{\tabcolsep}{0.5pt}
\centering
\resizebox{\linewidth}{!}{
\scriptsize
\begin{tabular}{ccccc}
Ground Truth & $\A^\dagger \y$ & DDRM & $\Pi$GDM & DPS\\
\includegraphics[width=0.09\textwidth]{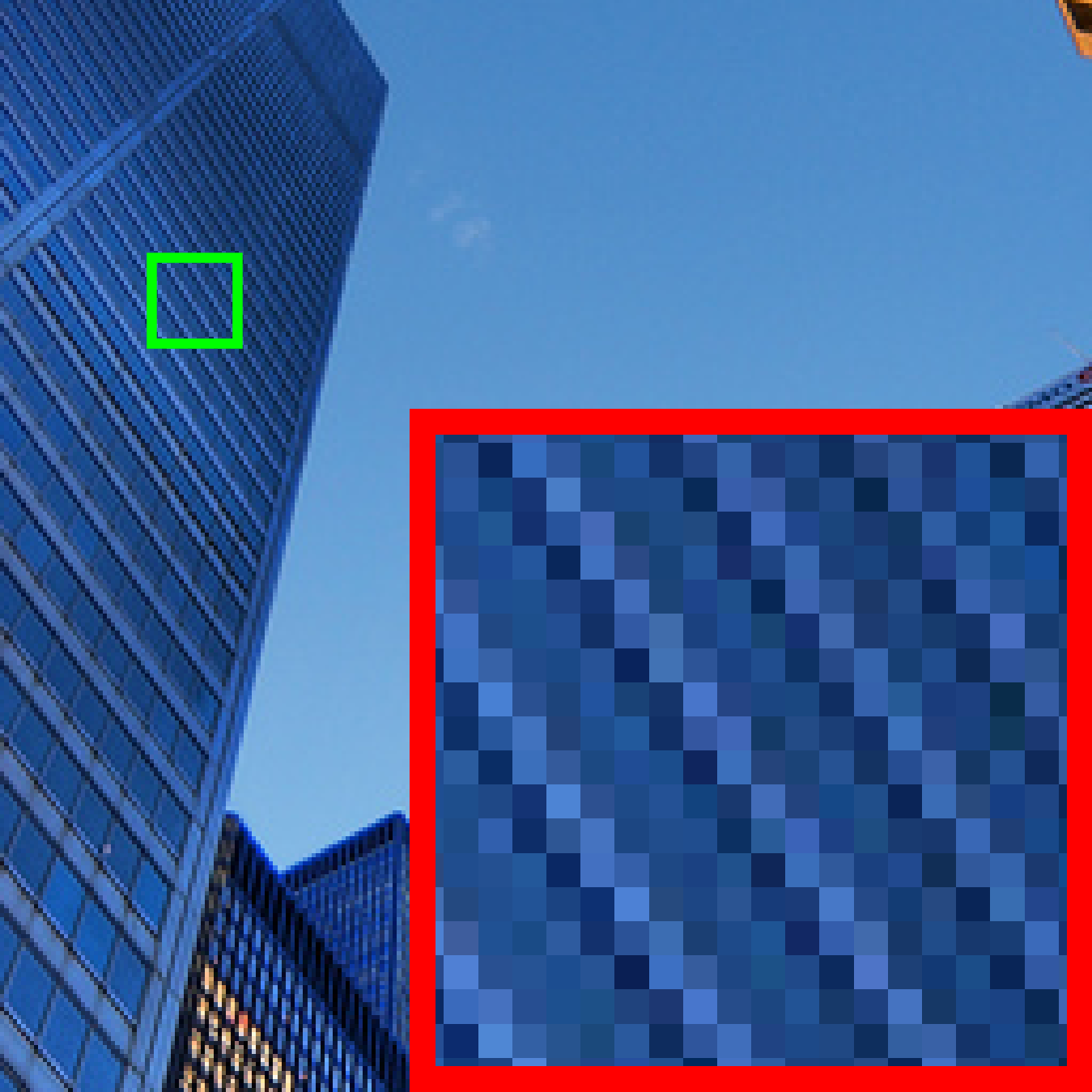}
&\includegraphics[width=0.09\textwidth]{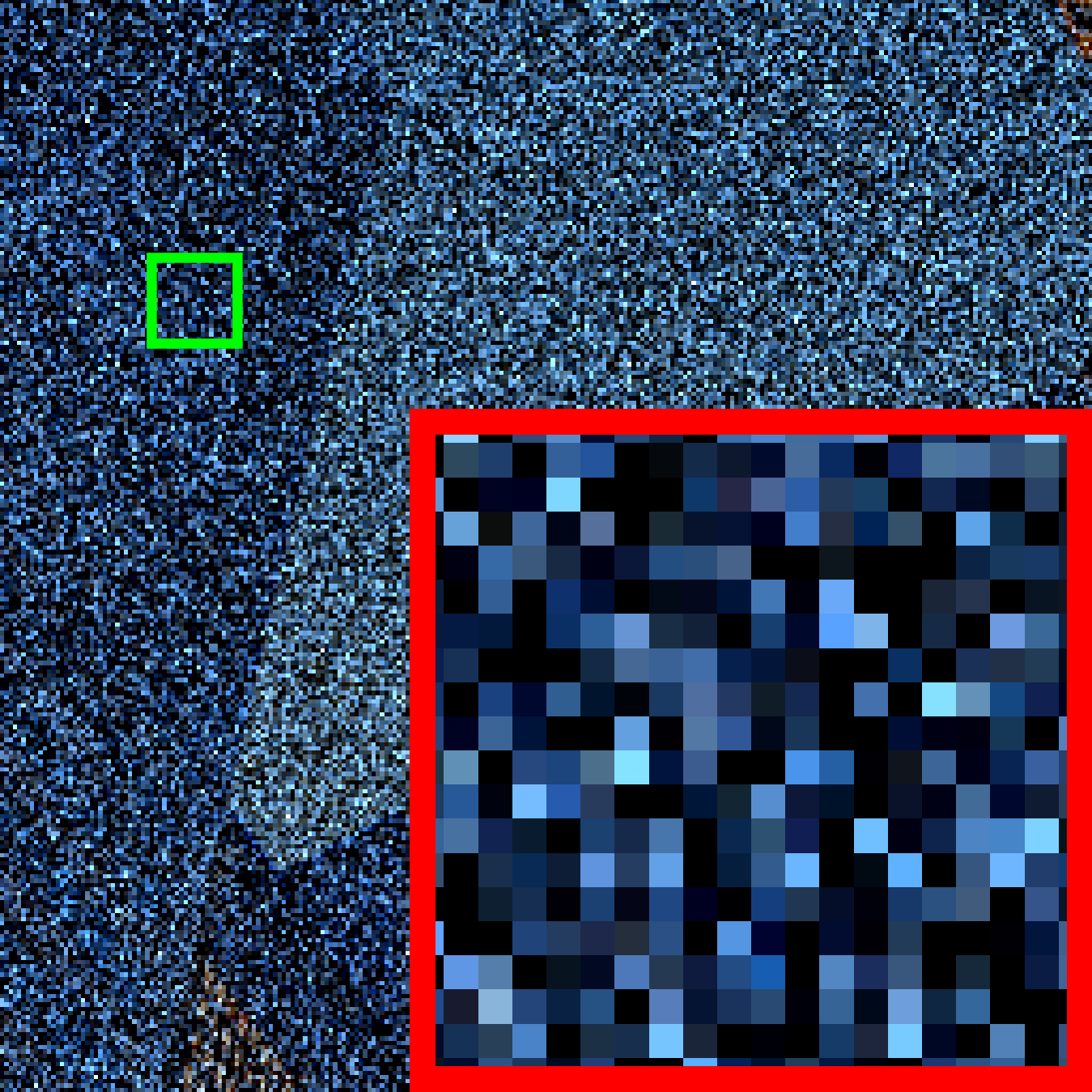}
&\includegraphics[width=0.09\textwidth]{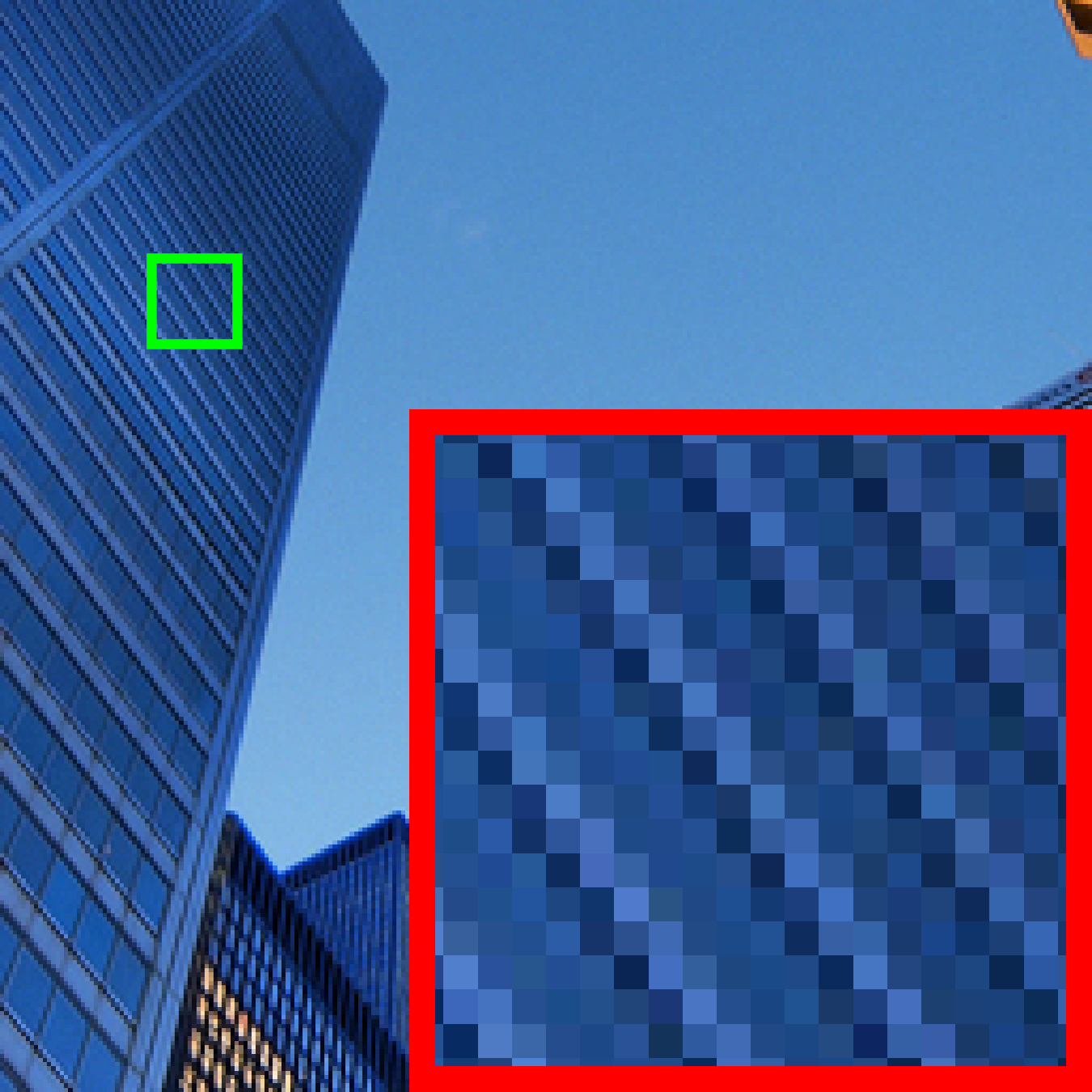}
&\includegraphics[width=0.09\textwidth]{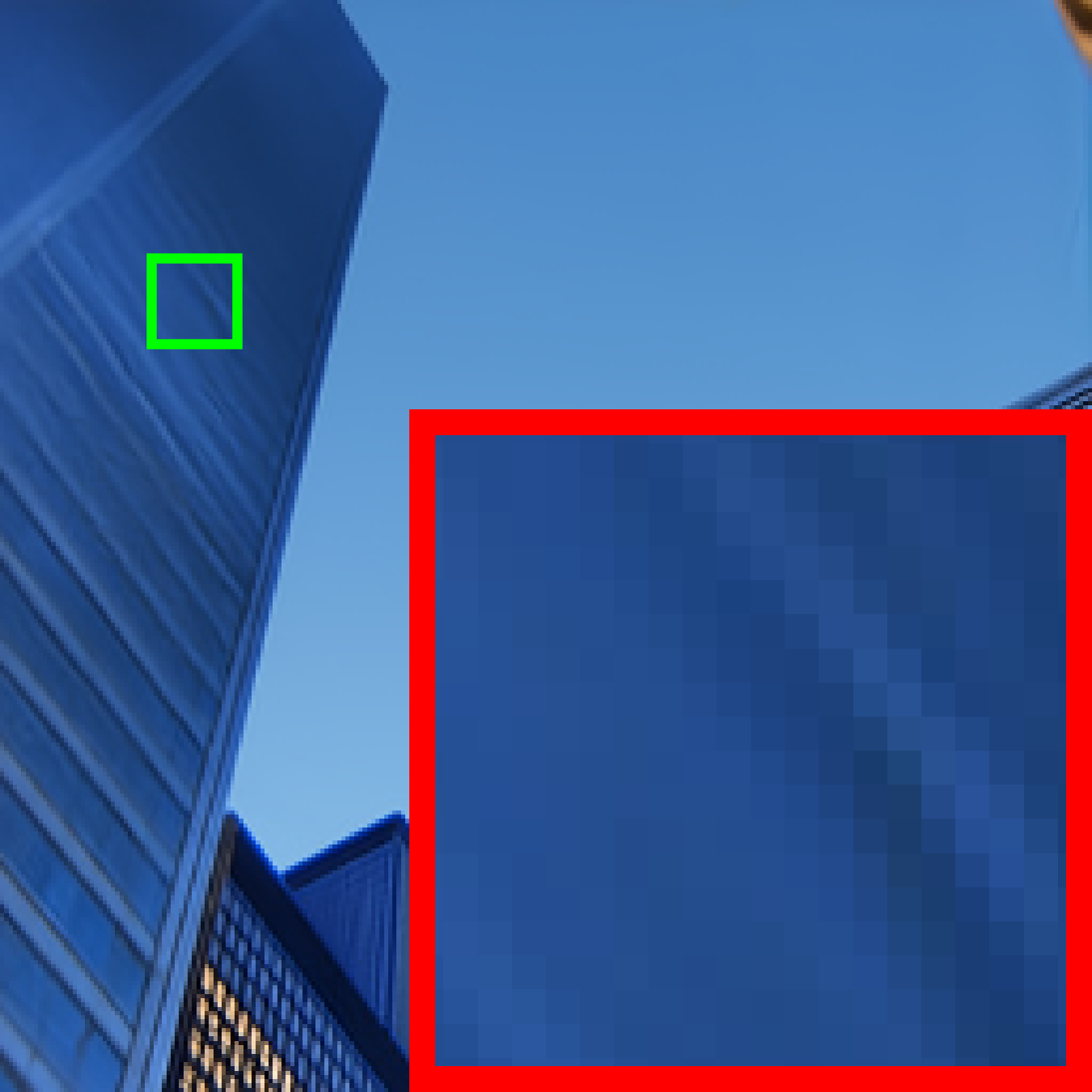}
&\includegraphics[width=0.09\textwidth]{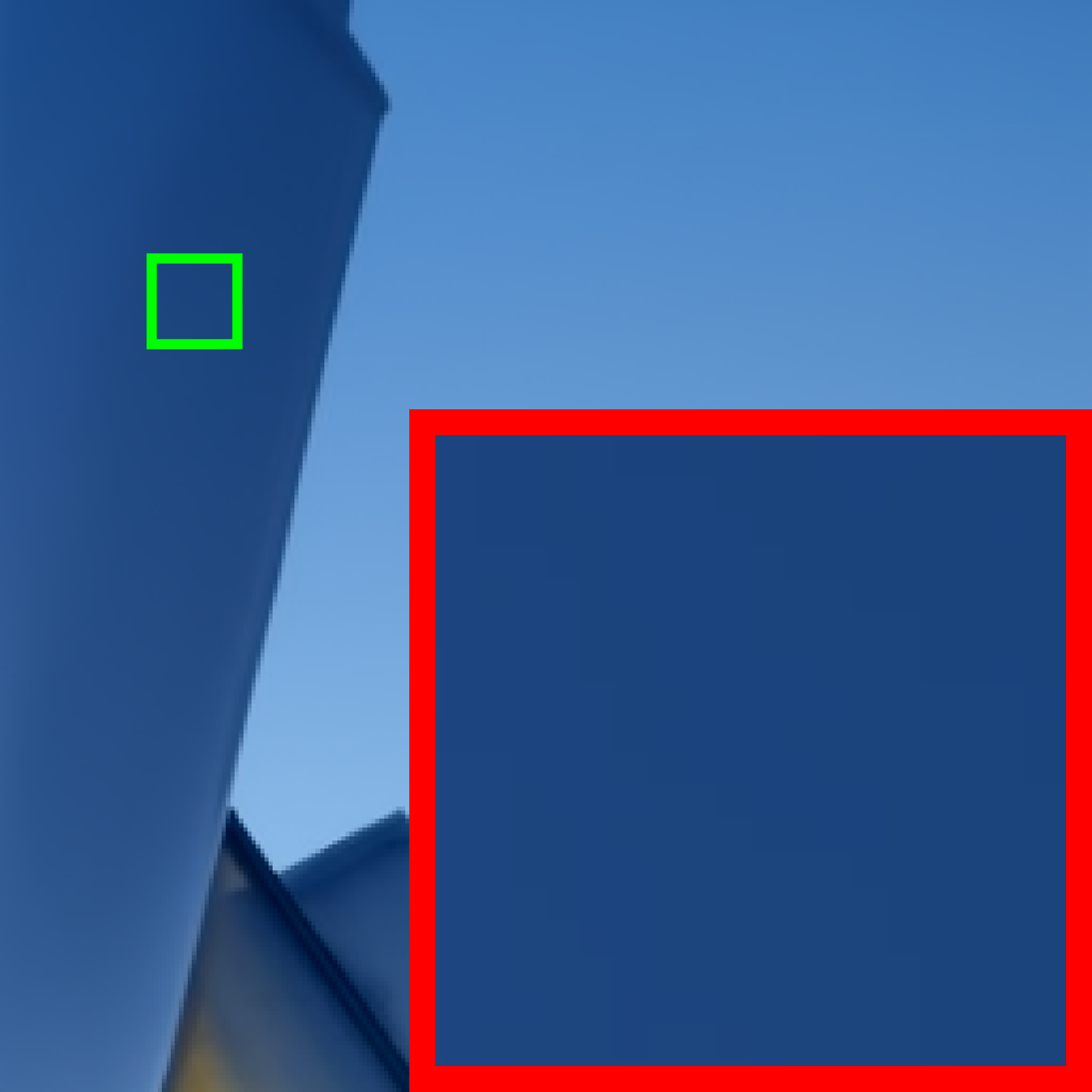}\\
PSNR/SSIM & 8.79/0.0600 & 36.20/\secondbest{0.9330} & 27.76/0.8317 & 22.27/0.7109
\end{tabular}}
\resizebox{\linewidth}{!}{
\scriptsize
\begin{tabular}{cccc>{\columncolor[HTML]{FFEEED}}c}
DDNM & GDP & PSLD & SR3 & \textbf{IDM (Ours)}\\
\includegraphics[width=0.09\textwidth]{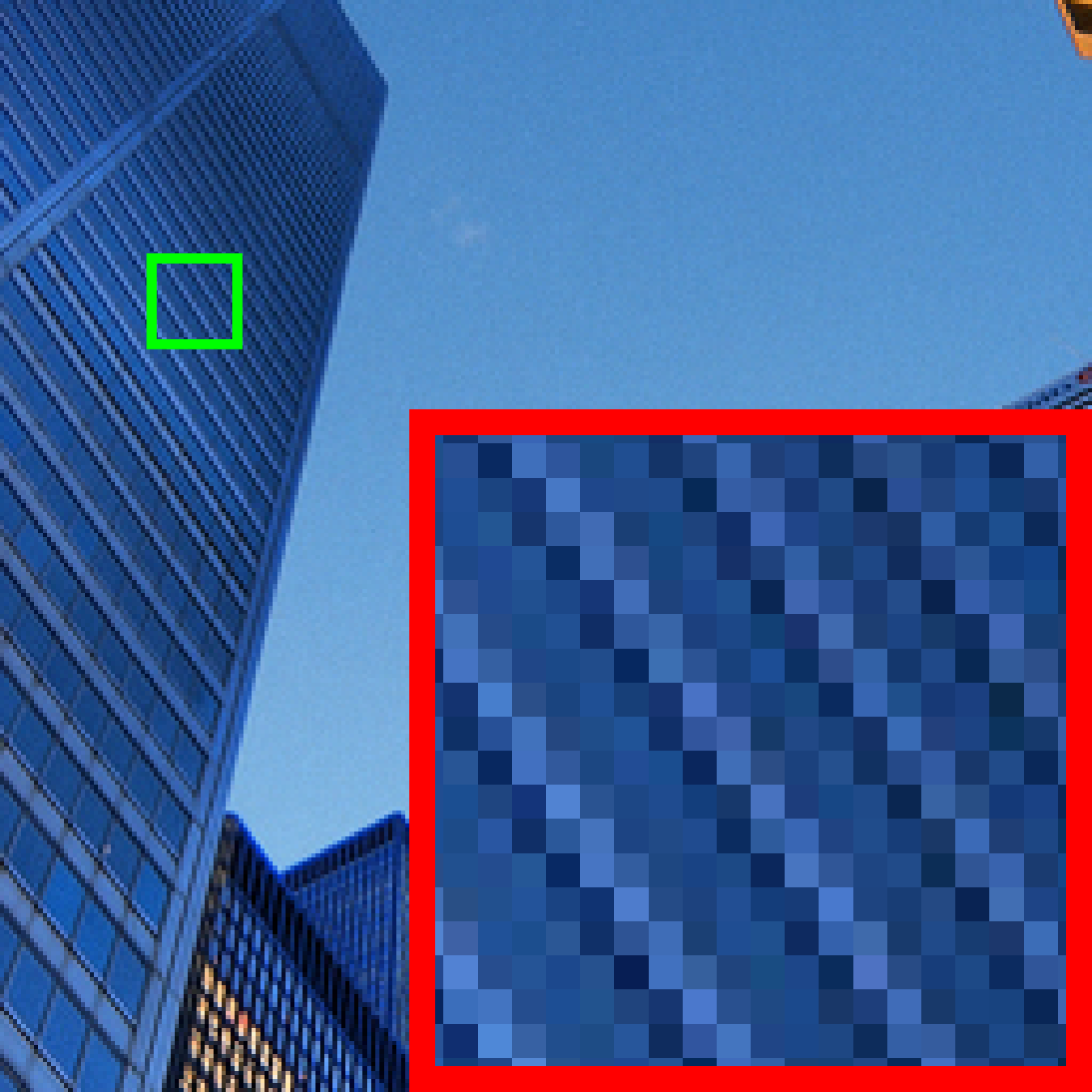}
&\includegraphics[width=0.09\textwidth]{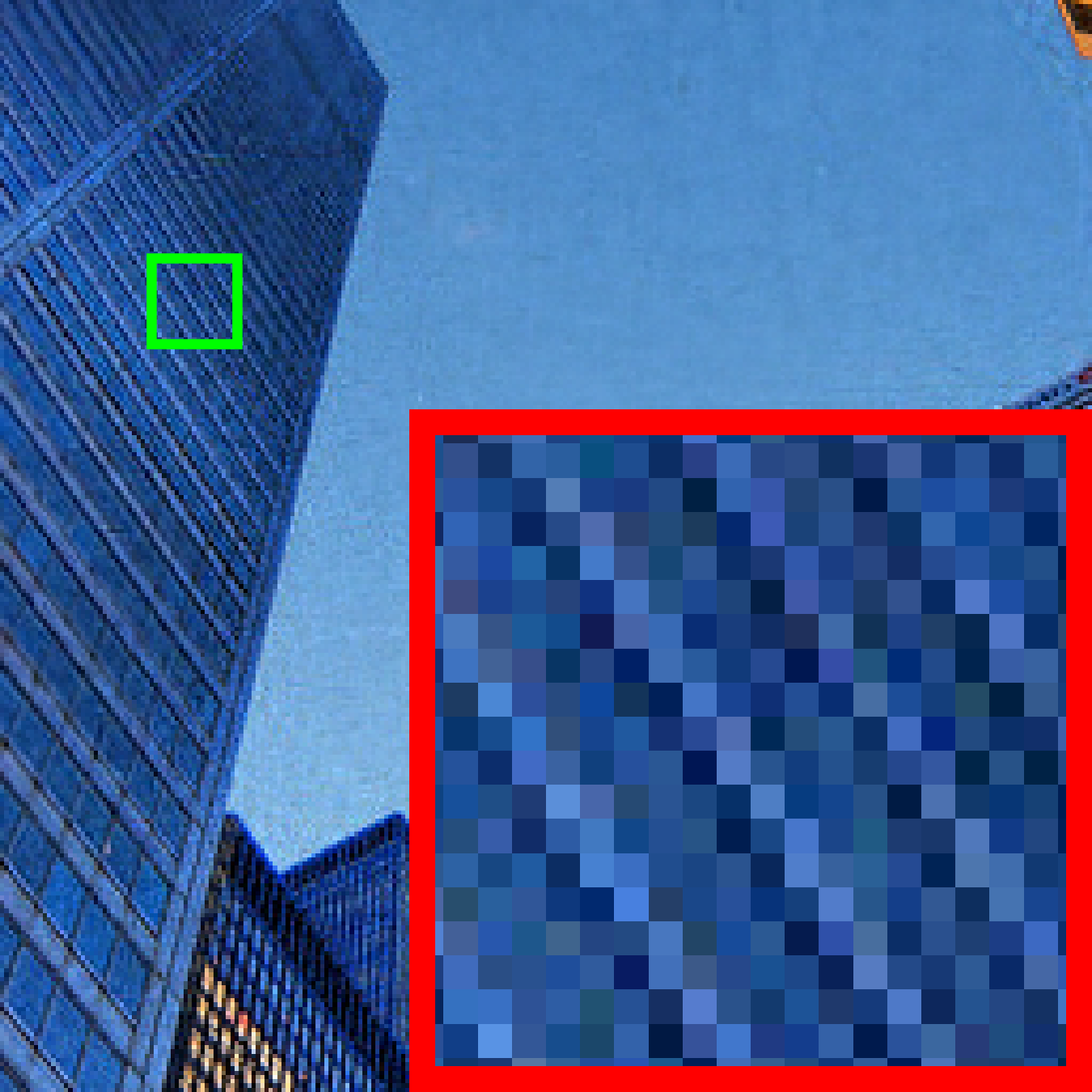}
&\includegraphics[width=0.09\textwidth]{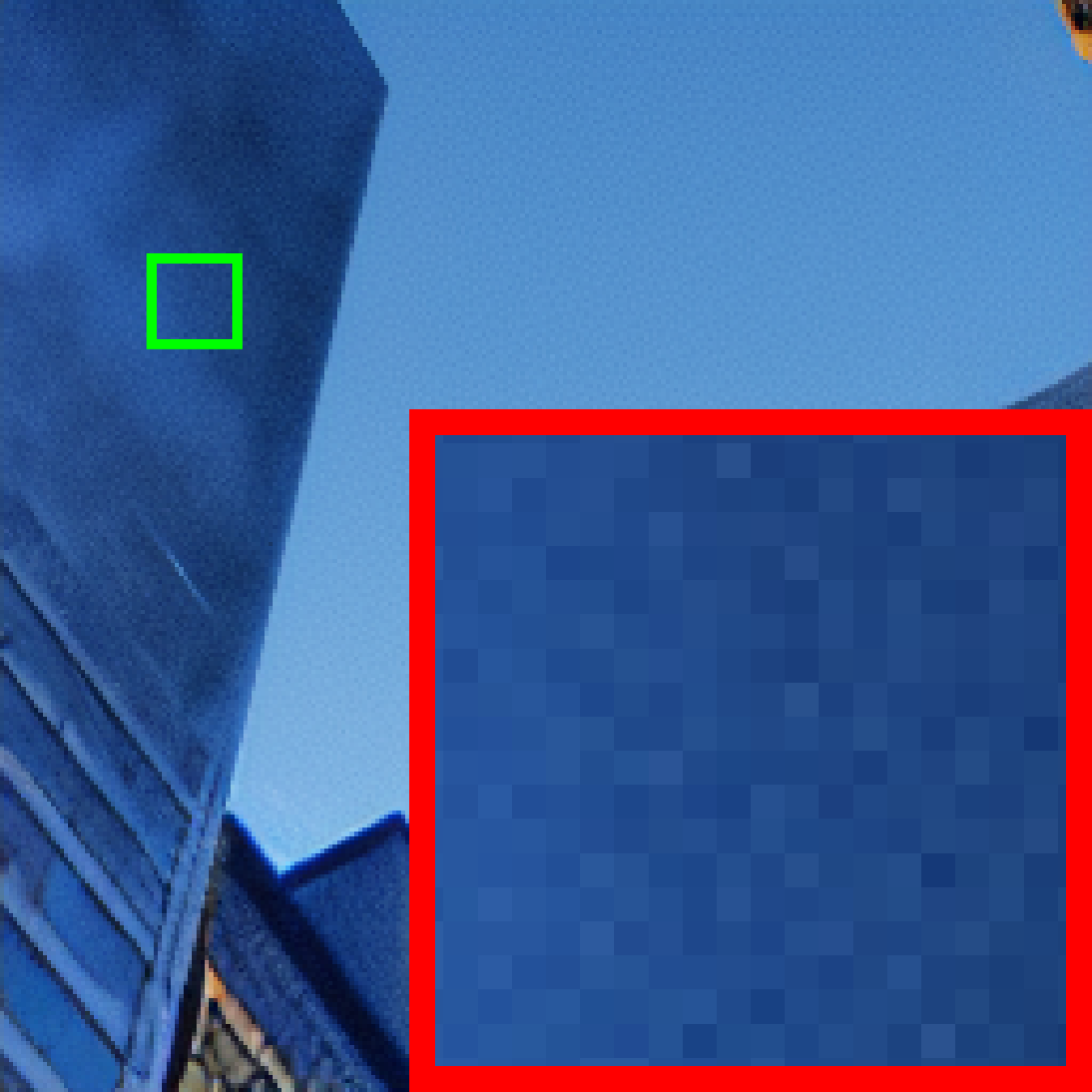}
&\includegraphics[width=0.09\textwidth]{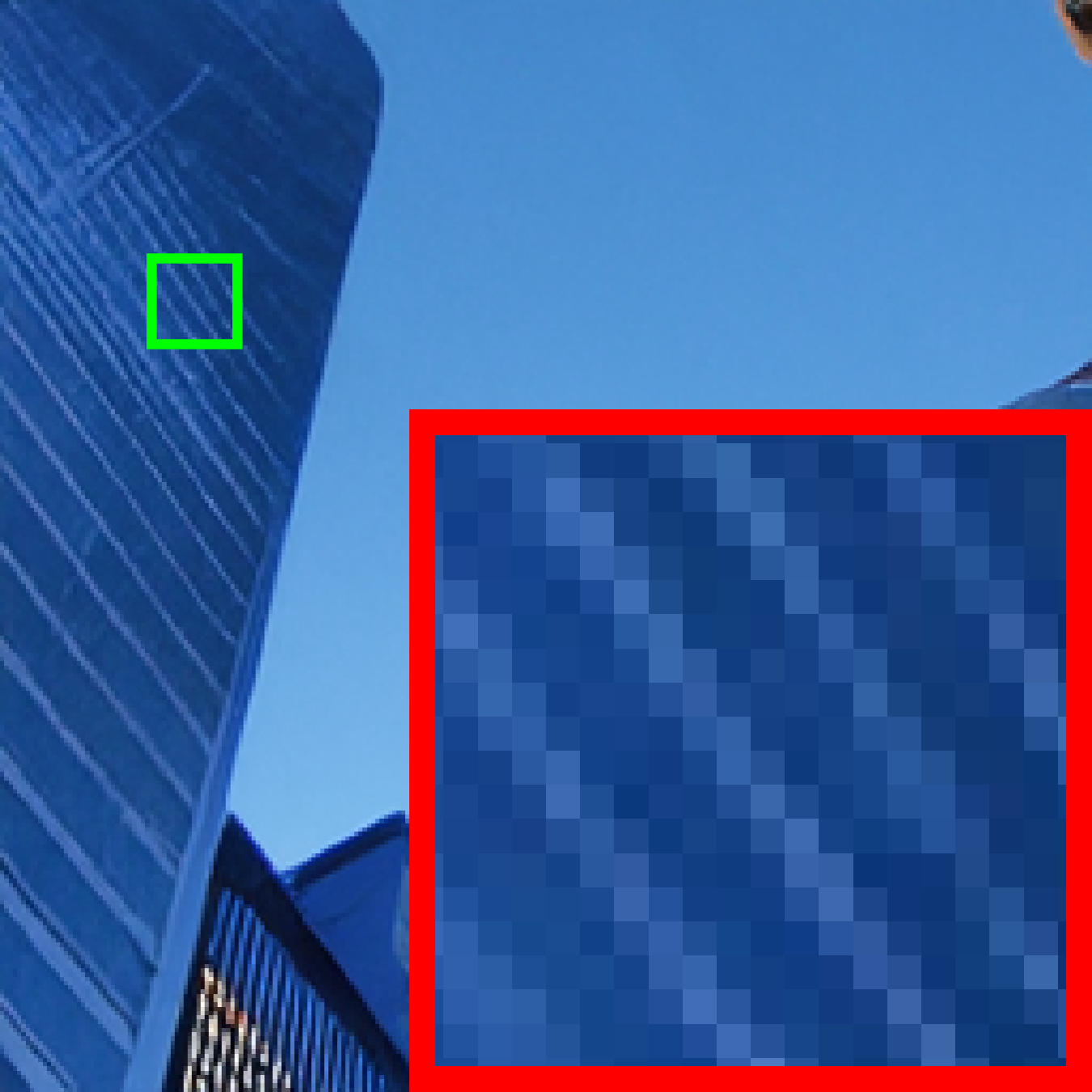}
&\includegraphics[width=0.09\textwidth]{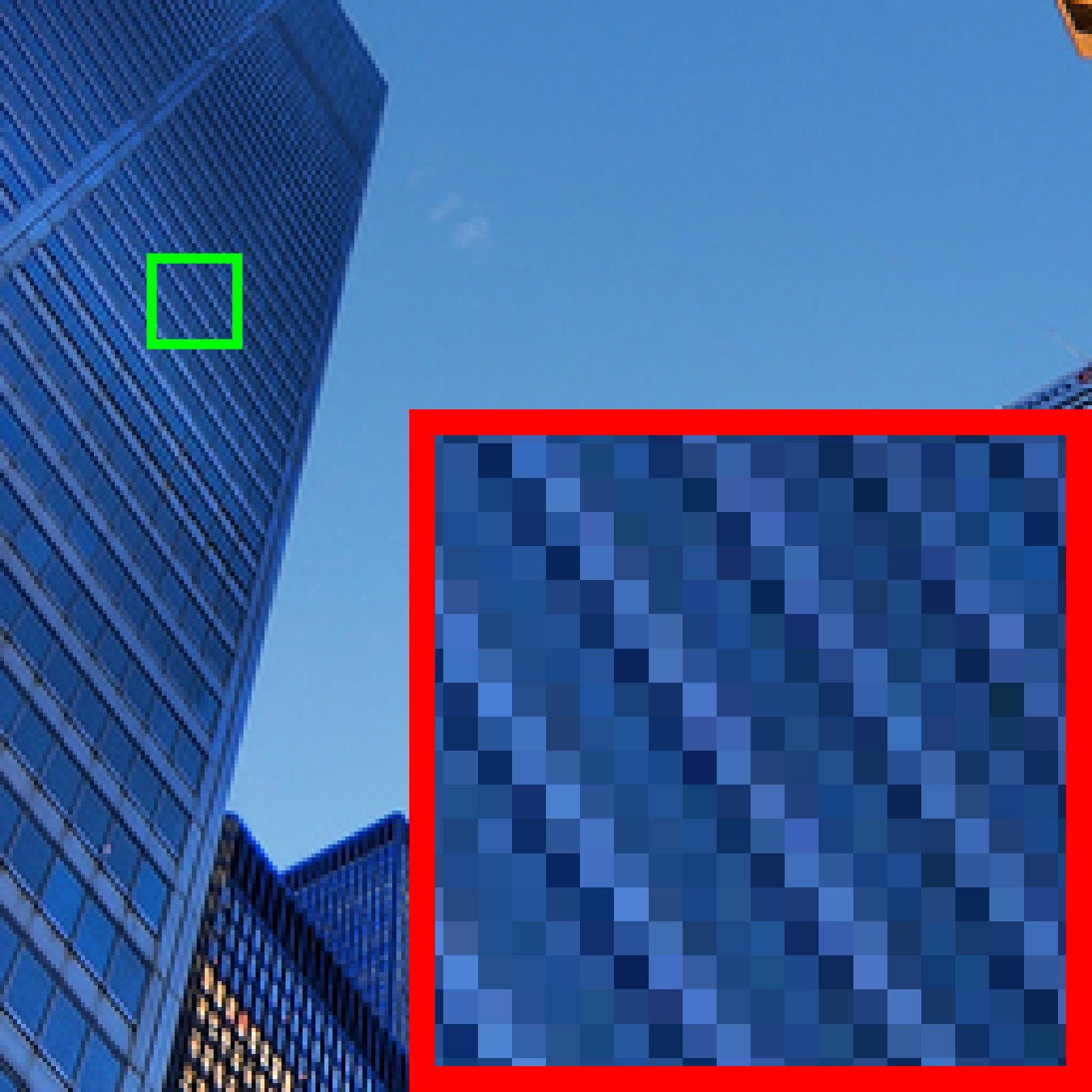}\\
\secondbest{37.41}/0.9319 & 30.30/0.8325 & 24.94/0.7217 & 25.27/0.8171 & \best{45.26}/\best{0.9882}
\end{tabular}}
\vspace{-10pt}
\caption{\textbf{Comparison of CS recovery results among various diffusion-based methods} on ``img\_012" image from Urban100 at $\gamma =50\%$.}
\label{fig:comp_diff_urban100}
\end{figure}

\begin{figure*}[!t]
\setlength{\tabcolsep}{0.5pt}\
\centering
\resizebox{1.0\textwidth}{!}{
\scriptsize
\begin{tabular}{ccccccc}
Ground Truth & ReconNet & ISTA-Net$^+$ & CSNet$^+$ & SCSNet & OPINE-Net$^+$ & COAST \\
\includegraphics[width=0.14\textwidth]{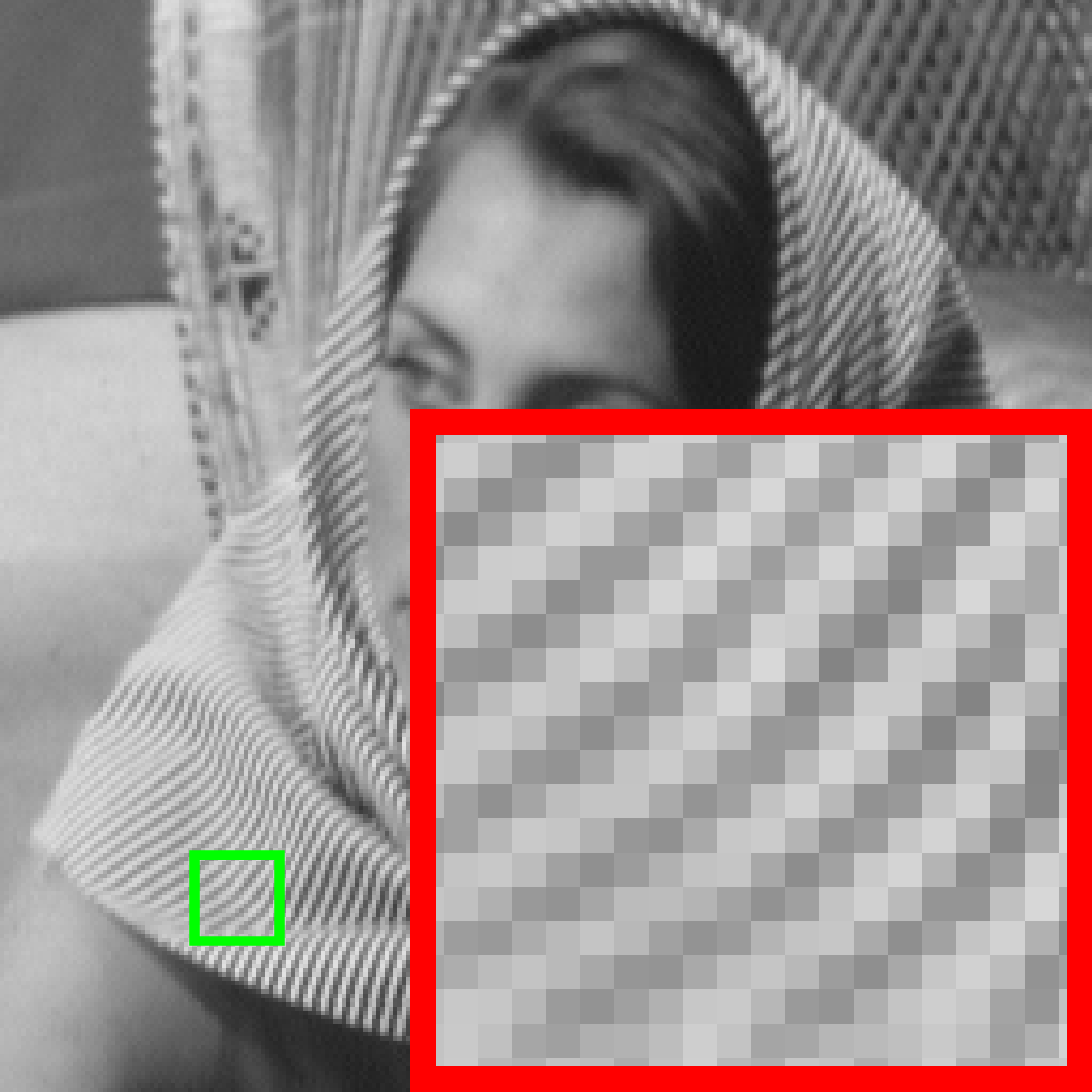}
&\includegraphics[width=0.14\textwidth]{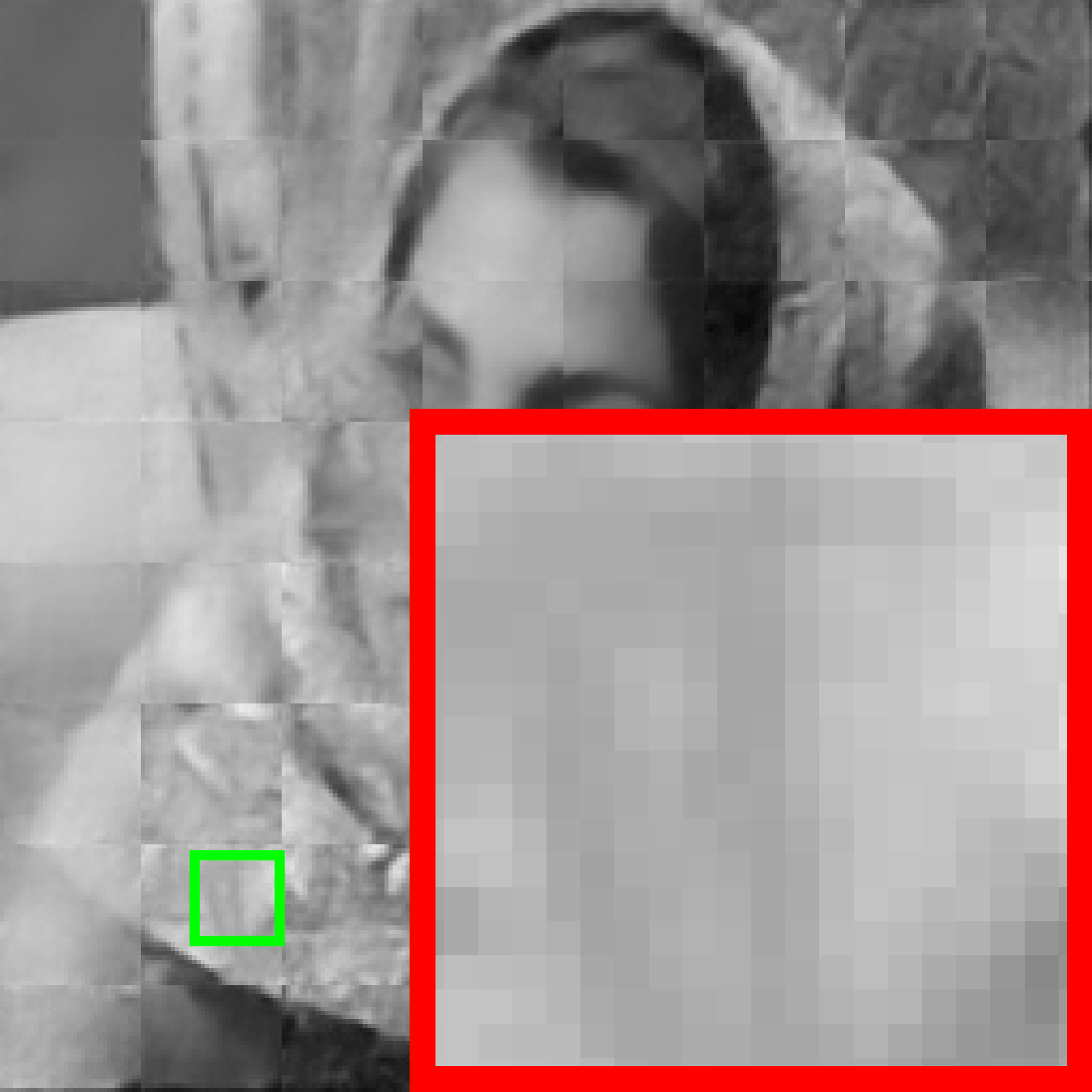}
&\includegraphics[width=0.14\textwidth]{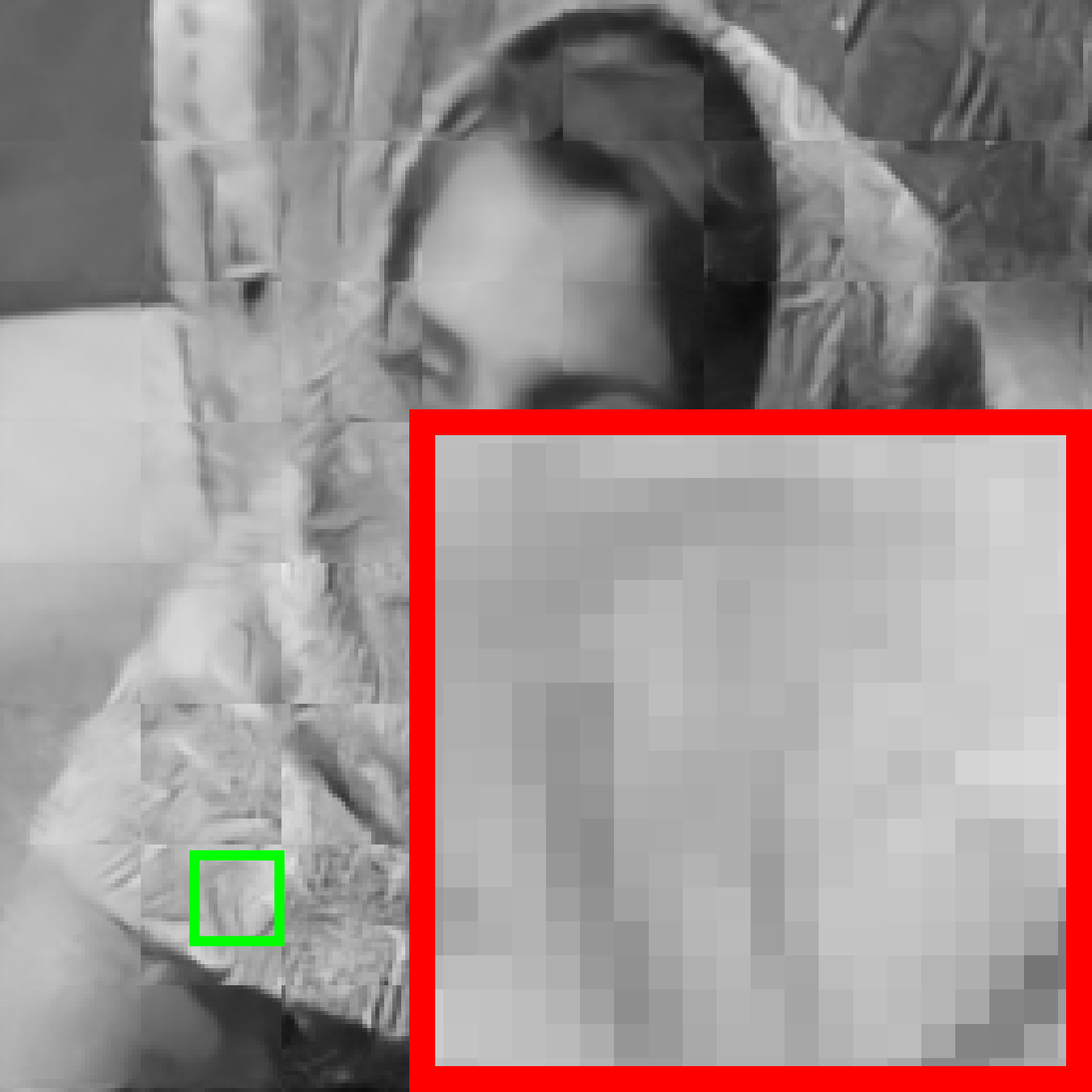}
&\includegraphics[width=0.14\textwidth]{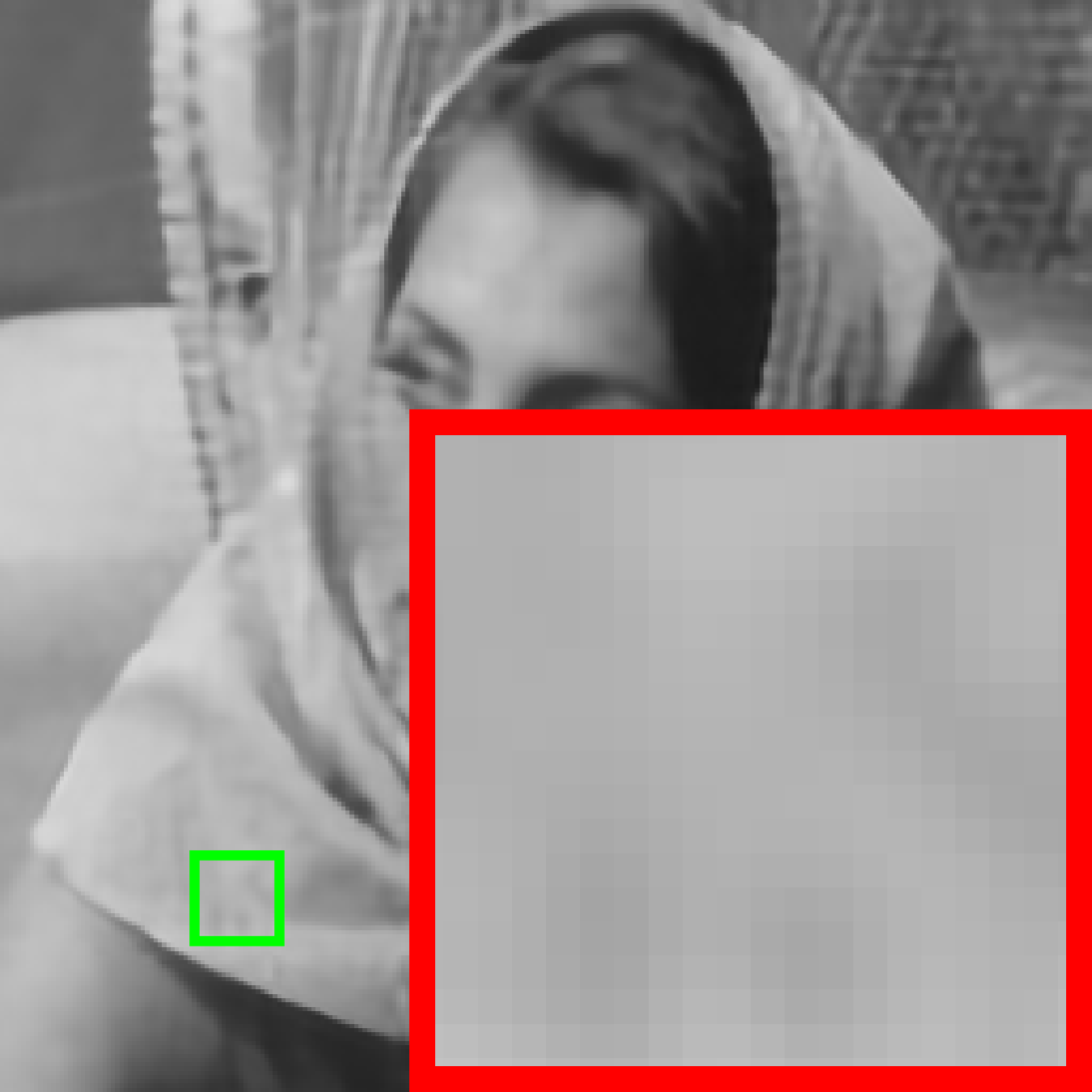}
&\includegraphics[width=0.14\textwidth]{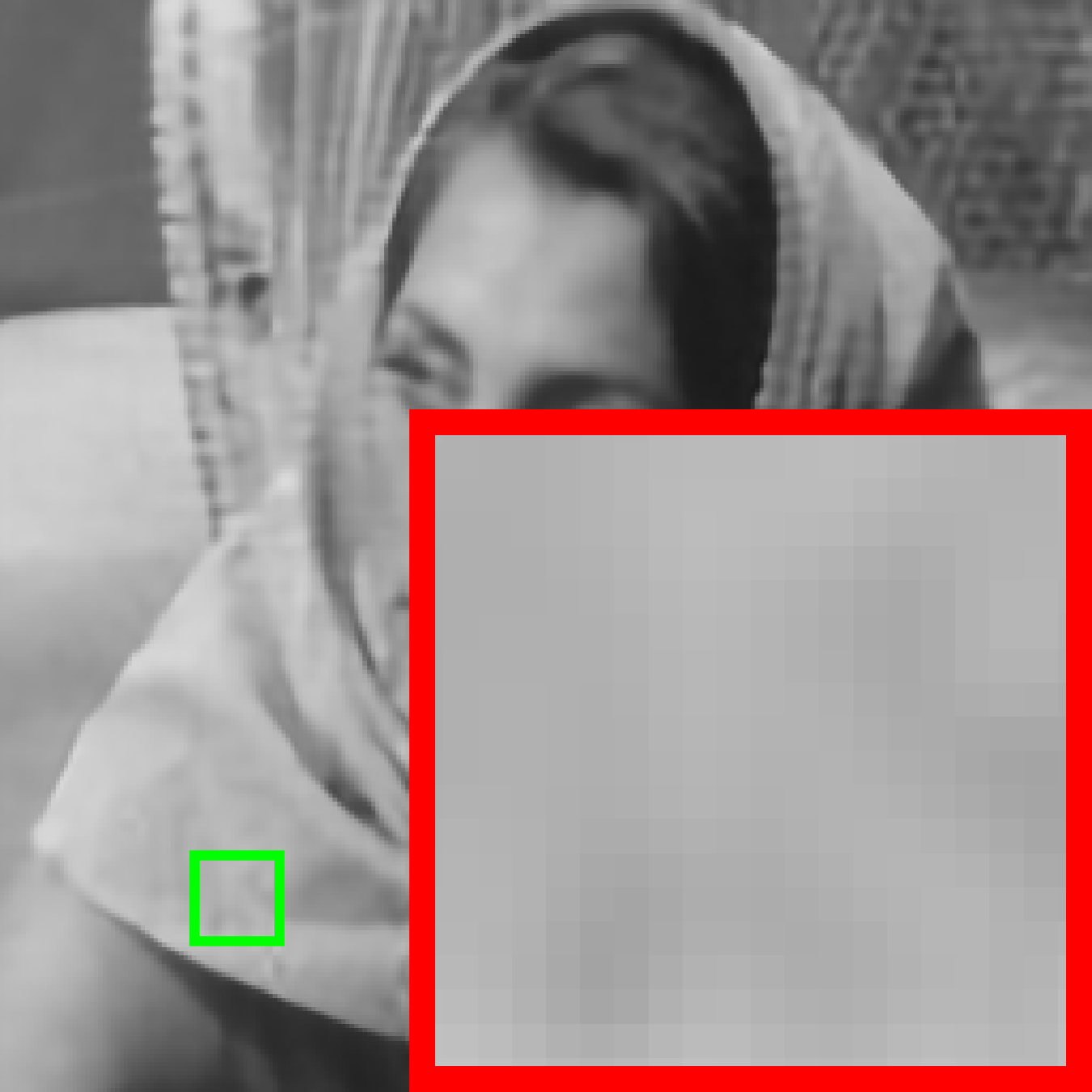}
&\includegraphics[width=0.14\textwidth]{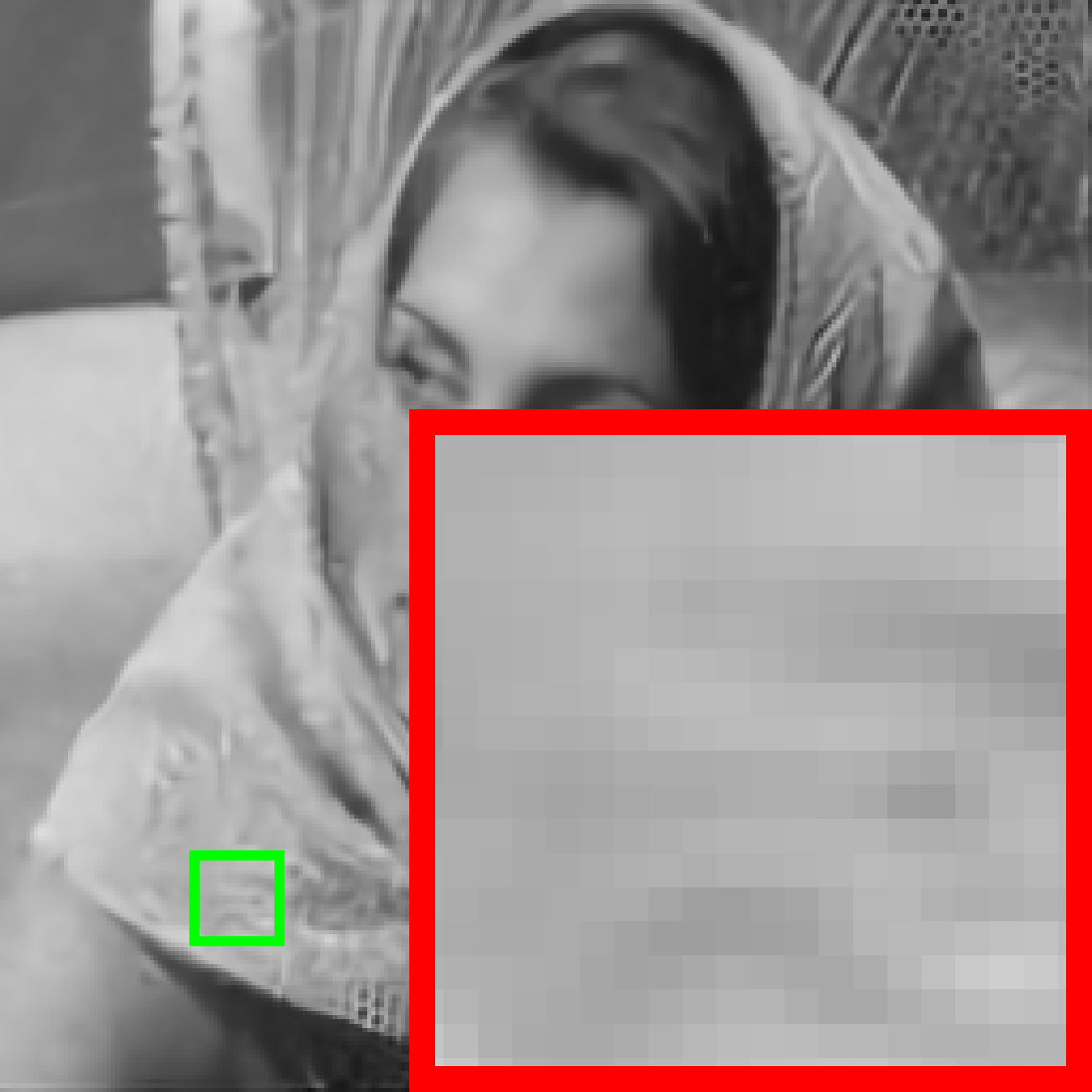}
&\includegraphics[width=0.14\textwidth]{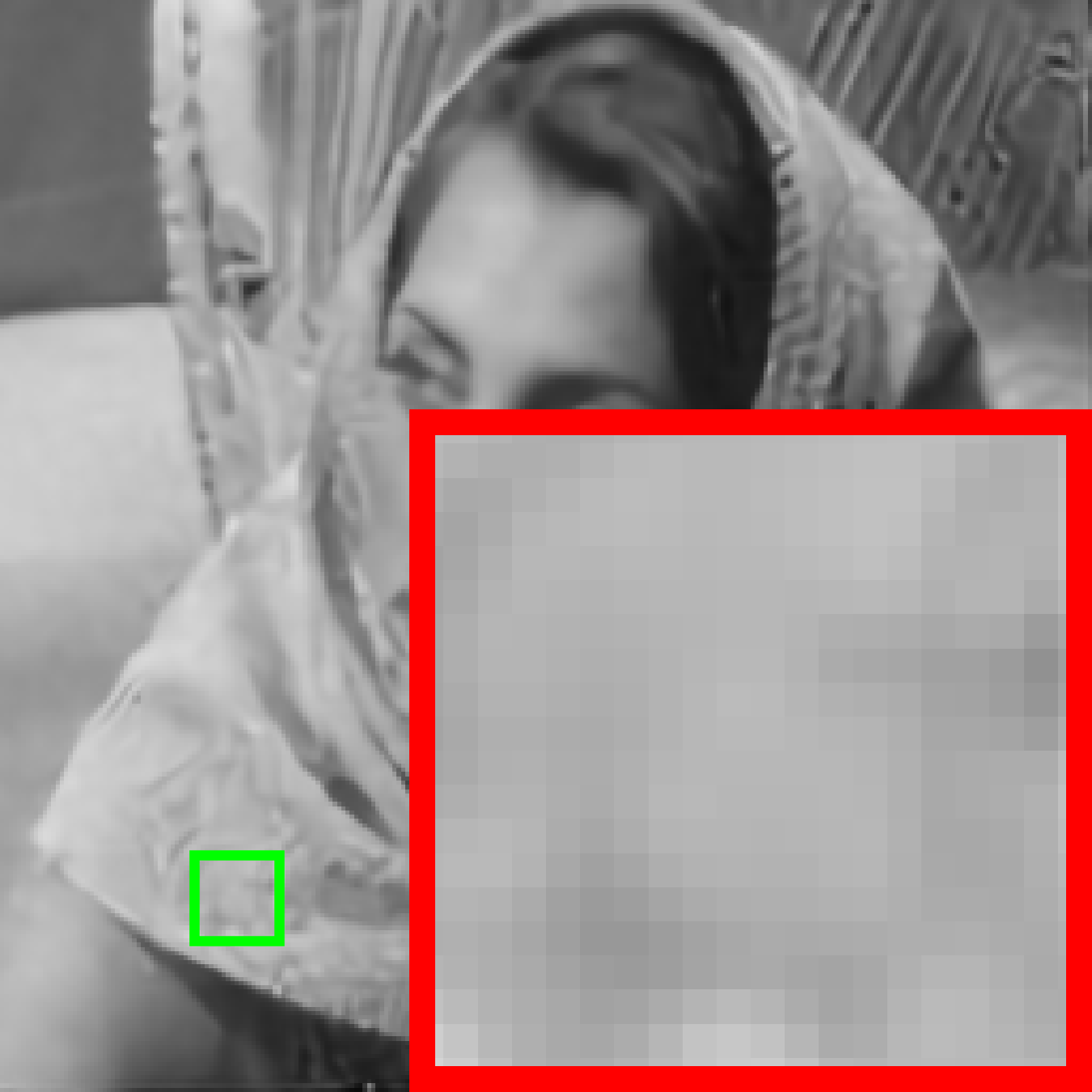}\\
PSNR/SSIM & 22.56/0.6260 & 23.52/0.6940 & 24.41/0.7249 & 24.43/0.7264 & 24.73/0.7674 & 24.61/0.7691
\end{tabular}}
\resizebox{1.0\textwidth}{!}{
\scriptsize
\begin{tabular}{cccccc>{\columncolor[HTML]{FFEEED}}c}
MADUN & DGUNet$^+$ & FSOINet & CASNet & TransCS & PRL-PGD$^+$ & \textbf{IDM (Ours)} \\
\includegraphics[width=0.14\textwidth]{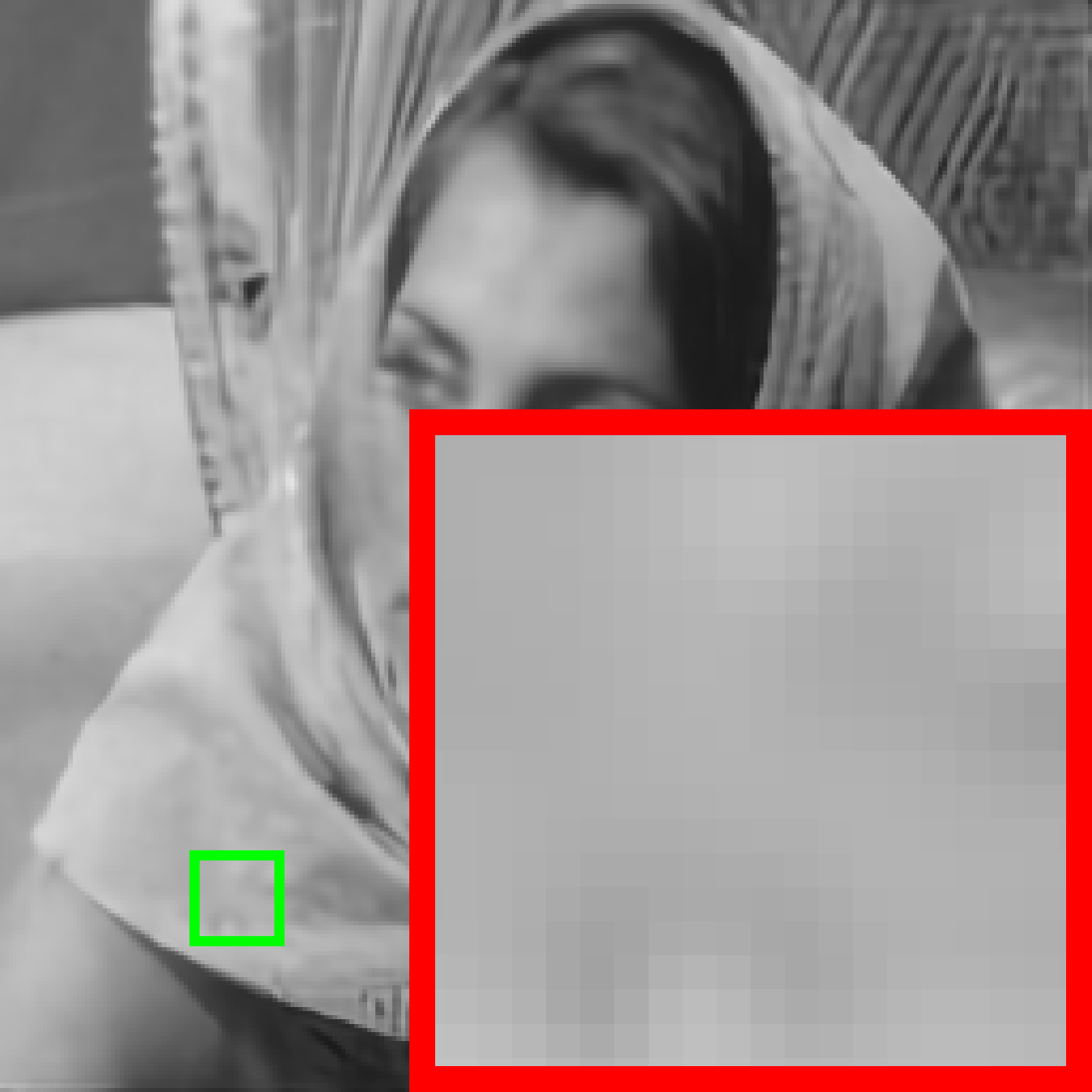}
&\includegraphics[width=0.14\textwidth]{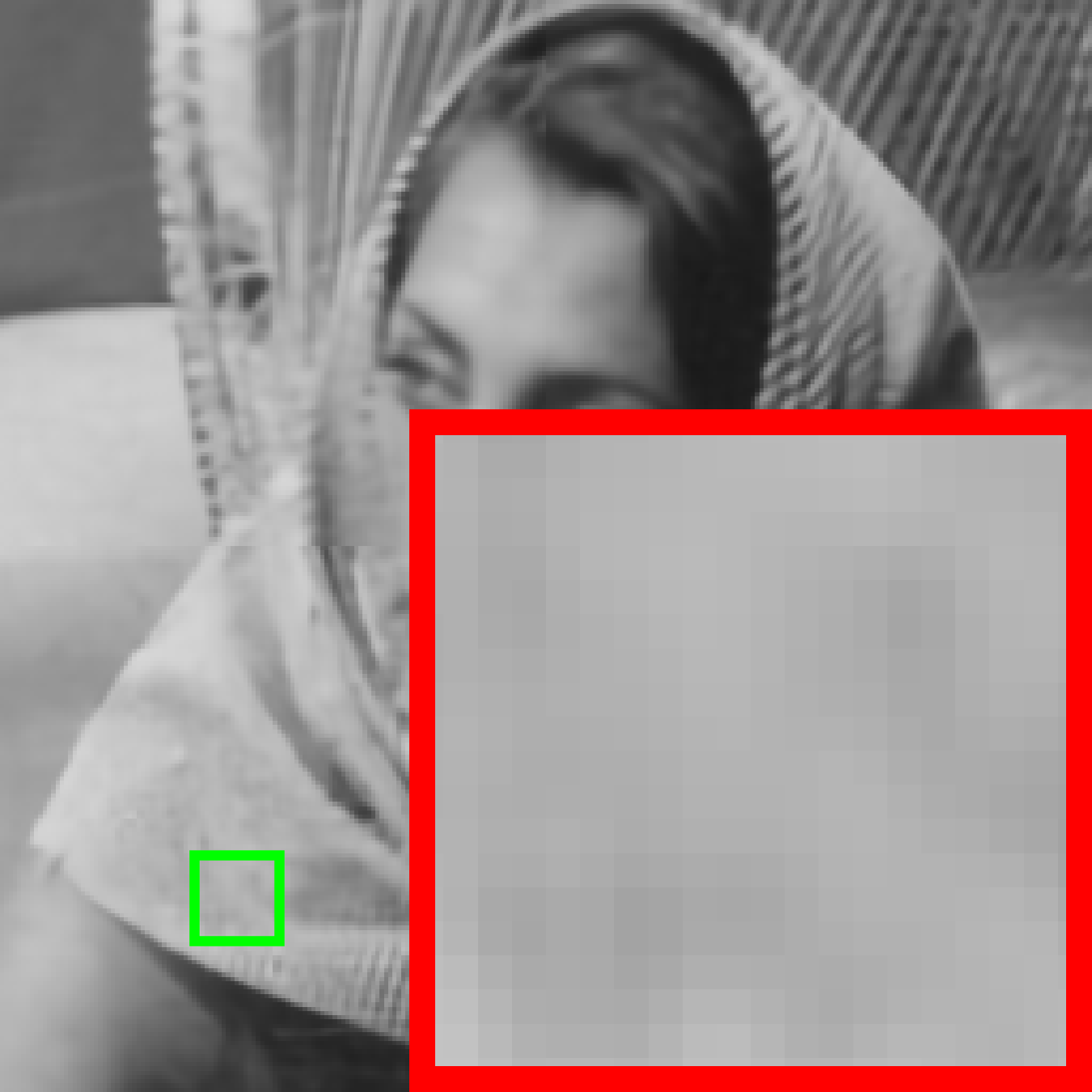}
&\includegraphics[width=0.14\textwidth]{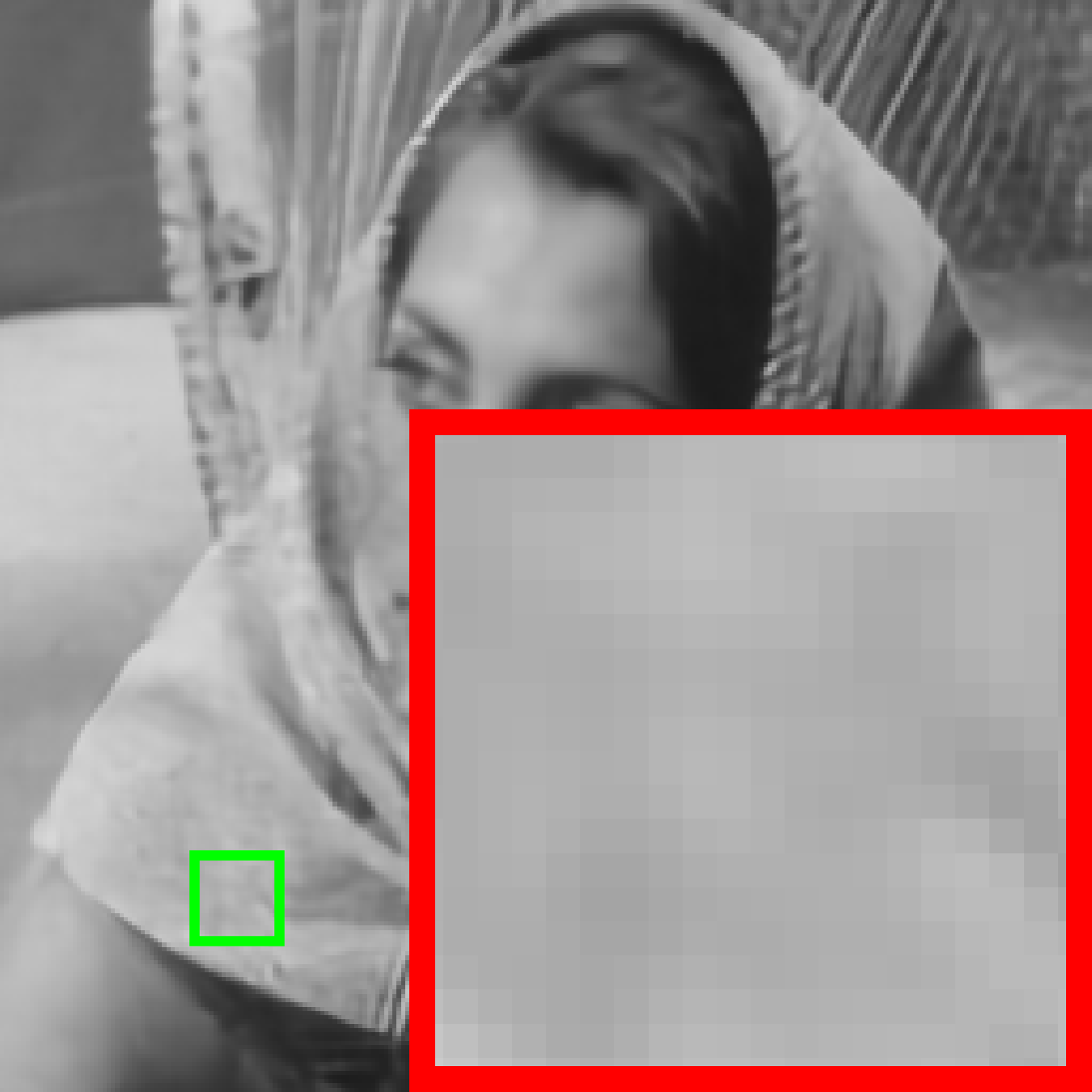}
&\includegraphics[width=0.14\textwidth]{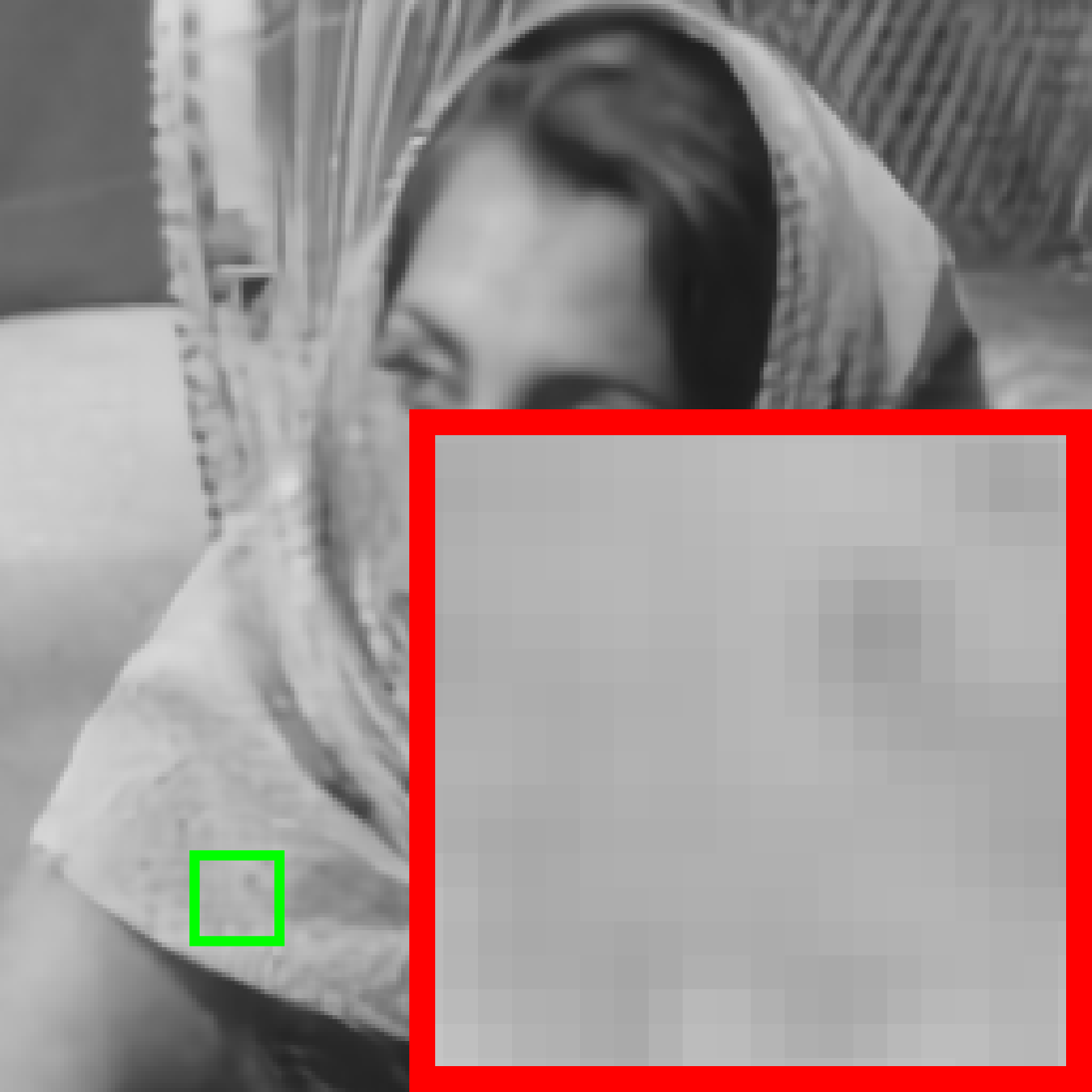}
&\includegraphics[width=0.14\textwidth]{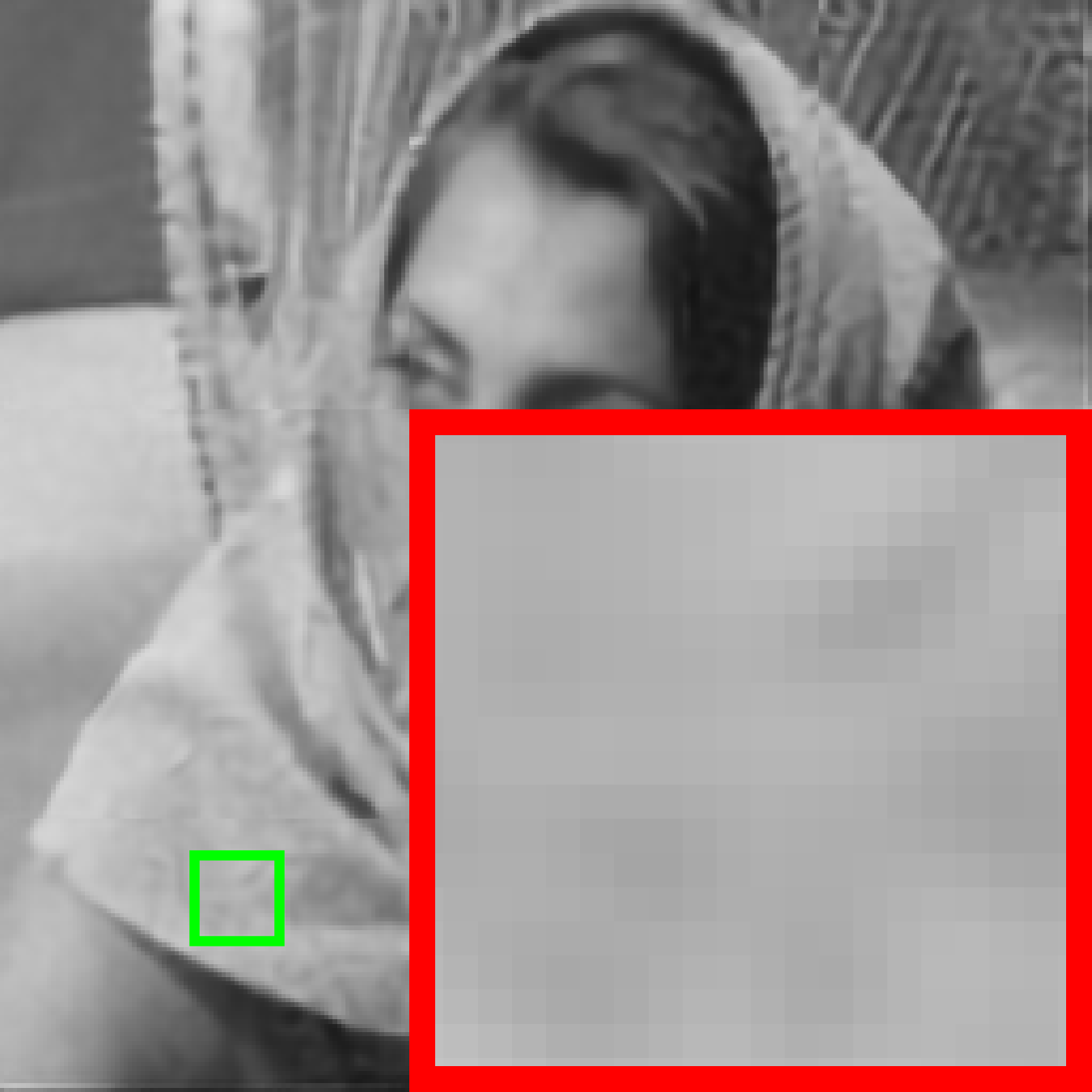}
&\includegraphics[width=0.14\textwidth]{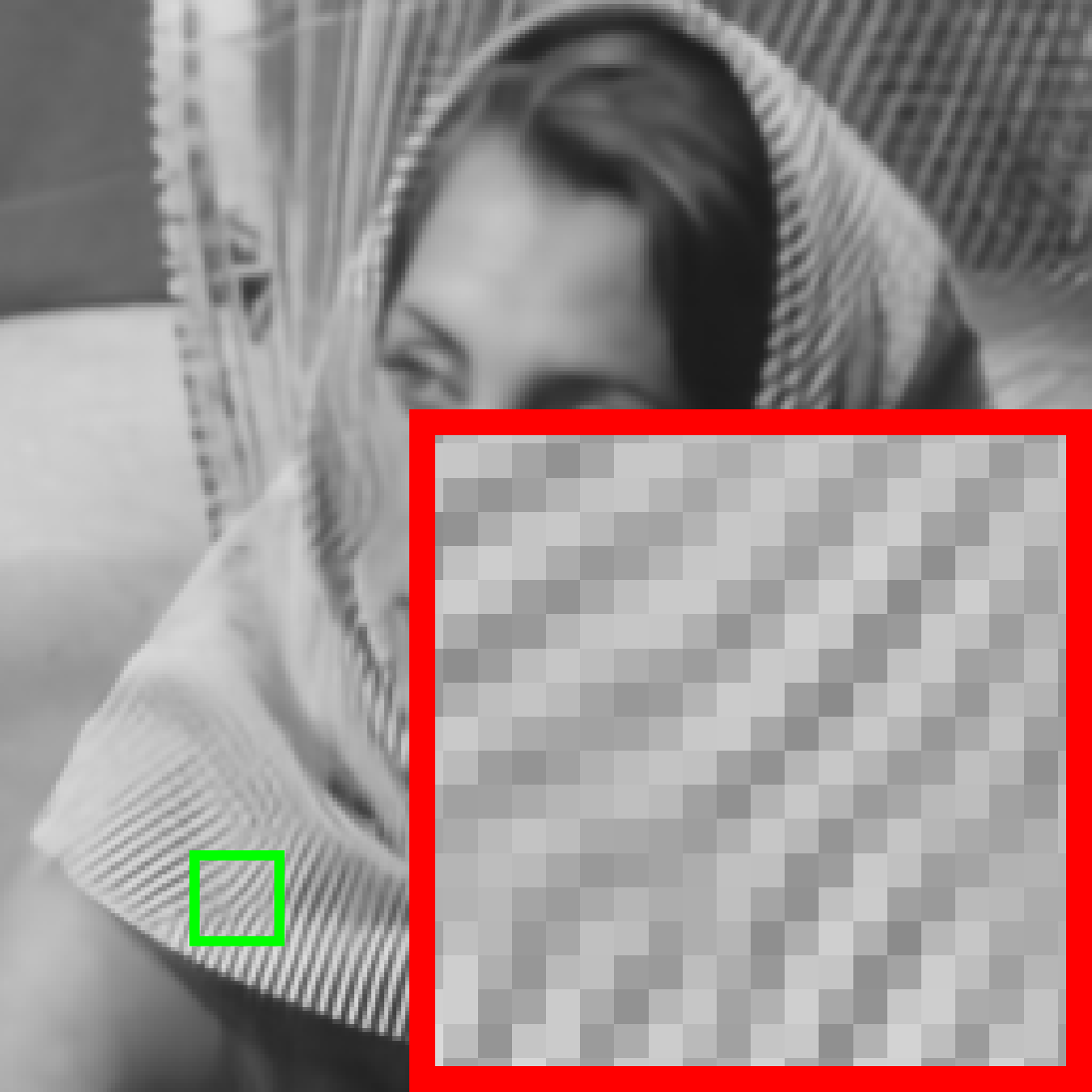}
&\includegraphics[width=0.14\textwidth]{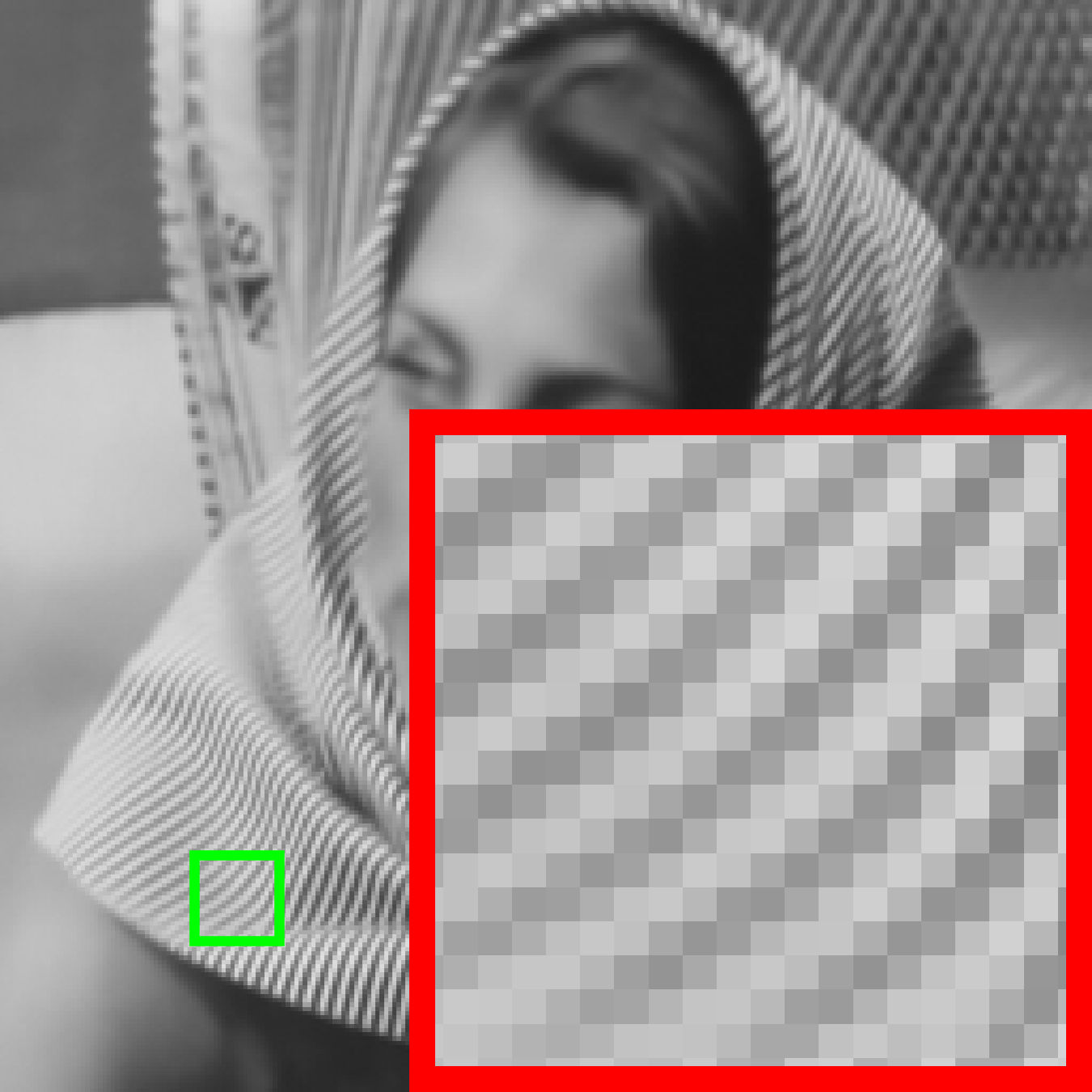}\\
25.14/0.7816 & 26.13/0.8145 & 25.25/0.7835 & 25.04/0.7809 & 24.92/0.7672 & \secondbest{28.44}/\secondbest{0.8747} & \best{32.78}/\best{0.9379}
\end{tabular}}
\resizebox{1.0\textwidth}{!}{
\scriptsize
\begin{tabular}{ccccccc}
Ground Truth & ReconNet & ISTA-Net$^+$ & CSNet$^+$ & SCSNet & OPINE-Net$^+$ & COAST \\
\includegraphics[width=0.14\textwidth]{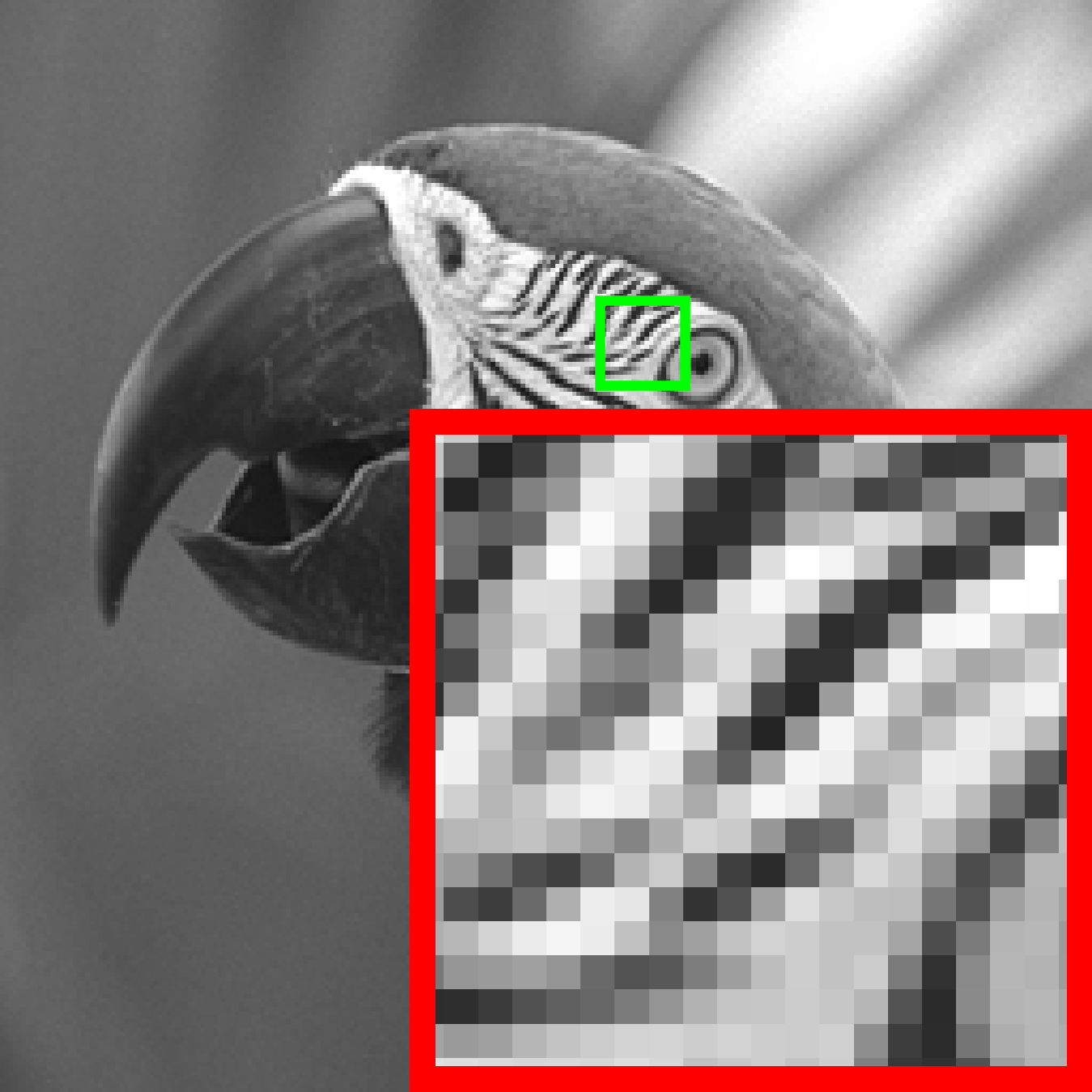}
&\includegraphics[width=0.14\textwidth]{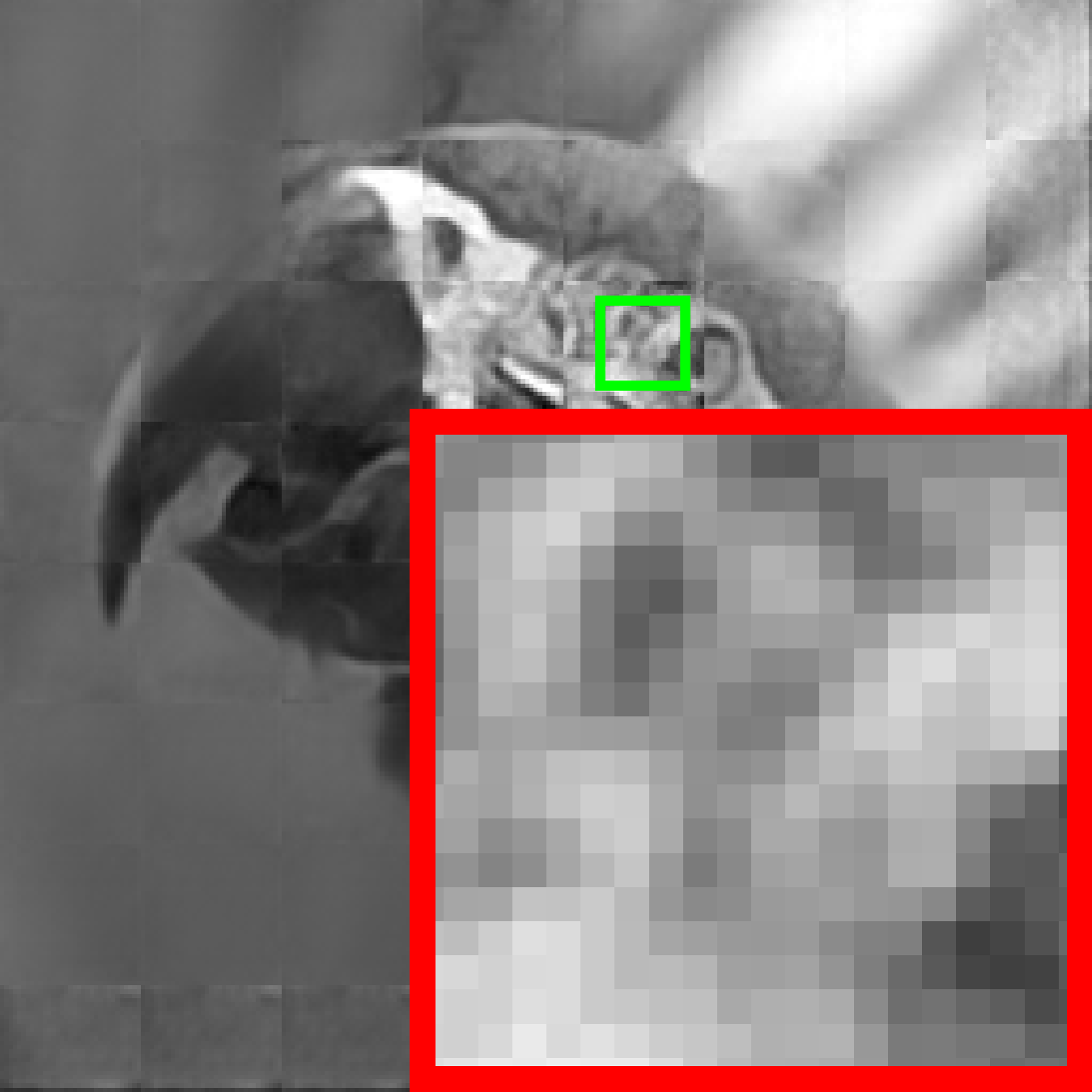}
&\includegraphics[width=0.14\textwidth]{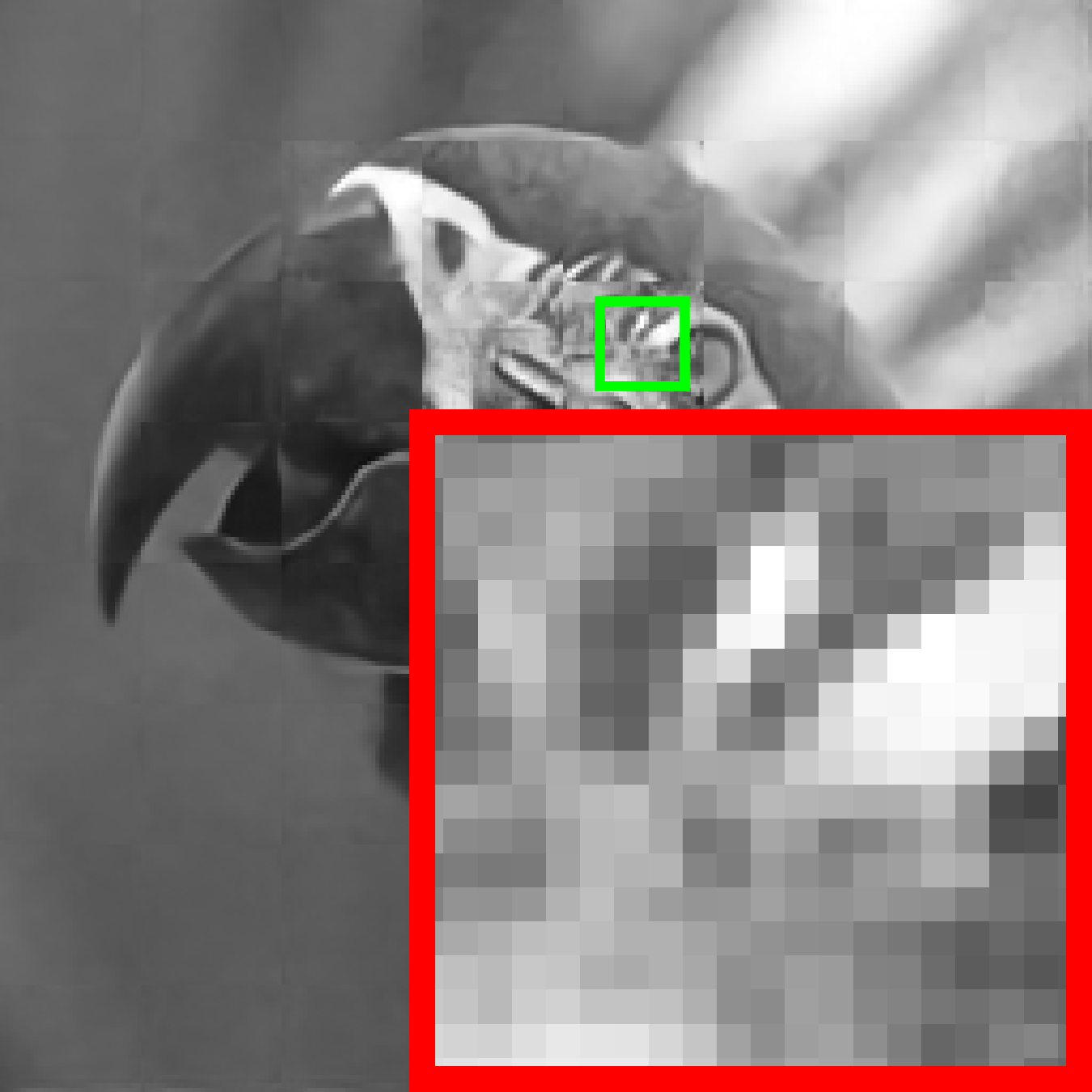}
&\includegraphics[width=0.14\textwidth]{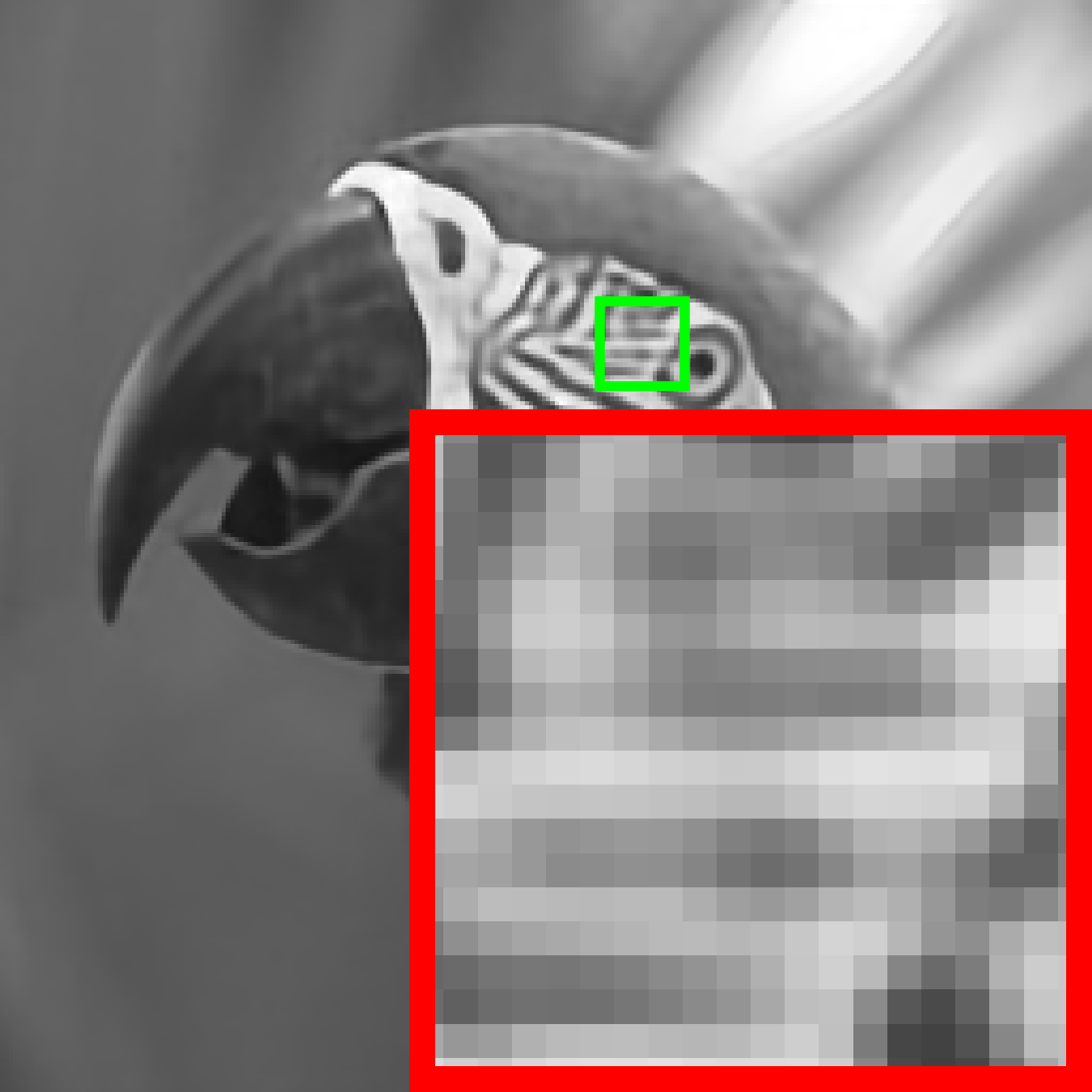}
&\includegraphics[width=0.14\textwidth]{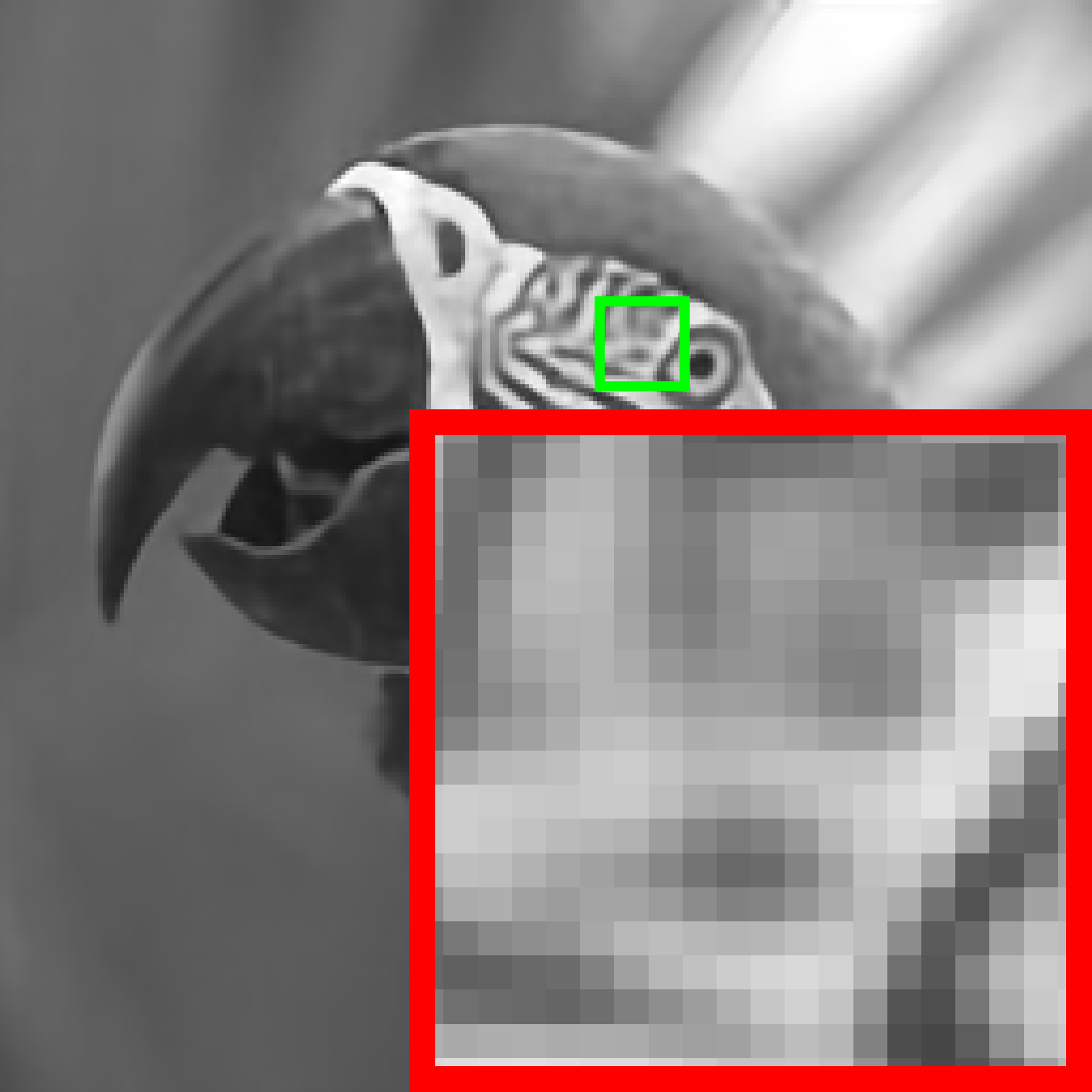}
&\includegraphics[width=0.14\textwidth]{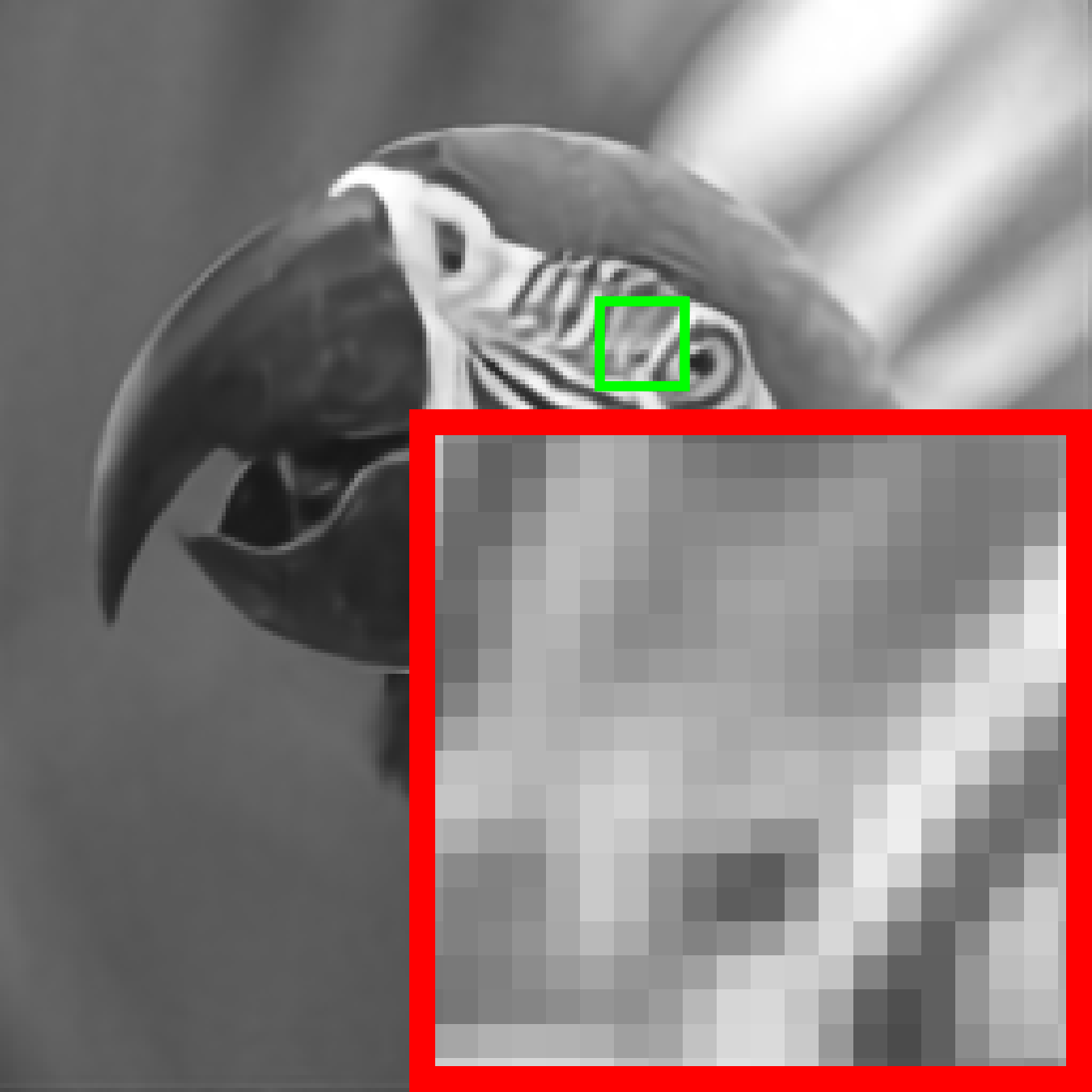}
&\includegraphics[width=0.14\textwidth]{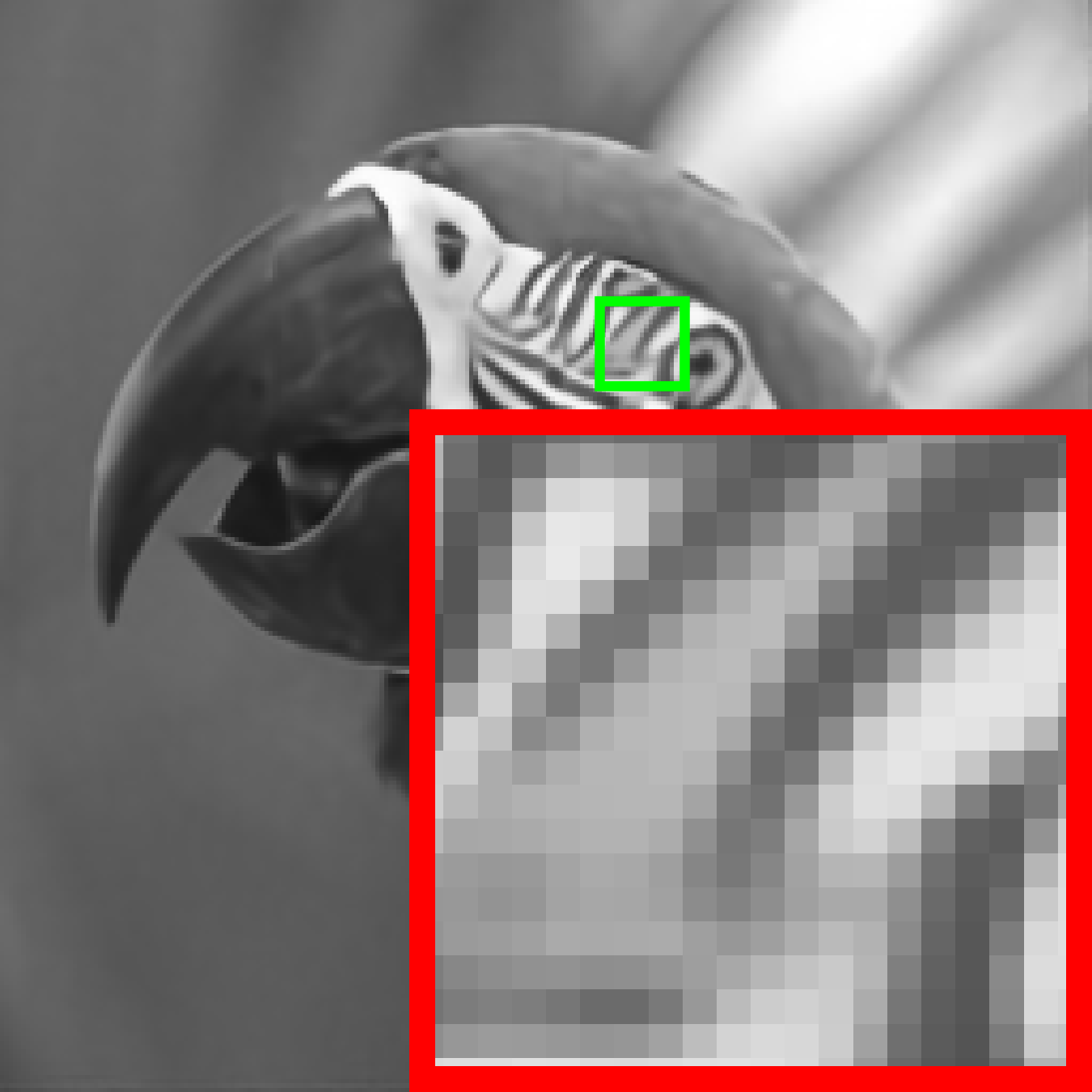}\\
PSNR/SSIM & 24.21/0.7660 & 26.37/0.8523 & 28.11/0.8899 & 28.10/0.8921 & 29.34/0.9155 & 29.30/0.9206
\end{tabular}}
\resizebox{1.0\textwidth}{!}{
\scriptsize
\begin{tabular}{cccccc>{\columncolor[HTML]{FFEEED}}c}
MADUN & DGUNet$^+$ & FSOINet & CASNet & TransCS & PRL-PGD$^+$ & \textbf{IDM (Ours)} \\
\includegraphics[width=0.14\textwidth]{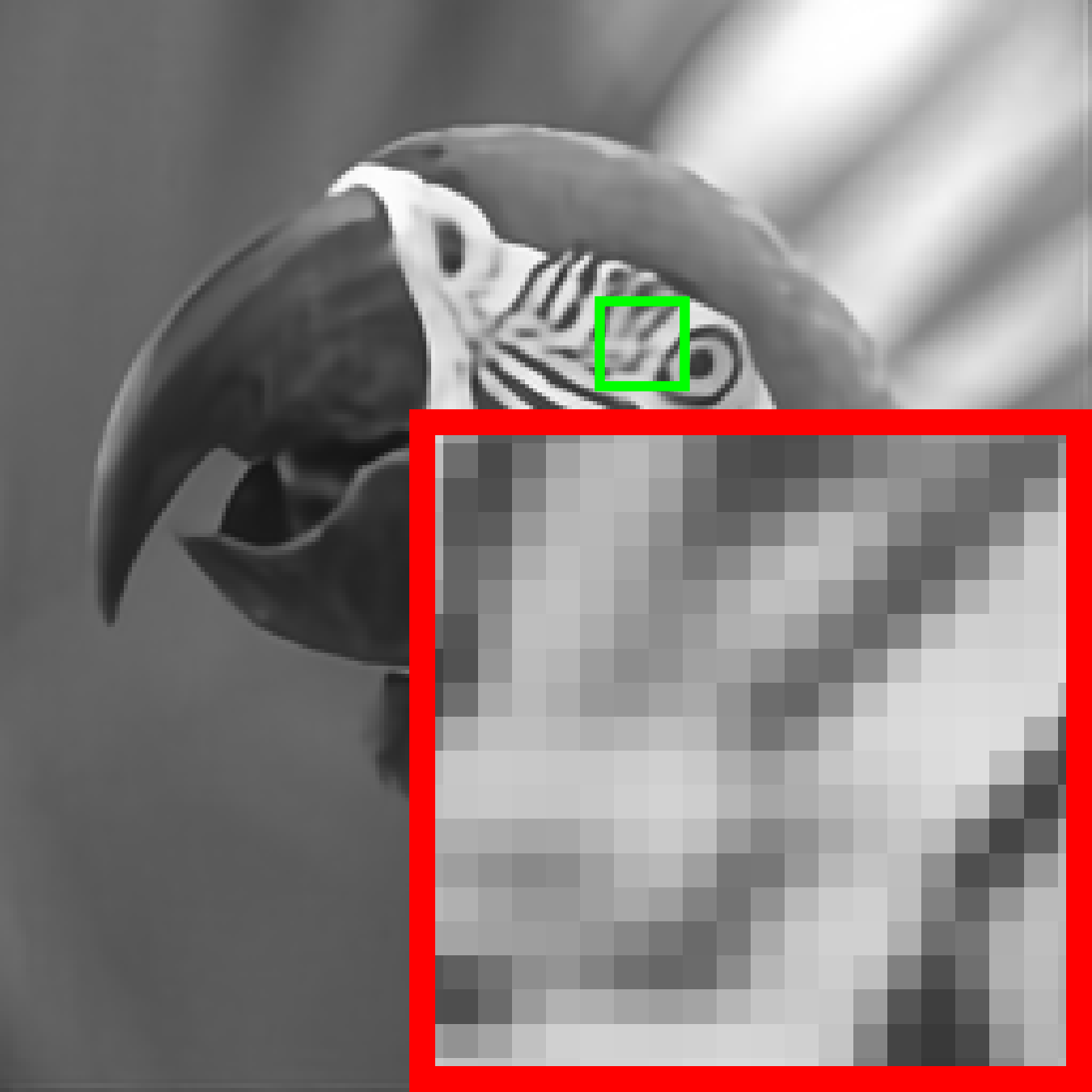}
&\includegraphics[width=0.14\textwidth]{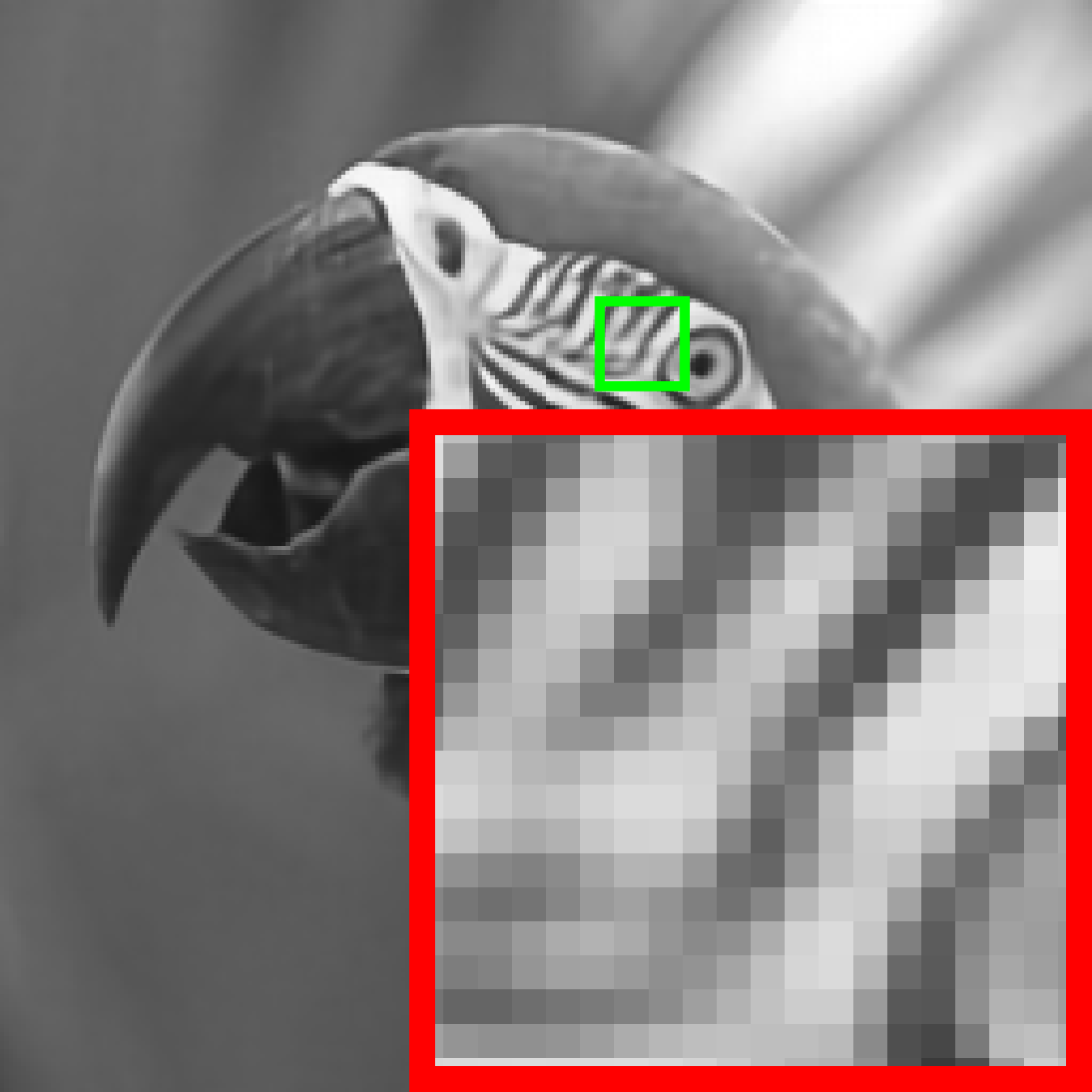}
&\includegraphics[width=0.14\textwidth]{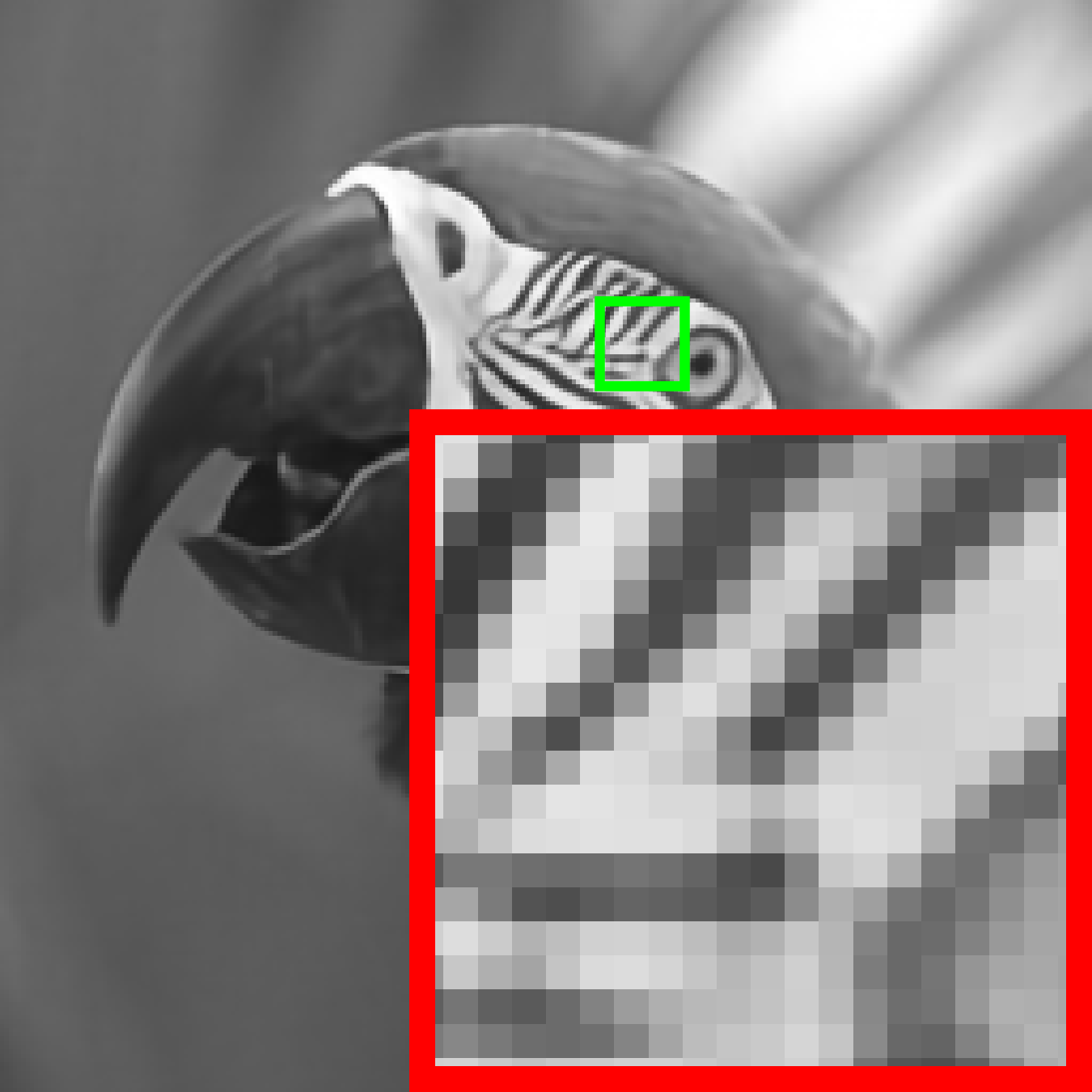}
&\includegraphics[width=0.14\textwidth]{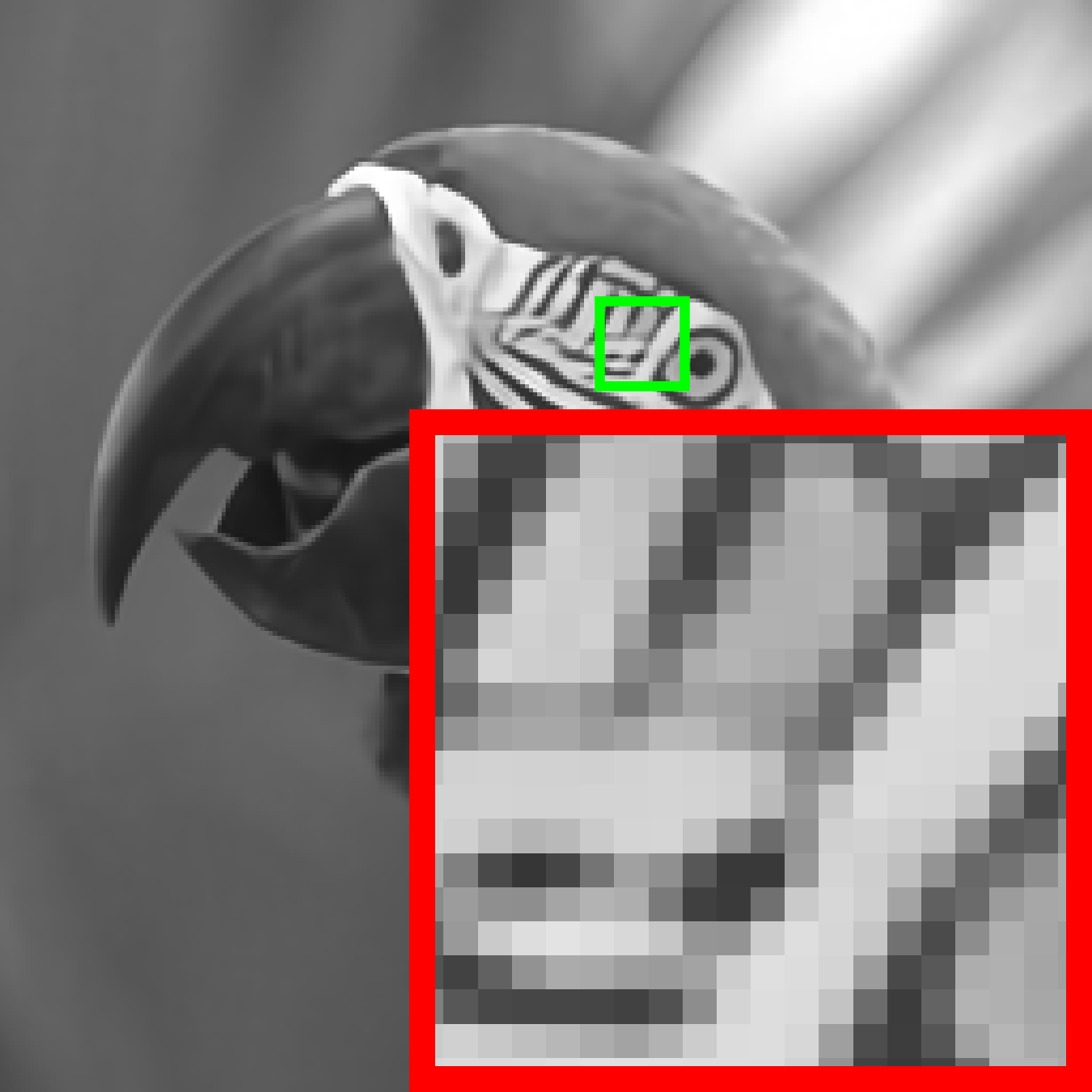}
&\includegraphics[width=0.14\textwidth]{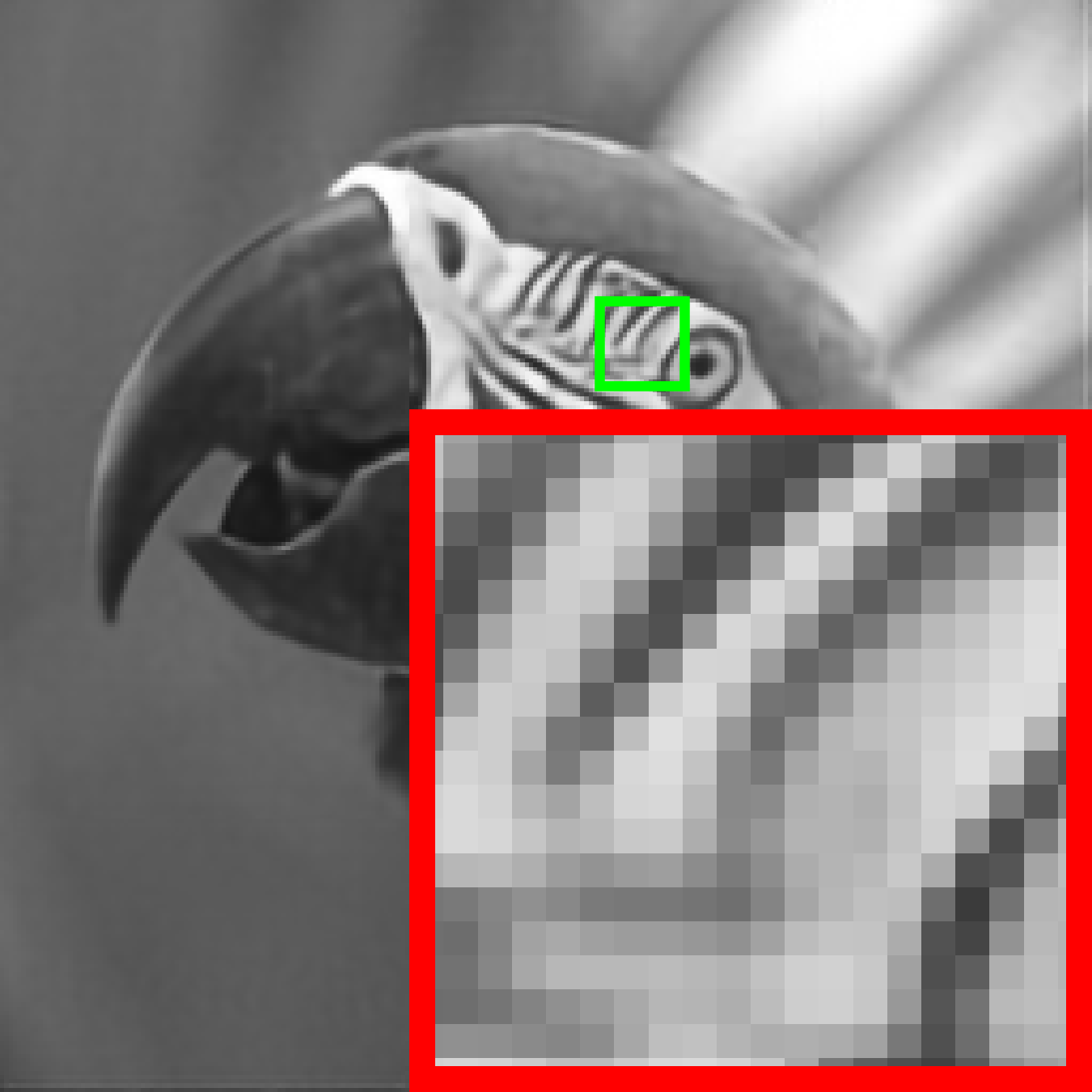}
&\includegraphics[width=0.14\textwidth]{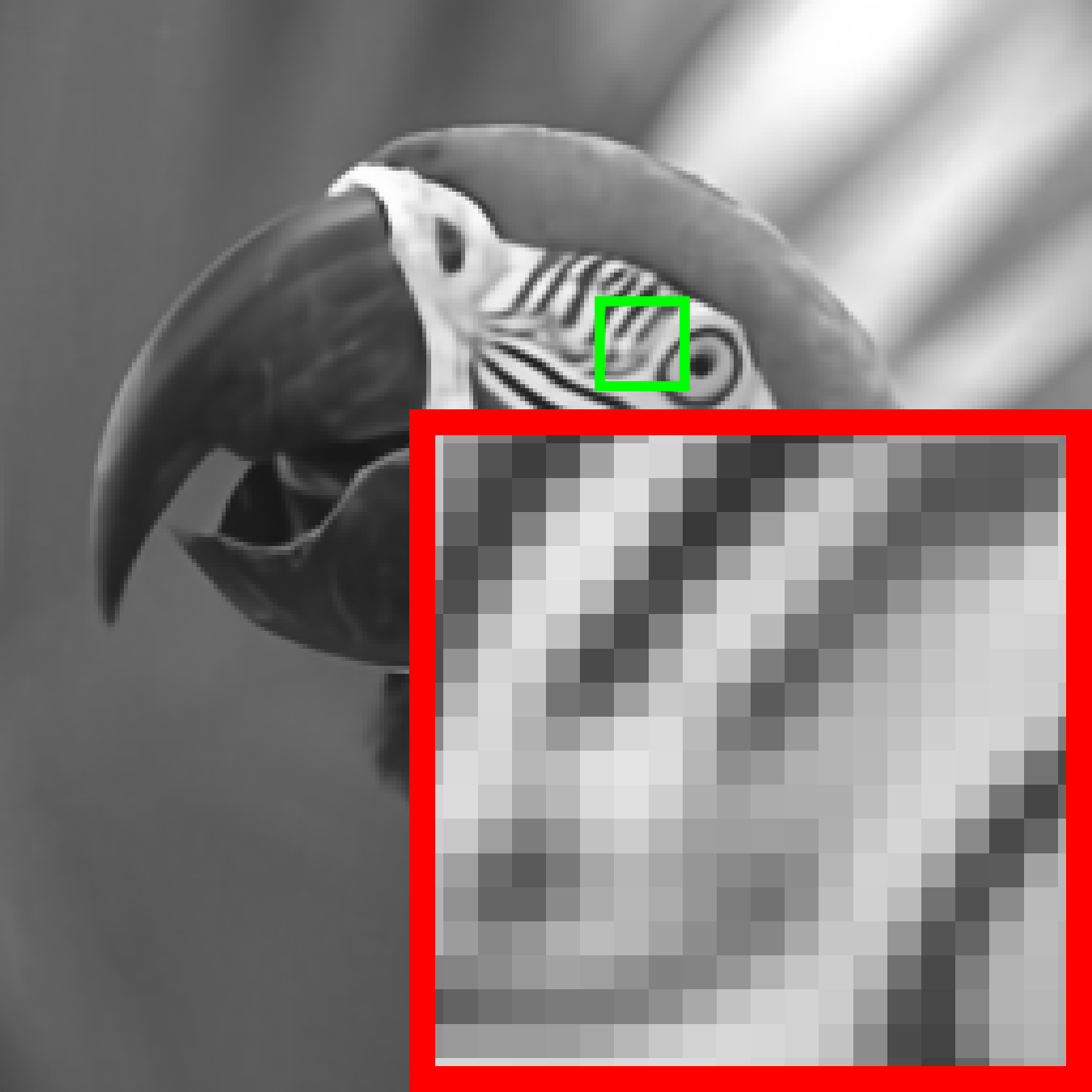}
&\includegraphics[width=0.14\textwidth]{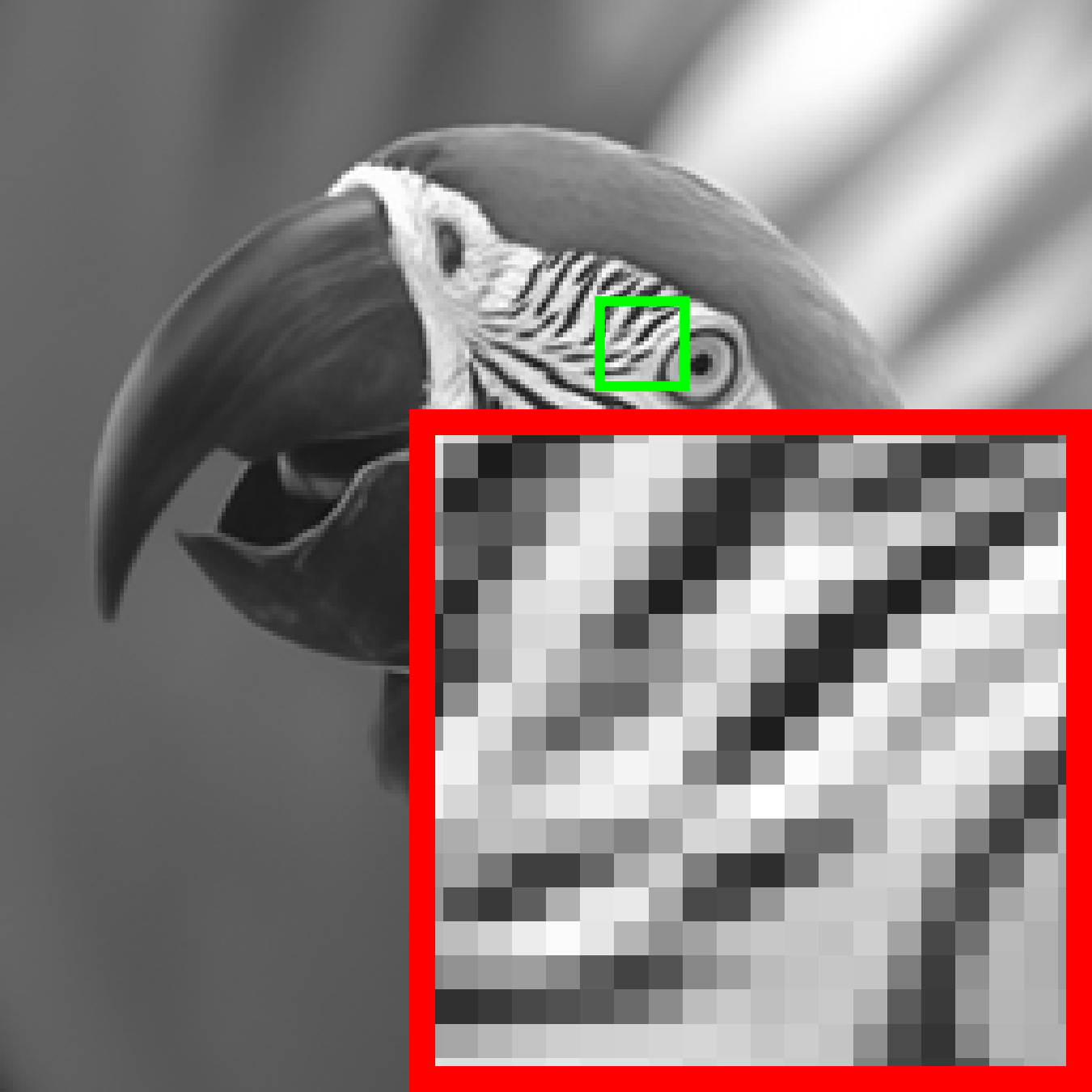}\\
29.46/0.9207 & 29.75/0.9226 & 29.81/0.9253 & 29.71/0.9253 & 29.44/0.9137 & \secondbest{30.81}/\secondbest{0.9303} & \best{34.85}/\best{0.9338}
\end{tabular}}
\vspace{-10pt}
\caption{\textbf{Comparison of CS reconstruction results} on two images named ``barbara'' \textcolor{blue}{\textbf{(top)}} and ``Parrots'' \textcolor{blue}{\textbf{(bottom)}} from Set11 at CS ratio $\gamma =10\%$.}
\label{fig:comp_set11}
\end{figure*}

\begin{figure*}[!t]
\setlength{\tabcolsep}{0.5pt}
\centering
\resizebox{1.0\textwidth}{!}{
\scriptsize
\begin{tabular}{ccccccccc>{\columncolor[HTML]{FFEEED}}c}
Ground Truth & $\A^\dagger \y$ & DDRM & $\Pi$GDM & DPS & DDNM & GDP & PSLD & SR3 & \textbf{IDM (Ours)}\\
\includegraphics[width=0.09\textwidth]{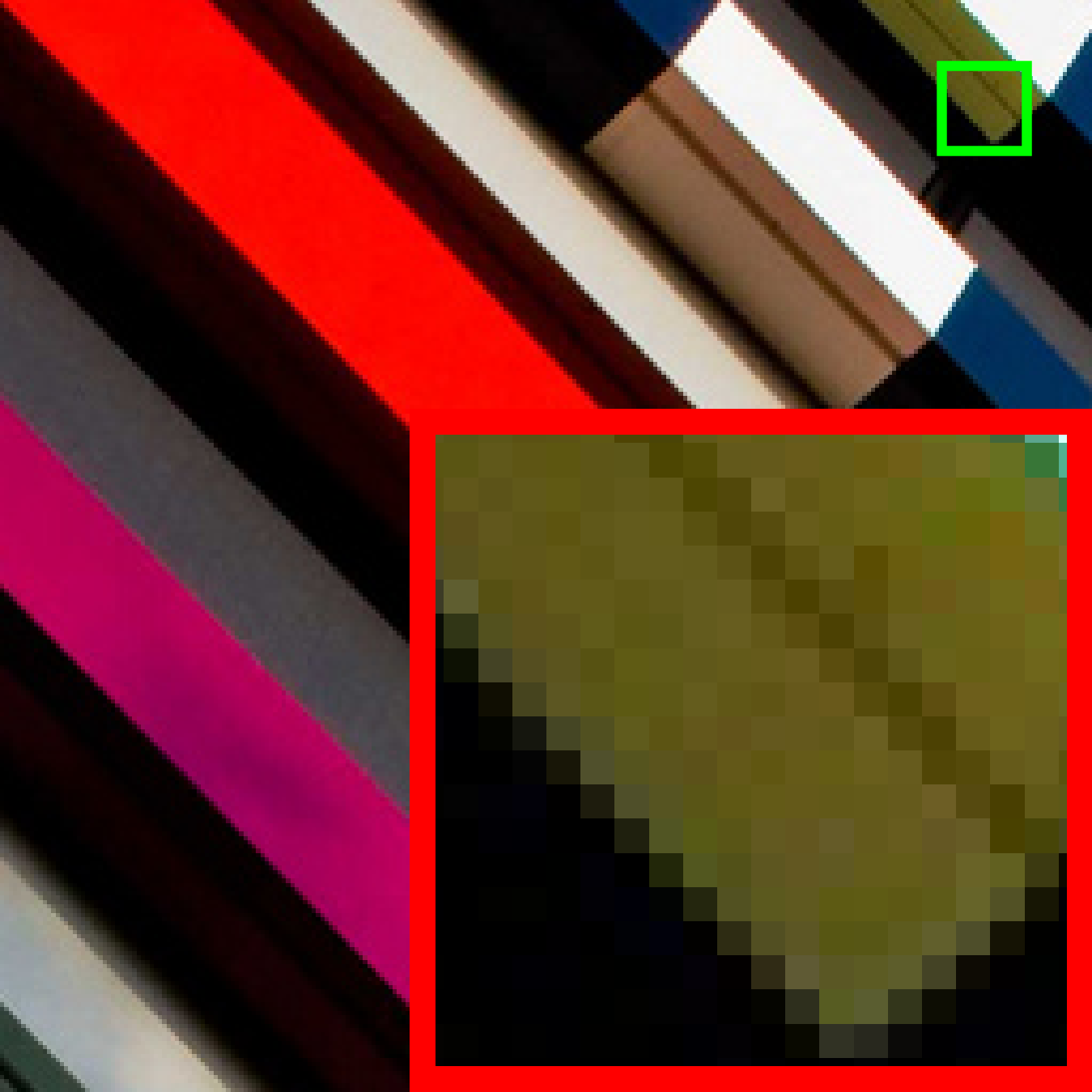}
&\includegraphics[width=0.09\textwidth]{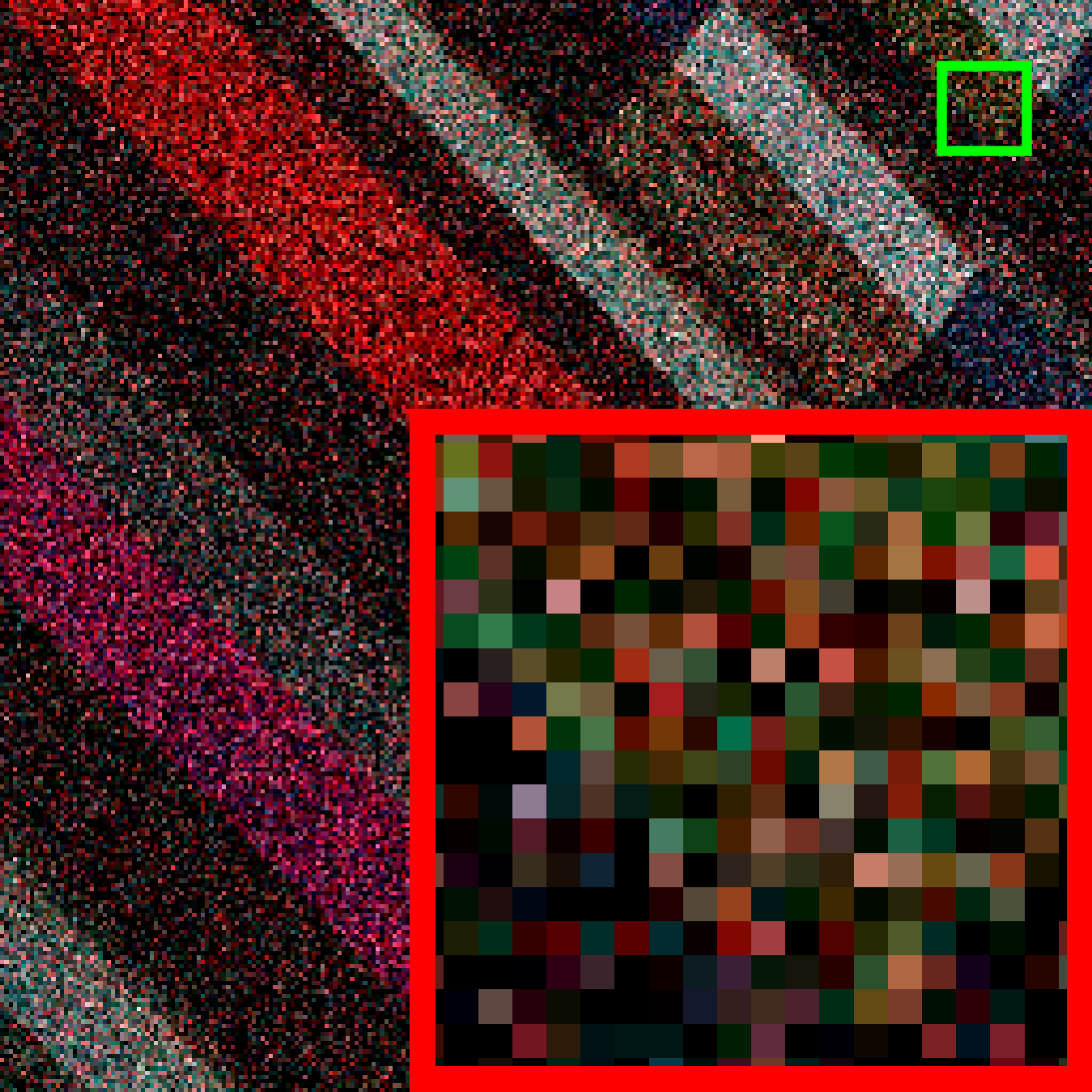}
&\includegraphics[width=0.09\textwidth]{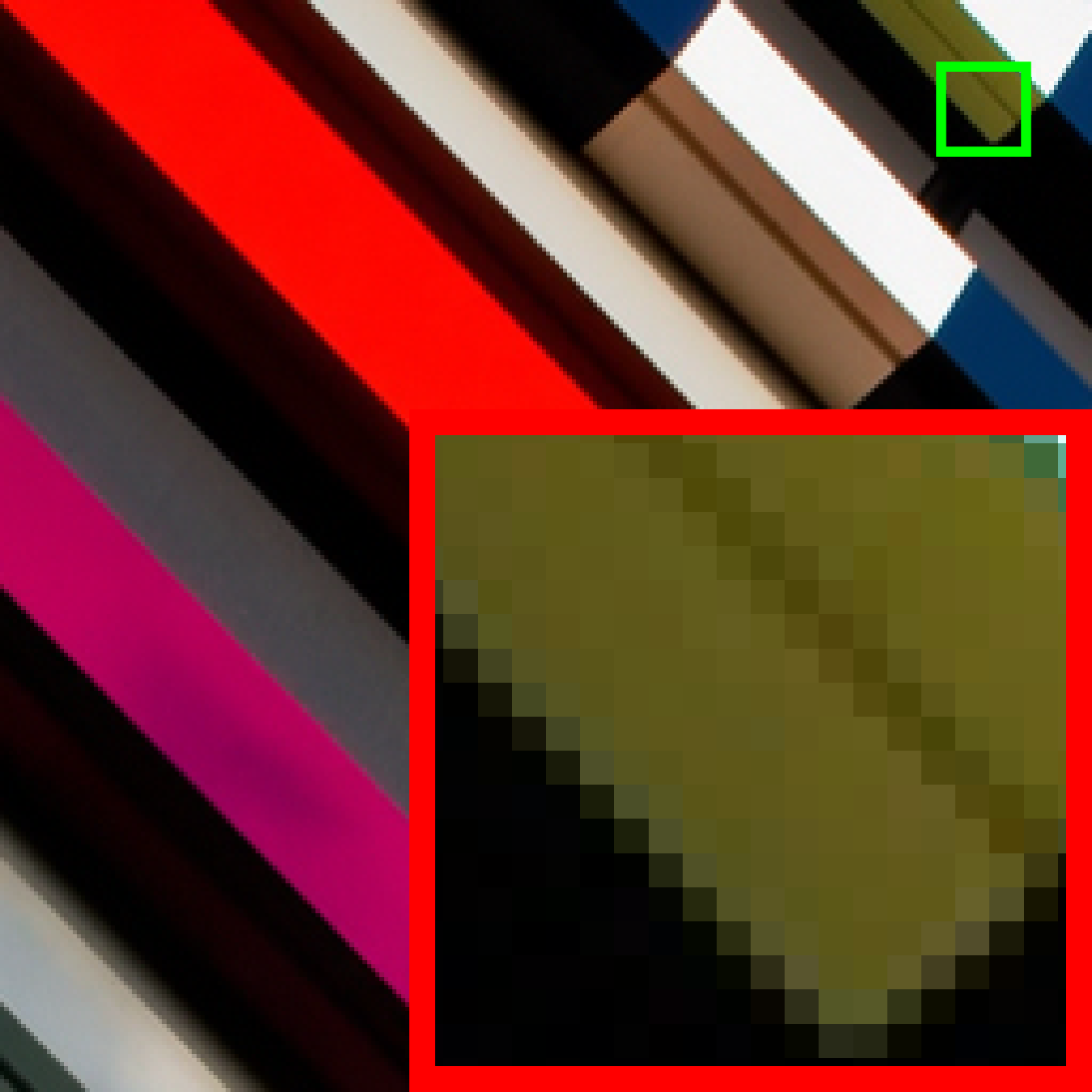}
&\includegraphics[width=0.09\textwidth]{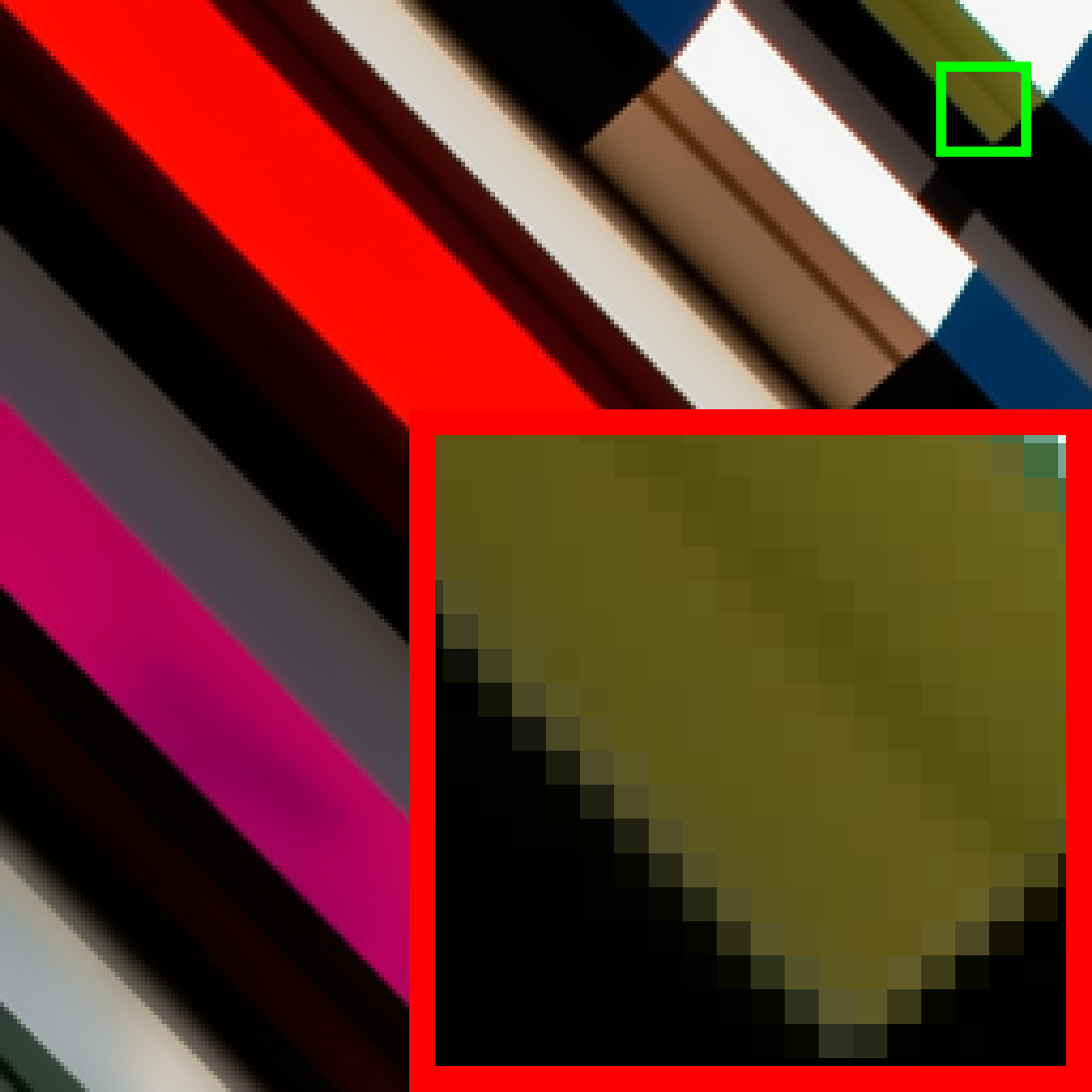}
&\includegraphics[width=0.09\textwidth]{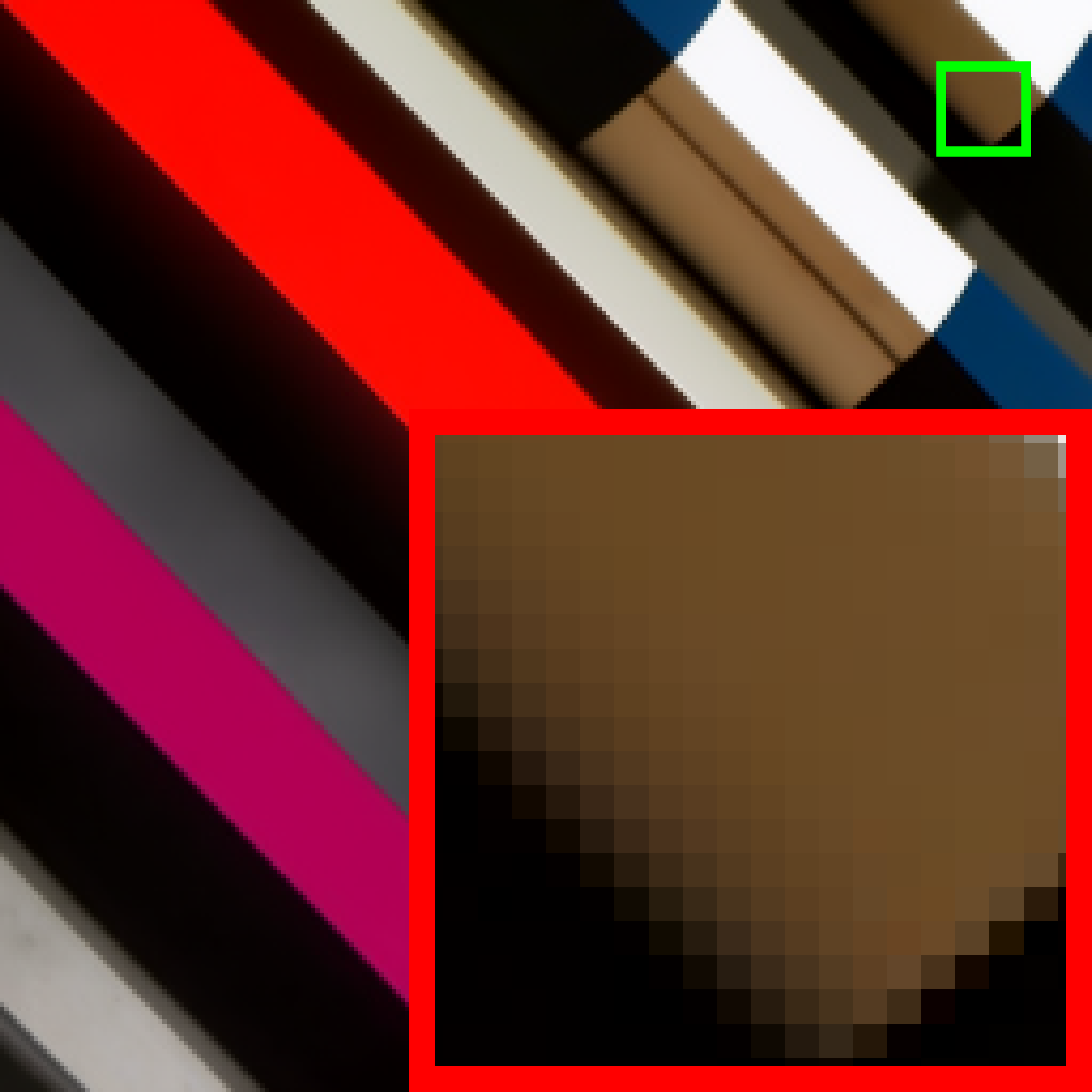}
&\includegraphics[width=0.09\textwidth]{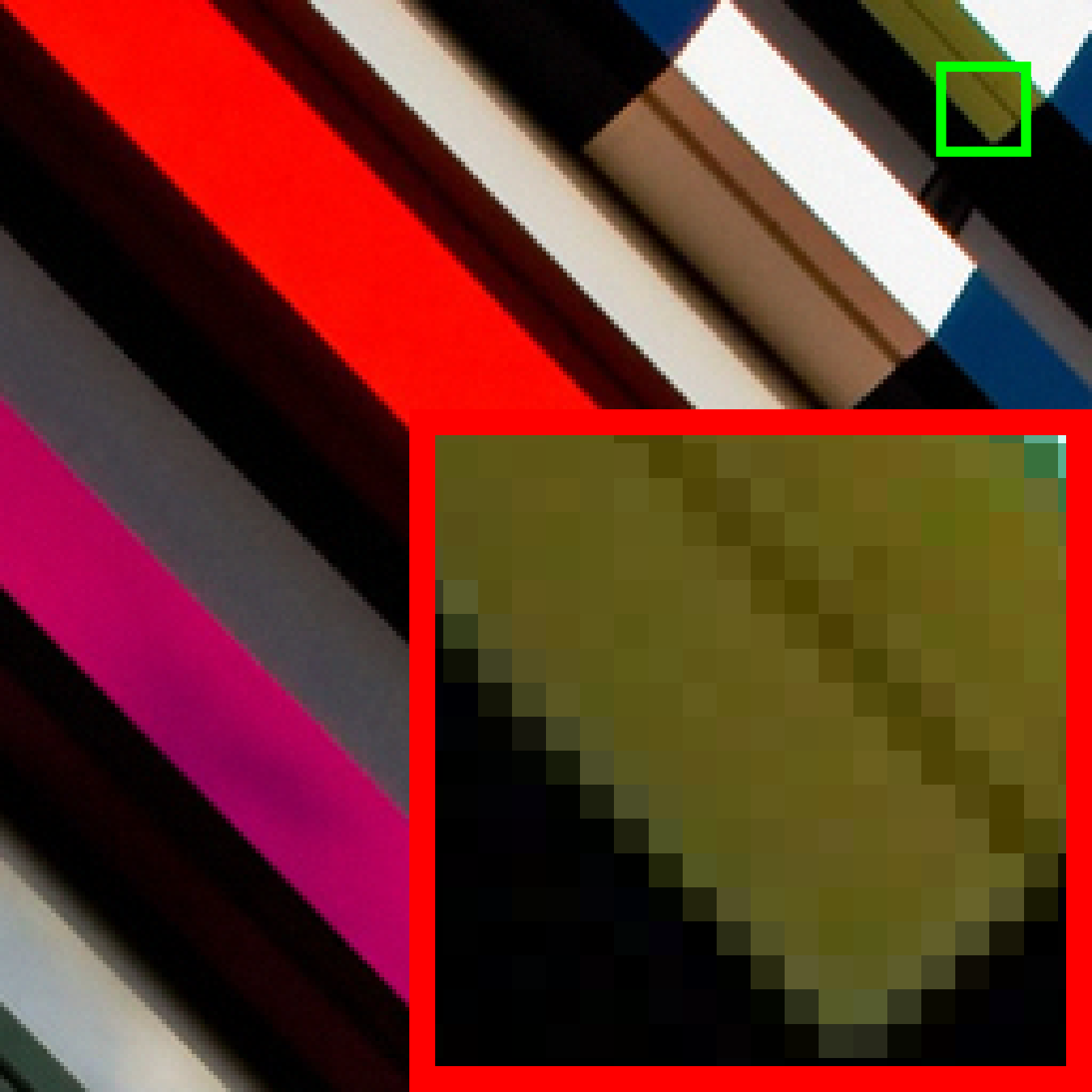}
&\includegraphics[width=0.09\textwidth]{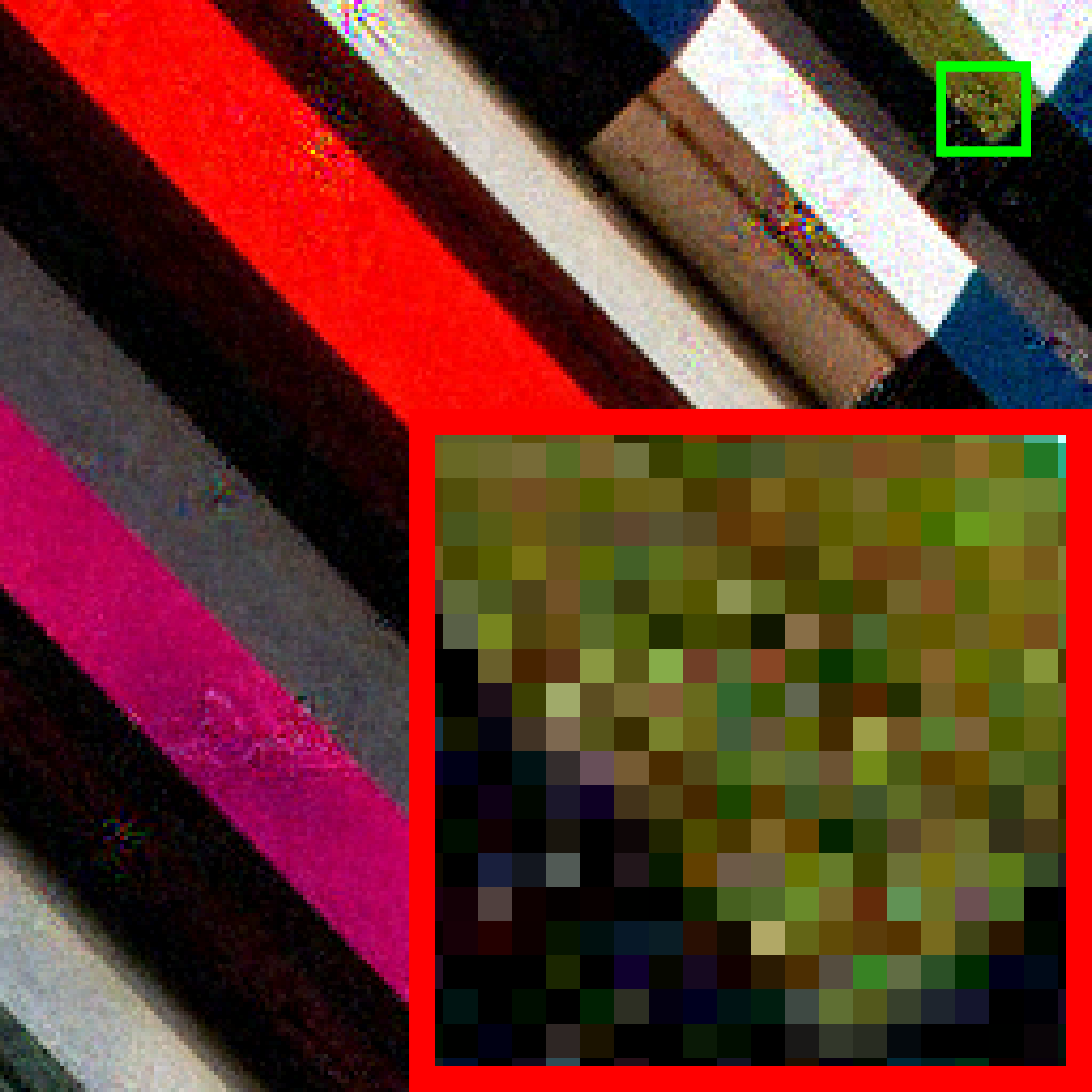}
&\includegraphics[width=0.09\textwidth]{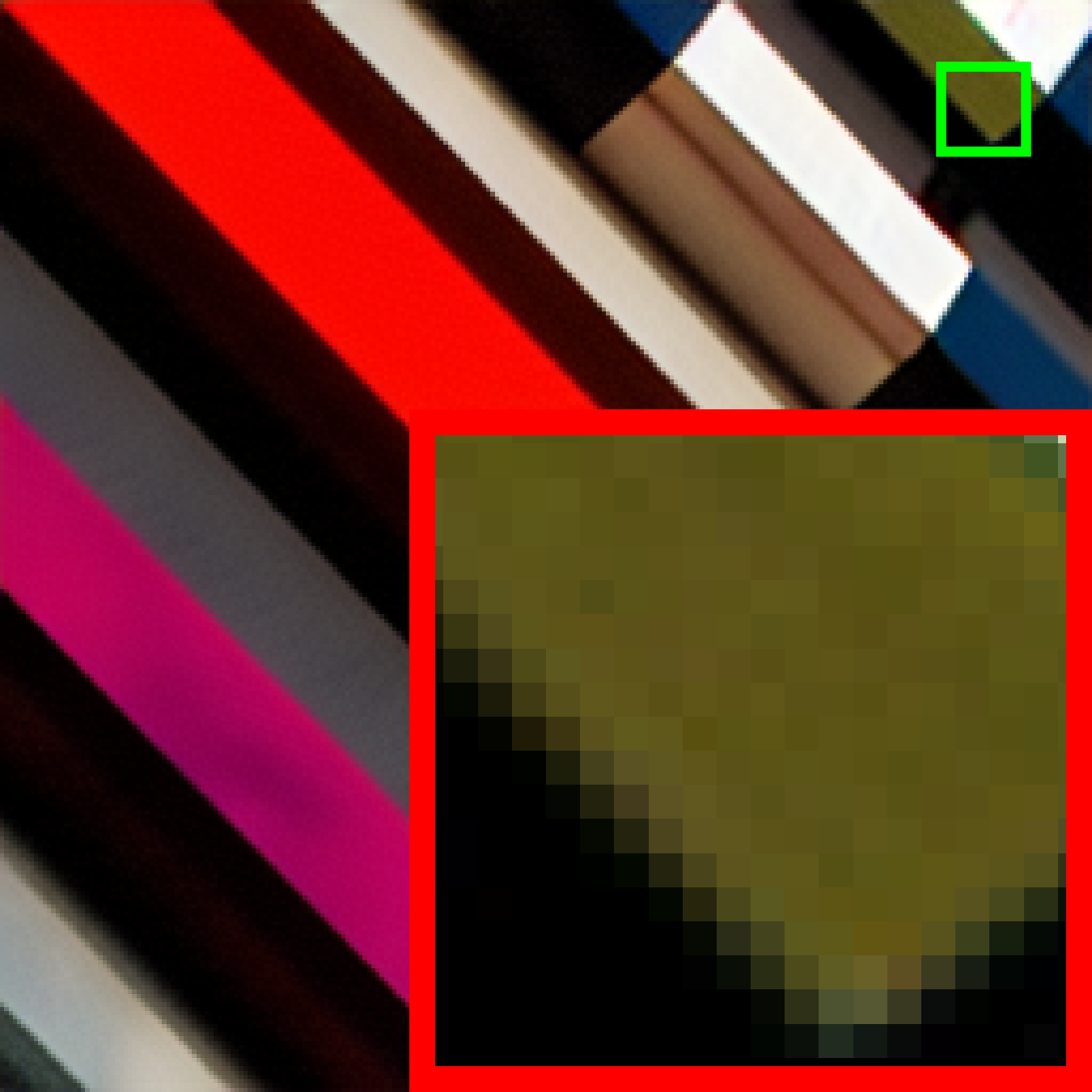}
&\includegraphics[width=0.09\textwidth]{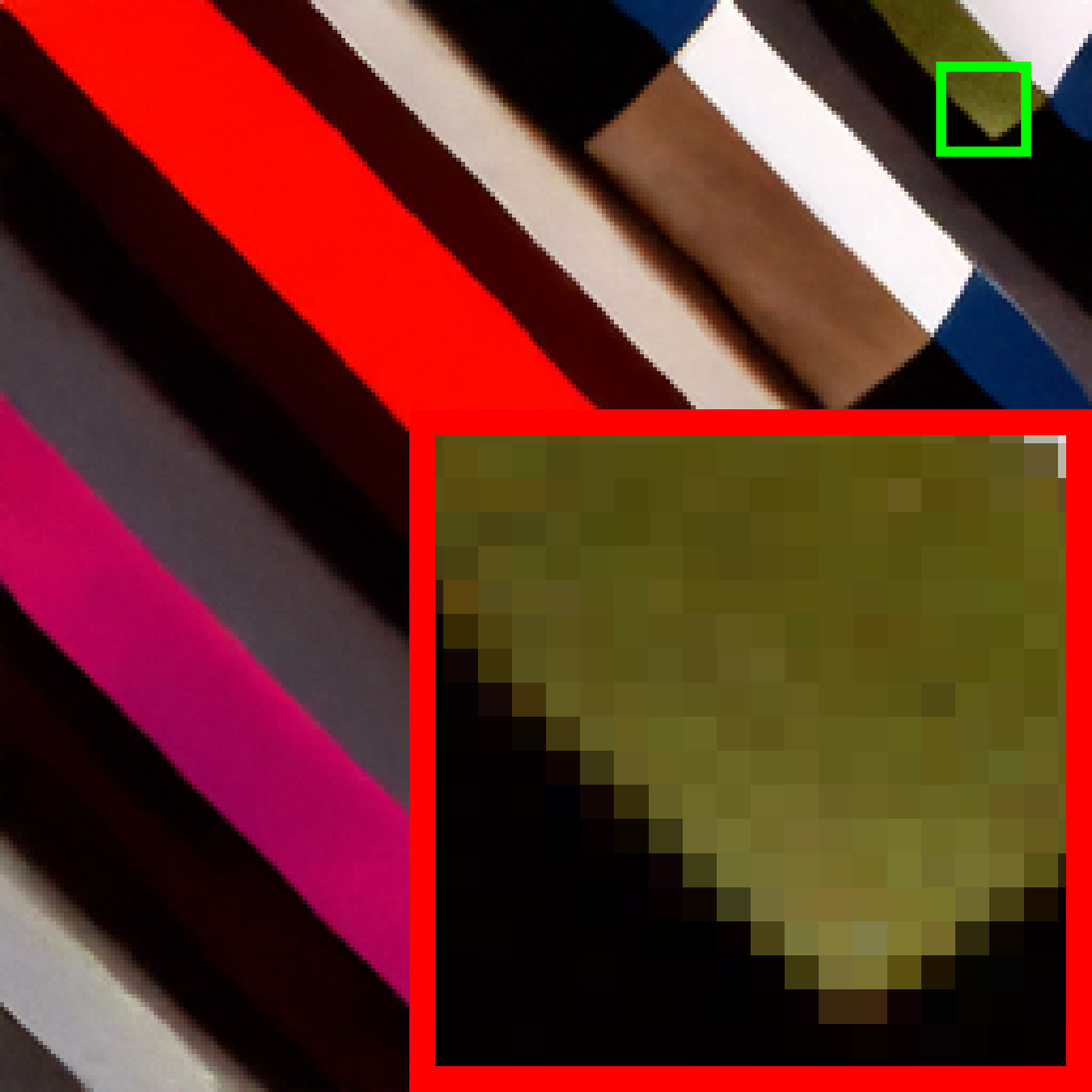}
&\includegraphics[width=0.09\textwidth]{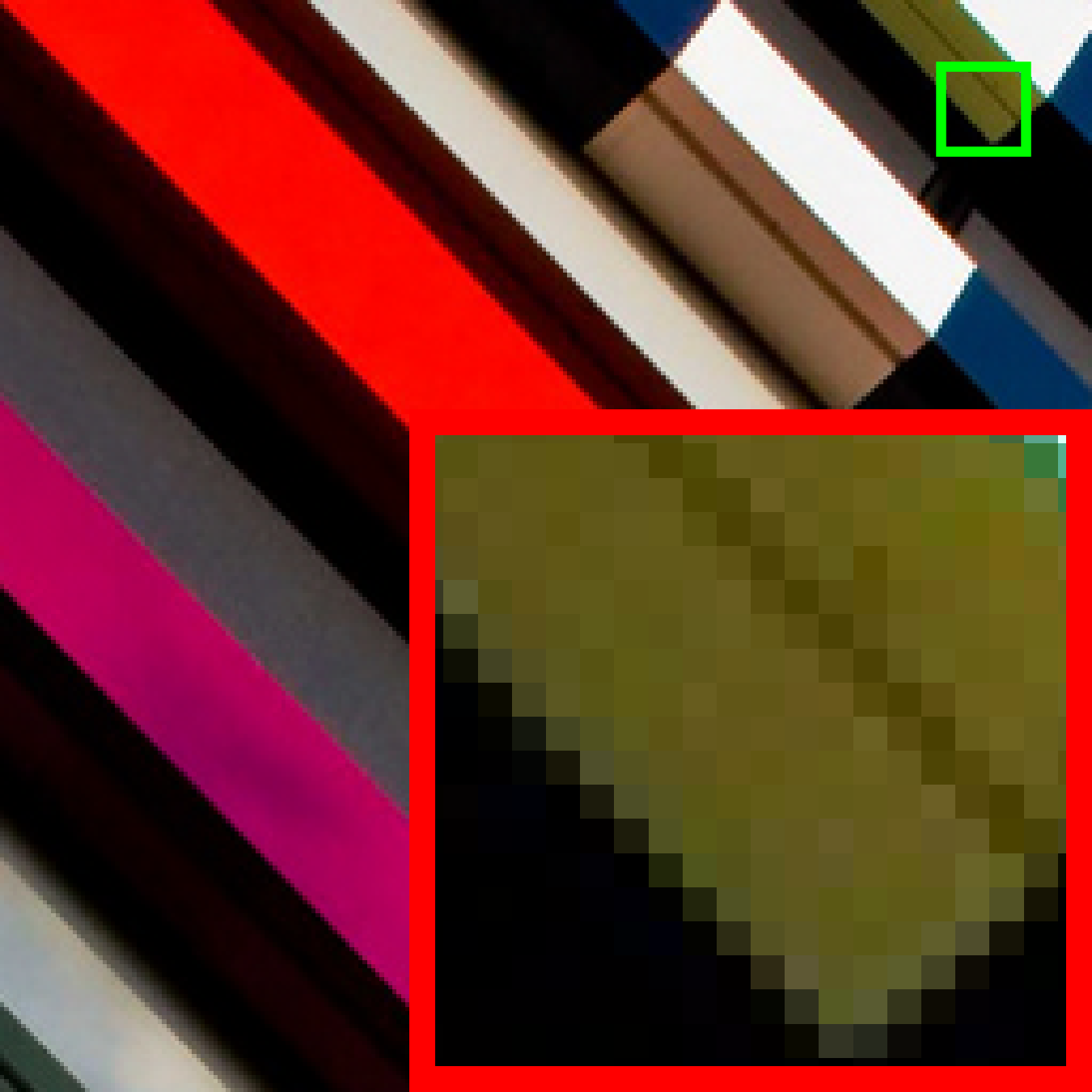}\\
PSNR/SSIM & 11.73/0.1015 & 40.29/0.9617 & 37.74/0.9459 & 29.55/0.8405 & \secondbest{42.26}/\secondbest{0.9684} & 27.94/0.6550 & 30.60/0.8905 & 27.76/0.8273 & \best{46.83}/\best{0.9895}
\end{tabular}}
\resizebox{1.0\textwidth}{!}{
\scriptsize
\begin{tabular}{ccccccccc>{\columncolor[HTML]{FFEEED}}c}
Ground Truth & $\A^\dagger \y$ & DDRM & $\Pi$GDM & DPS & DDNM & GDP & PSLD & SR3 & \textbf{IDM (Ours)}\\
\includegraphics[width=0.09\textwidth]{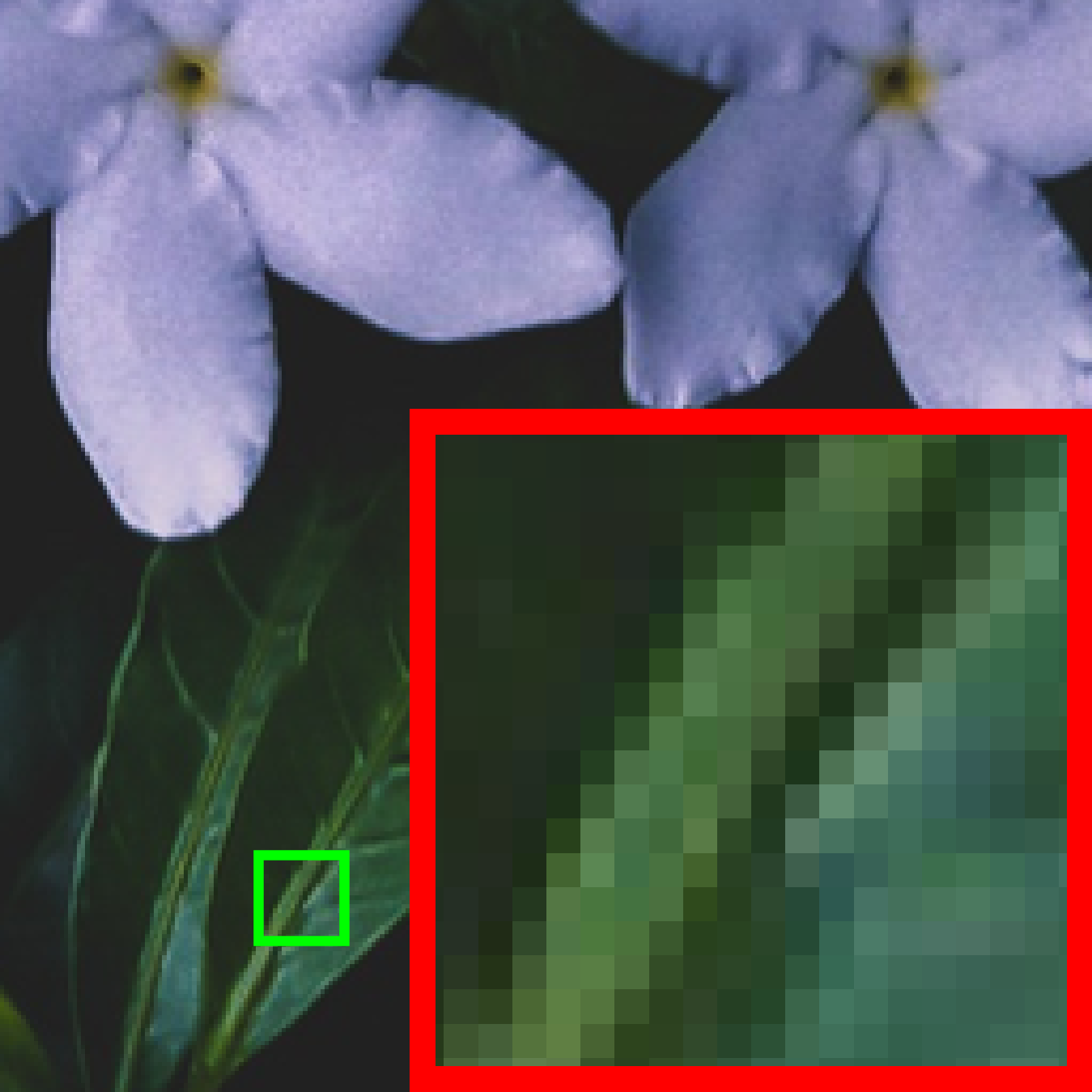}
&\includegraphics[width=0.09\textwidth]{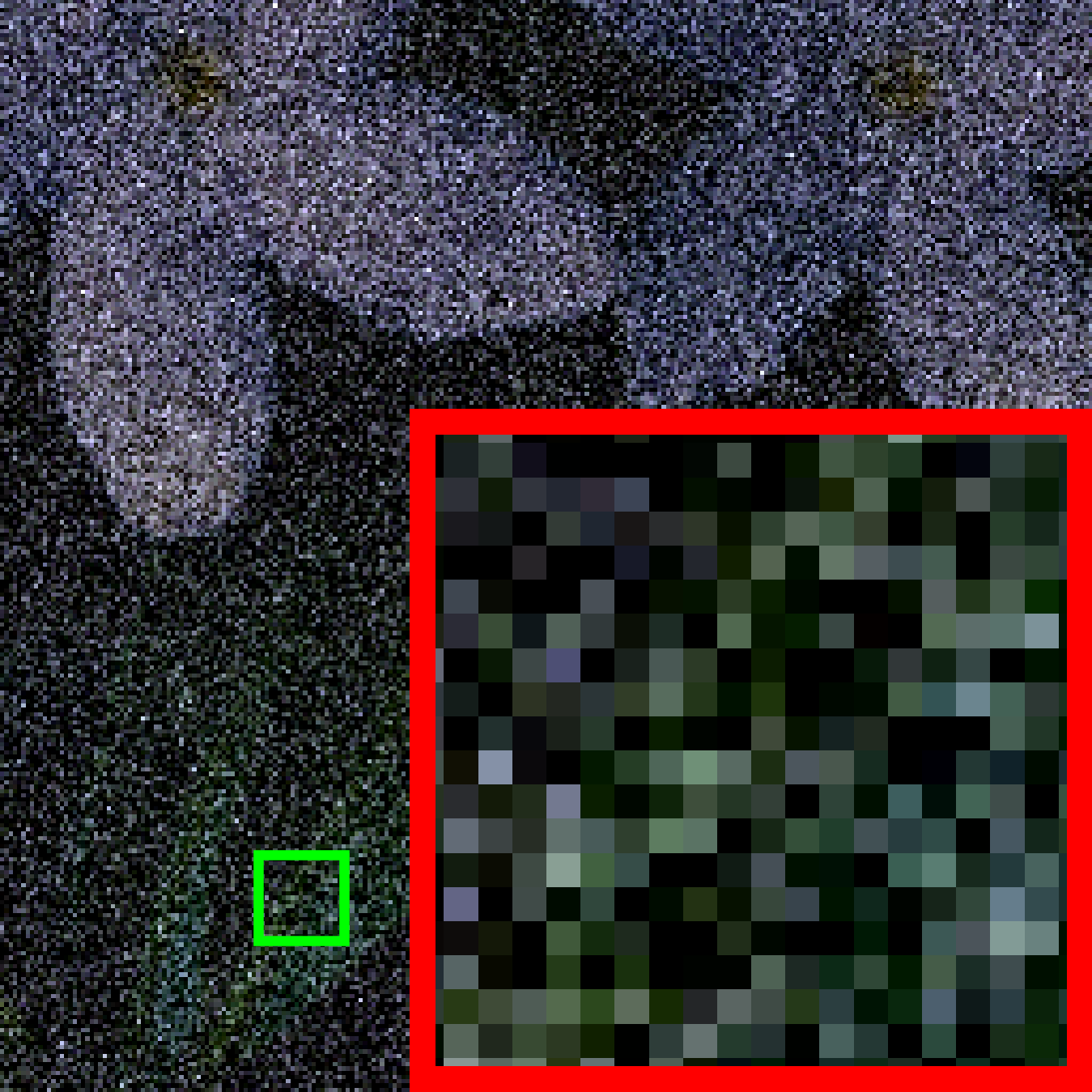}
&\includegraphics[width=0.09\textwidth]{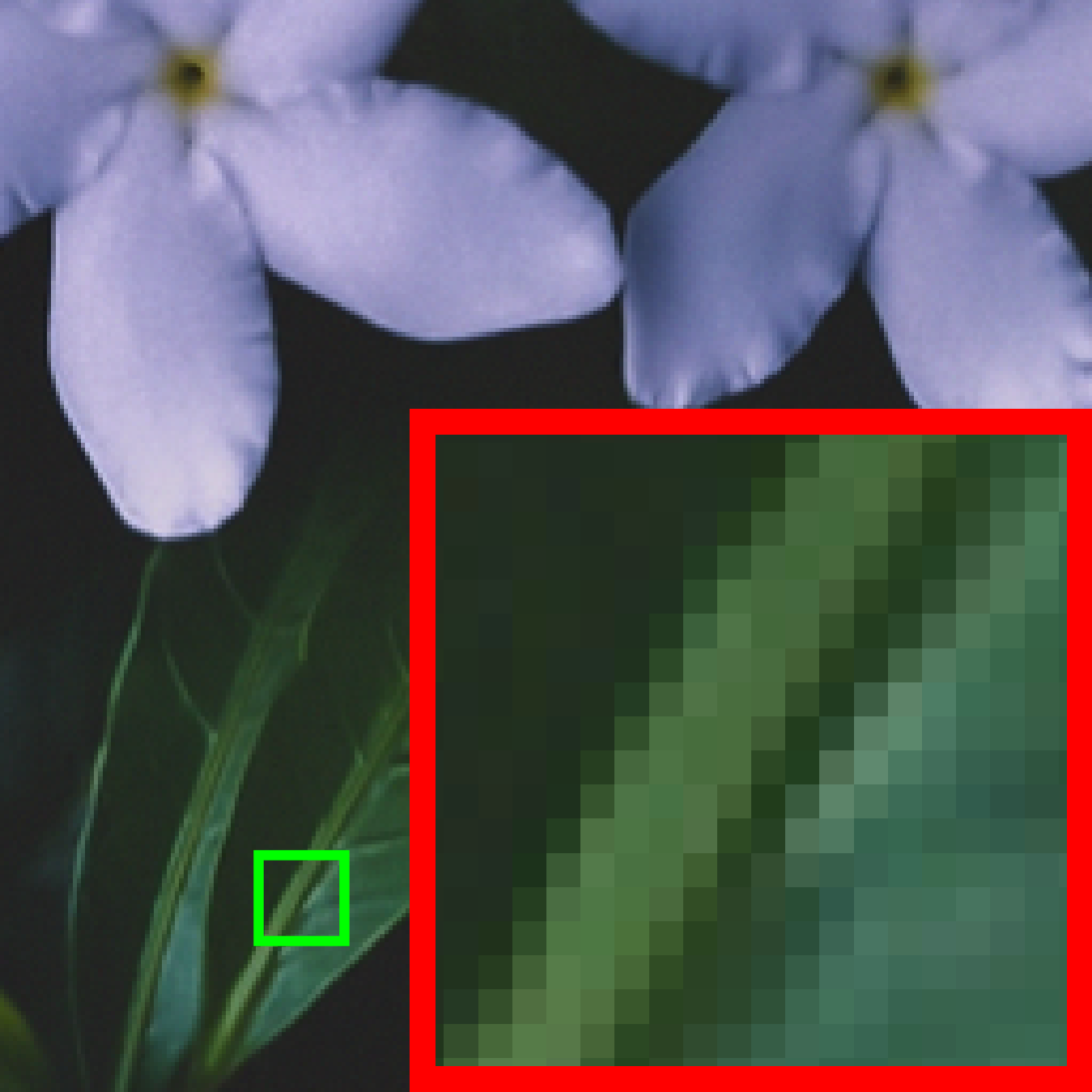}
&\includegraphics[width=0.09\textwidth]{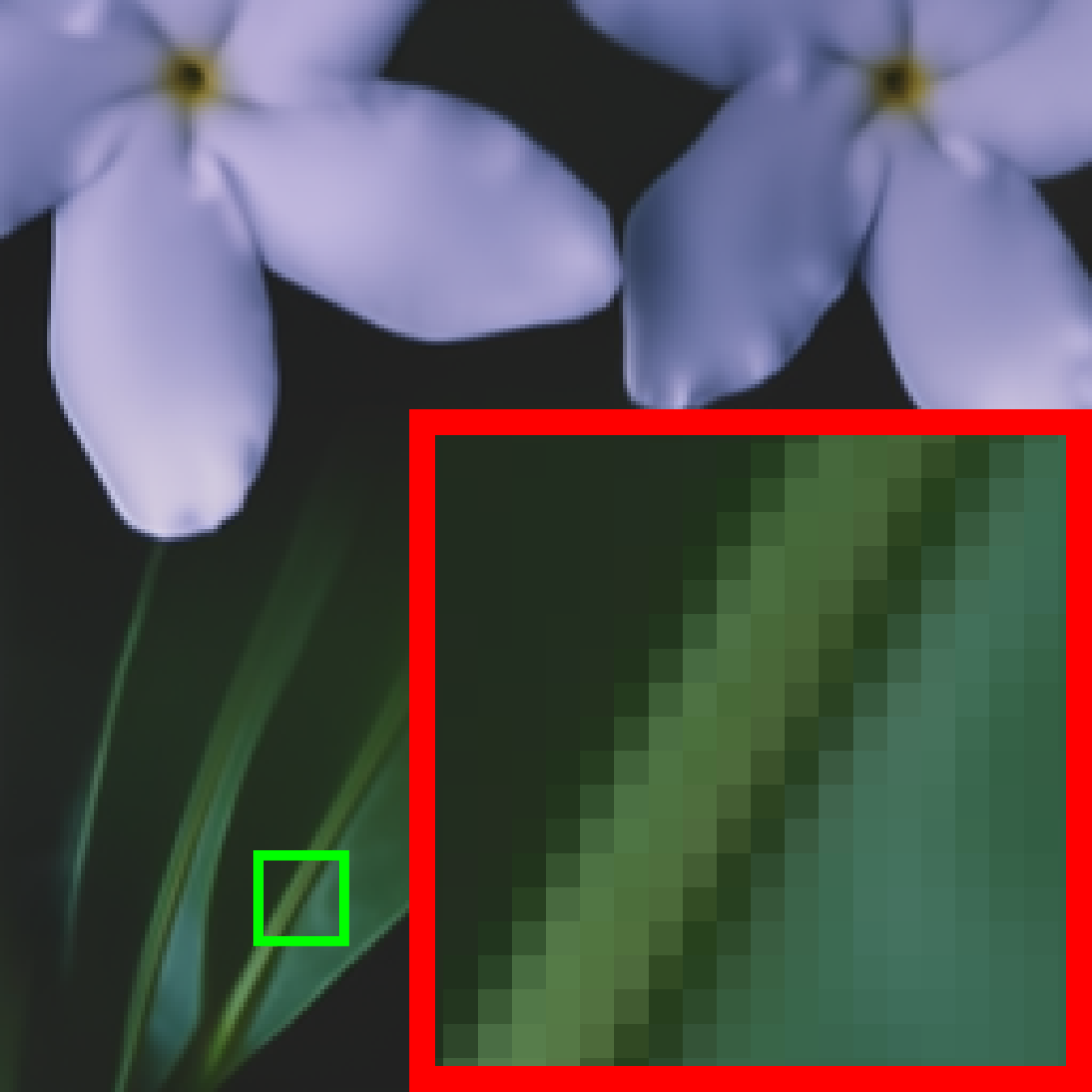}
&\includegraphics[width=0.09\textwidth]{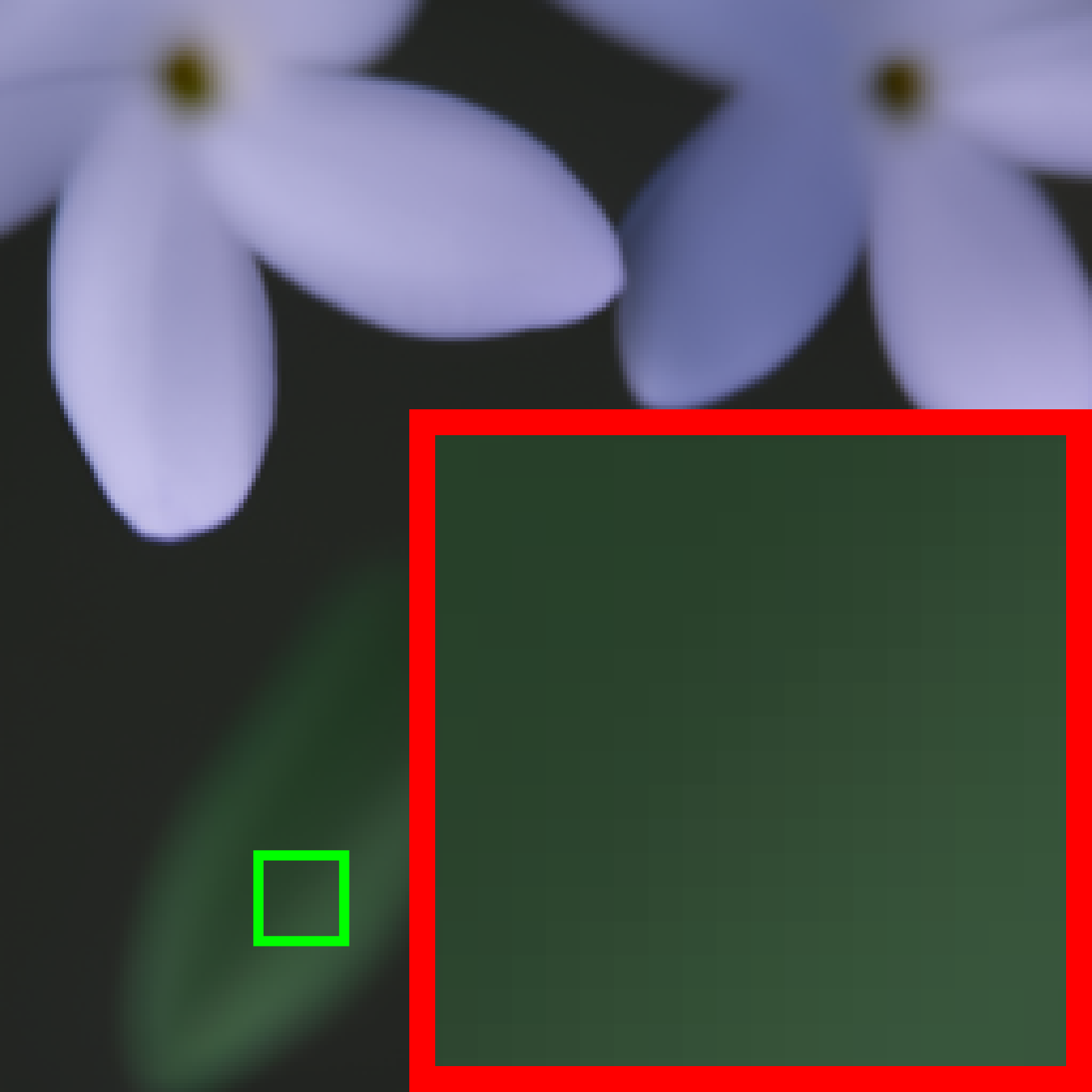}
&\includegraphics[width=0.09\textwidth]{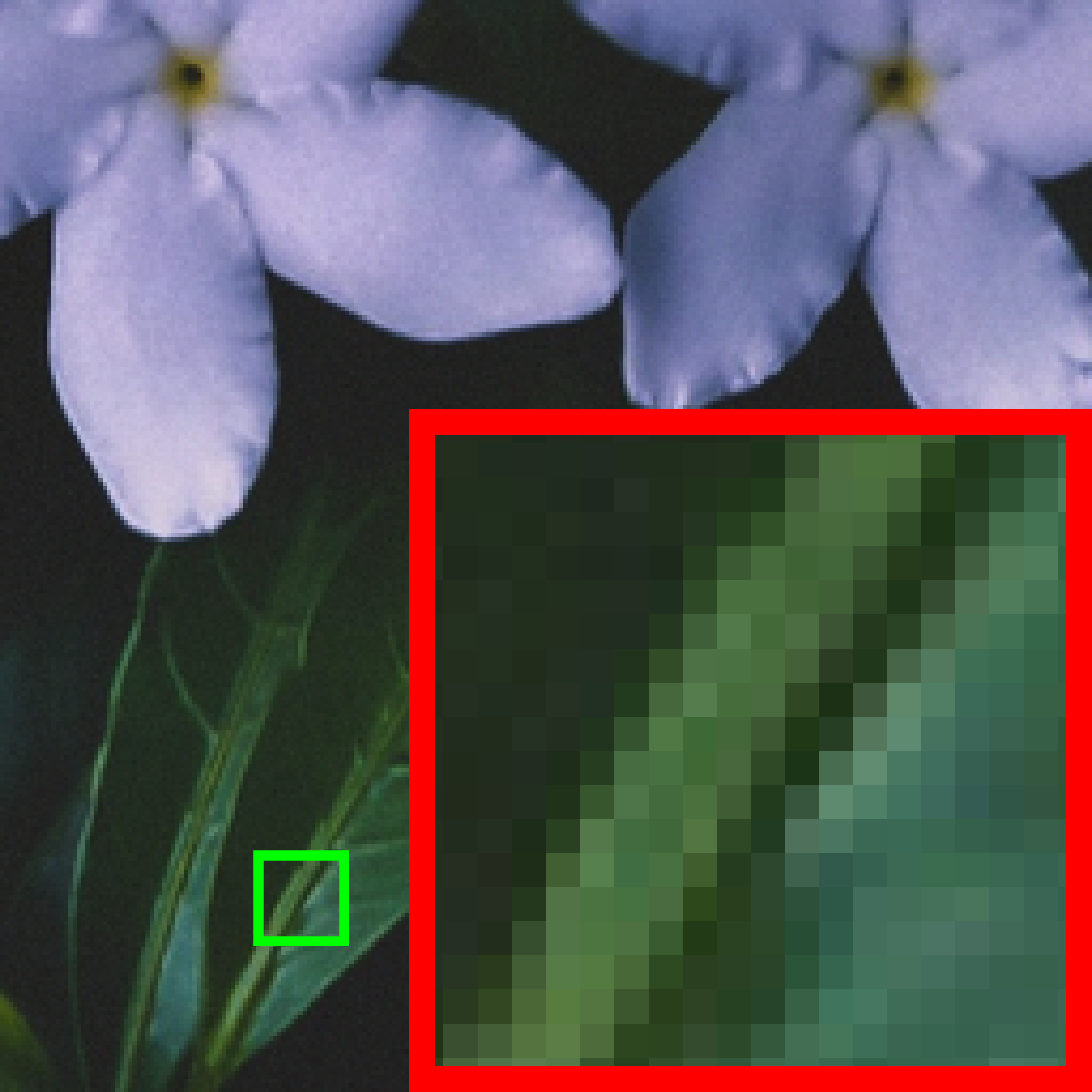}
&\includegraphics[width=0.09\textwidth]{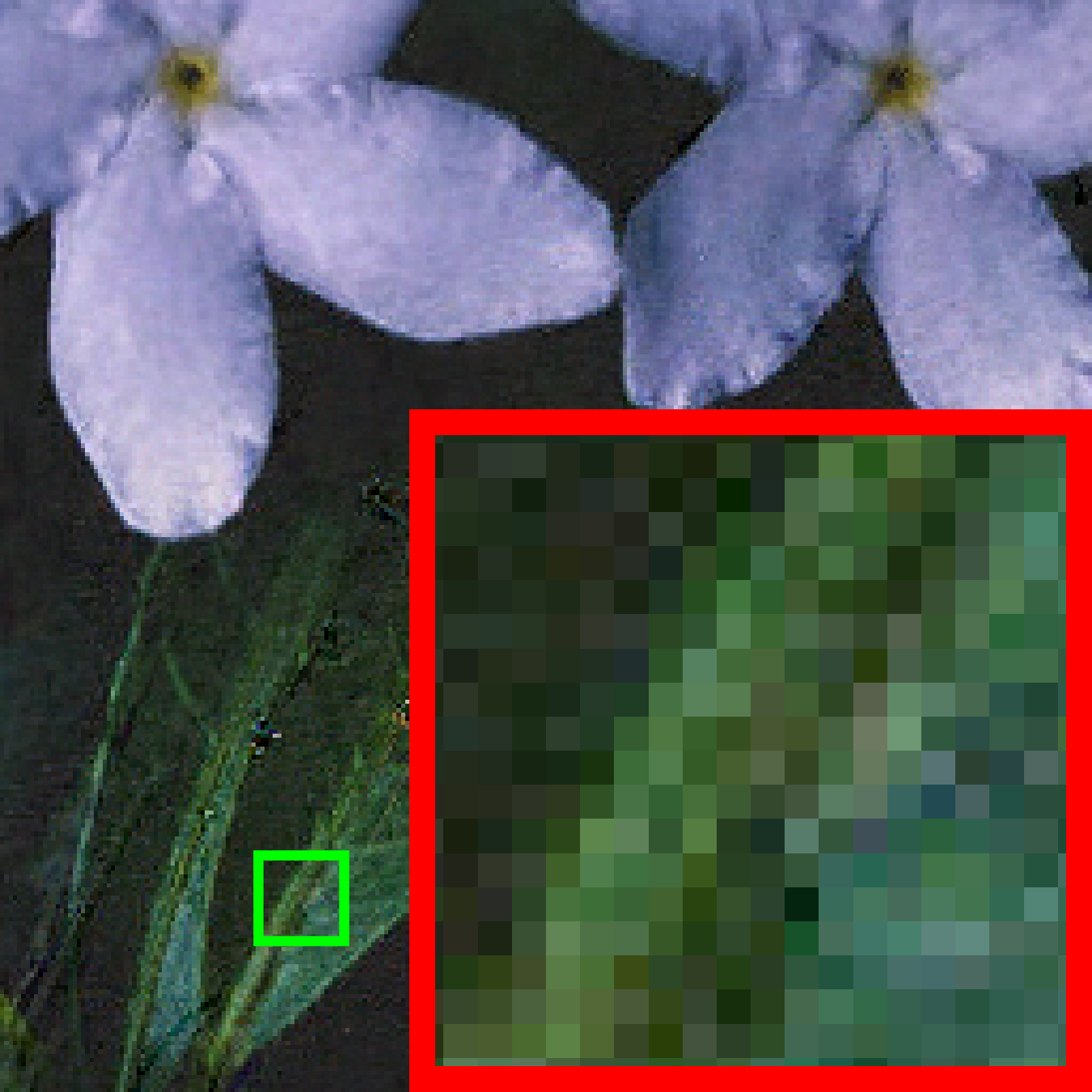}
&\includegraphics[width=0.09\textwidth]{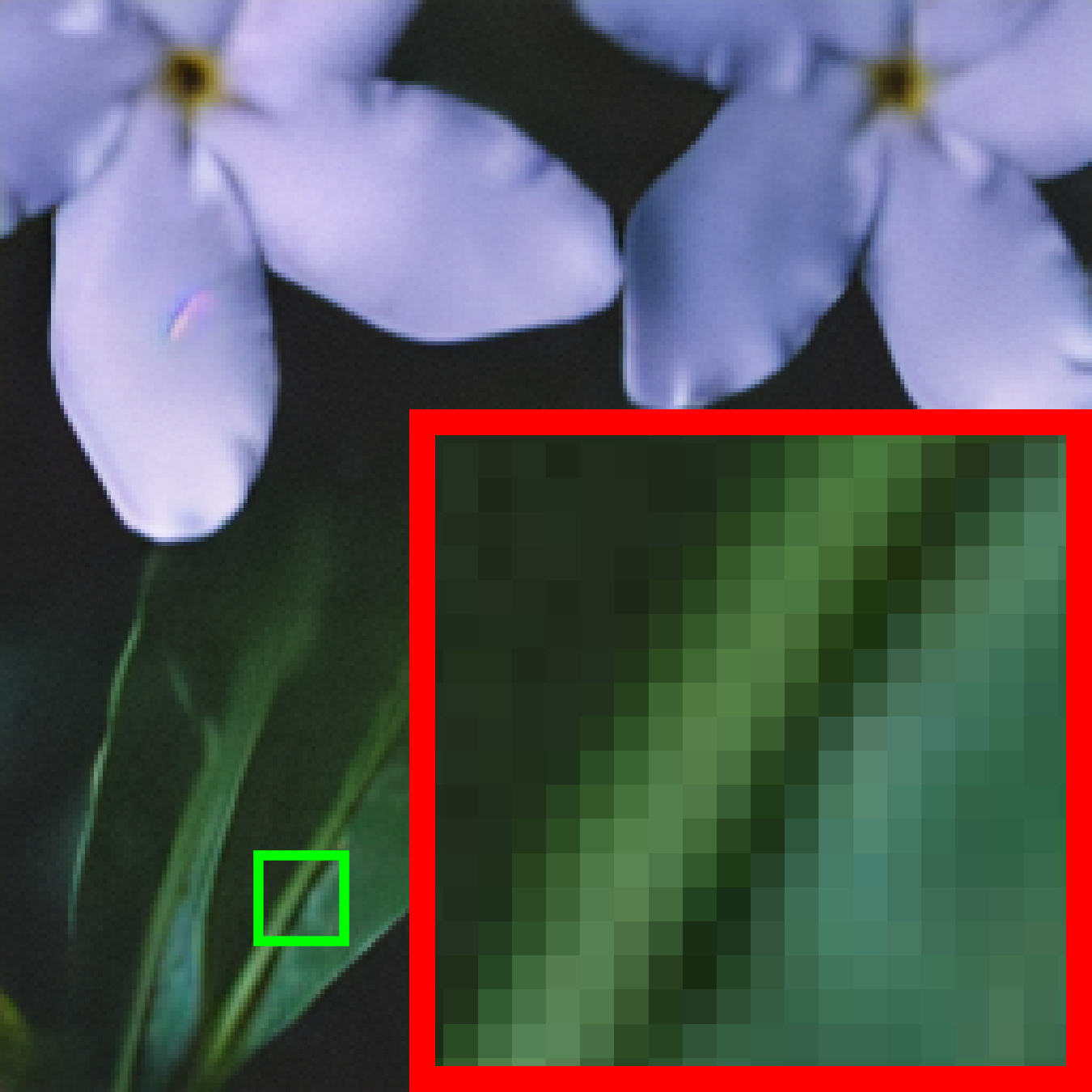}
&\includegraphics[width=0.09\textwidth]{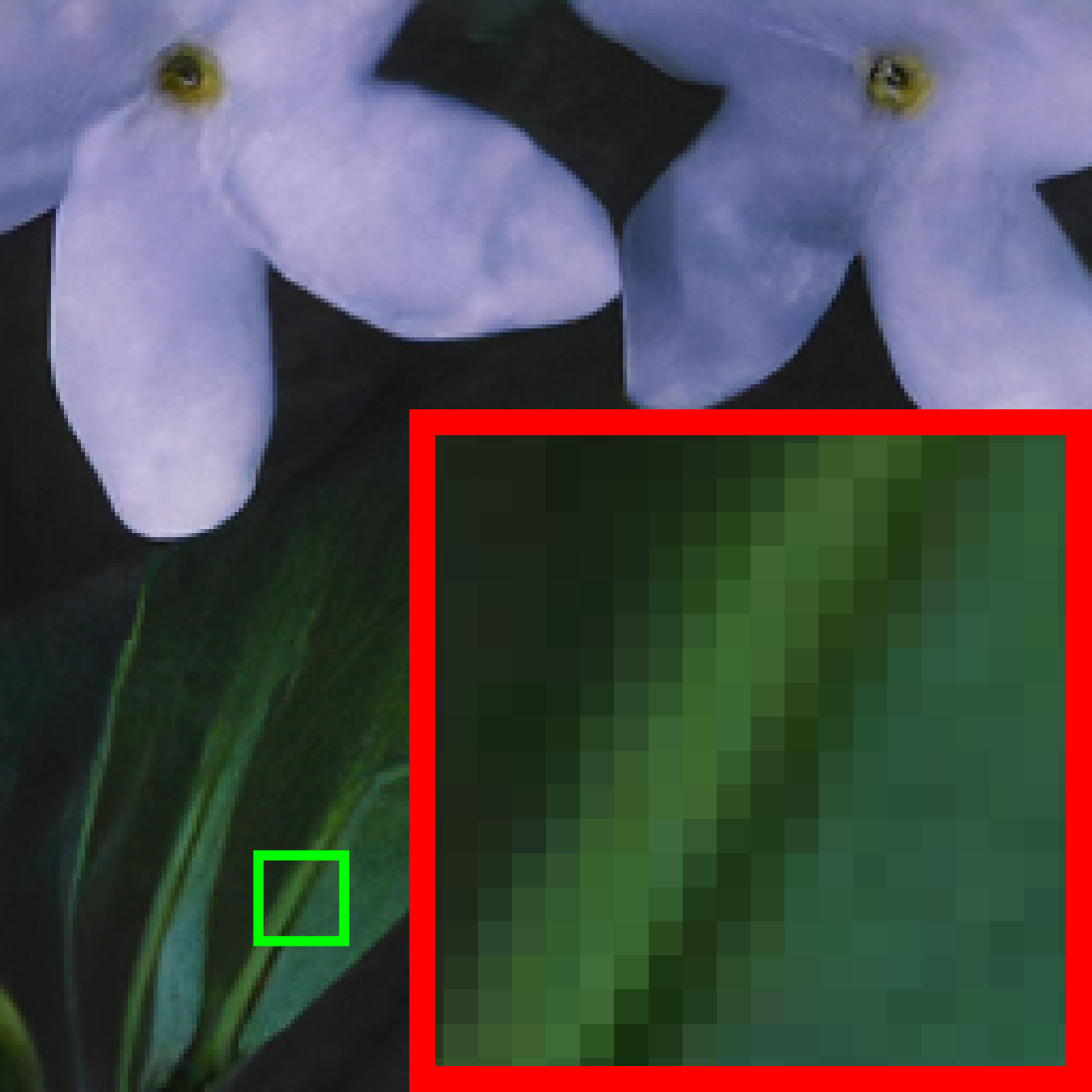}
&\includegraphics[width=0.09\textwidth]{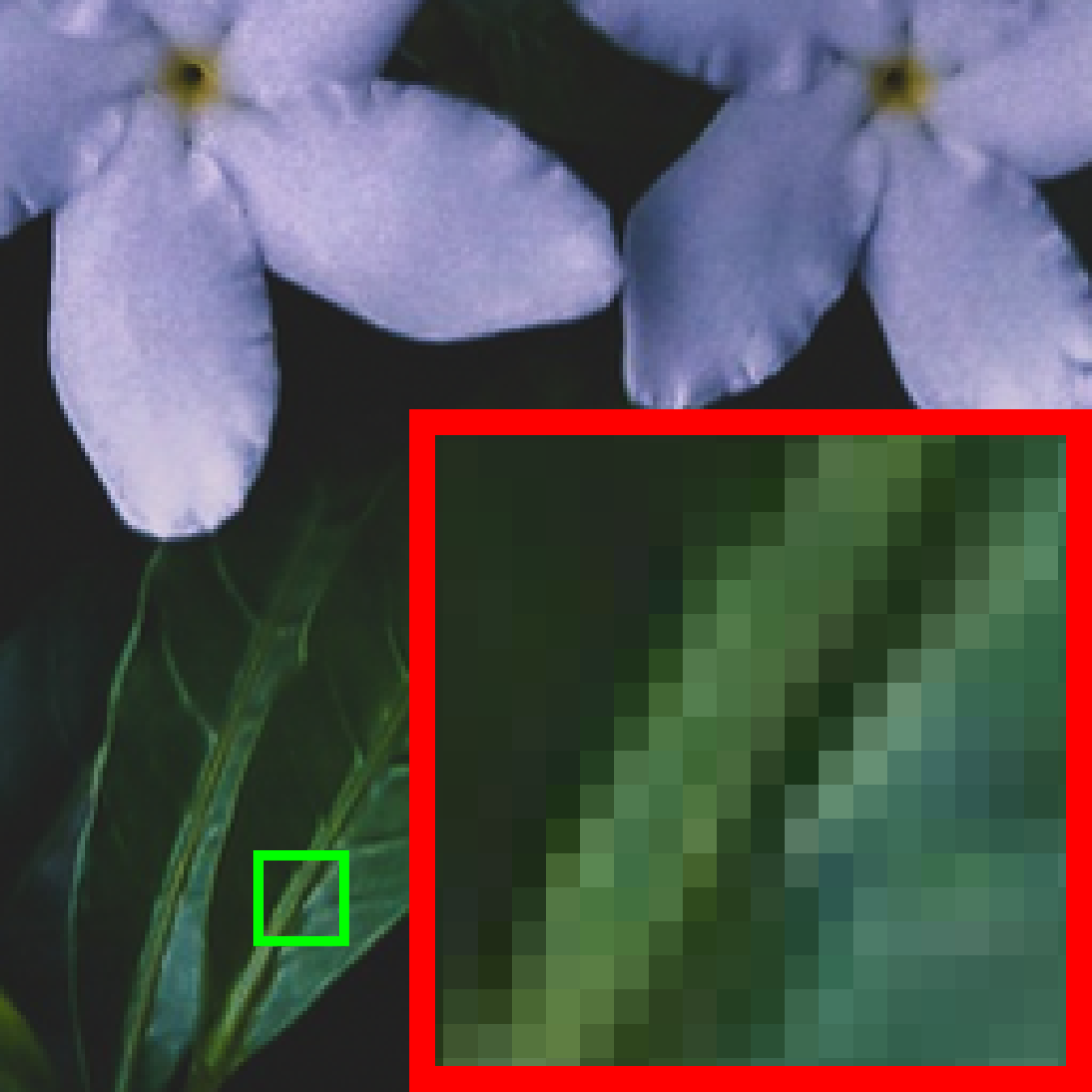}\\
PSNR/SSIM & 12.79/0.0768 & 39.66/0.9581 & 35.85/0.9292 & 27.71/0.8150 & \secondbest{40.72}/\secondbest{0.9562} & 29.74/0.7529 & 28.18/0.9025 & 27.65/0.8151 & \best{52.17}/\best{0.9969}
\end{tabular}}
\vspace{-10pt}
\caption{\textbf{Comparison of CS recovery results} on two images named ``img\_081" from Urban100 \textcolor{blue}{\textbf{(top)}} and ``0877" from DIV2K \textcolor{blue}{\textbf{(bottom)}} at $\gamma =50\%$.}
\label{fig:comp_diff_urban100_and_div2k}
\end{figure*}

\section{Experiment\protect \footnote{Please refer to Sec. C in the \SM for our more experimental results and analyses.}}
\subsection{Implementation Details}
\textbf{Architectural Customization.} Moving beyond the original DDNM \cite{wang2023zero} approach based on an unconditional image diffusion model\footnote{{\url{https://github.com/openai/guided-diffusion}}} \cite{dhariwal2021diffusion} pre-trained on the ImageNet dataset \cite{deng2009imagenet}, our method reuses the noise estimation NN of SD v1.5\footnote{{\url{https://huggingface.co/stable-diffusion-v1-5/stable-diffusion-v1-5}}} pre-trained on the LAION dataset \cite{schuhmann2022laion}, yielding improved performance after fine-tuning (see Tab.~\ref{tab:abla} (6) \vs (11)). We adapt this noise estimator to IDM by introducing PixelUnshuffle and PixelShuffle \cite{shi2016real} layers with scaling ratio $s=2$ at the first and last convolutions to match the four-channel data format of SD. As illustrated in Fig.~\ref{fig:unet} (a), we further simplify the U-Net by removing its time embeddings, cross-attention layers\footnote{In this study, for diffusion models using the SD U-Net architecture that retain text input, embedding, and cross-attention layers, a null text prompt (i.e., ``'') is used as the default condition. For IDM, which removes these components entirely, no text input is applied.}, and the final three scales to balance efficiency and performance. Notably, this customized framework differs from other zero-shot, plug-and-play, or conditional models, as it allows rapid end-to-end fine-tuning of all reused weights for CS, which is advantageous for applications where the prior of the modified, pre-trained diffusion model does not fully align with the target task.

\begin{table*}[!t]
\caption{\textbf{Ablation study on our contributions and developed techniques for enhancement on performance and efficiency.} \textcolor{blue}{Arch.:} Architecture choice of the noise estimator. \textcolor{blue}{E2E:} End-to-end training. \textcolor{blue}{Inv.:} Invertibility. \textcolor{blue}{Reu.:} Reuse of pre-trained NN weights. \textcolor{blue}{Inj.:} Injectors. \textcolor{blue}{Pru.:} Pruning. \textcolor{blue}{PSNR:} Average PSNR (dB) for RGB image CS on Urban100/DIV2K at $\gamma = 10\%$. \textcolor{blue}{Mem.:} GPU memory usage (GB). \textcolor{blue}{Tra. $t$:} Training time (hours) to convergence. \textcolor{blue}{NFEs:} Neural function evaluations. \textcolor{blue}{Inf. $t$:} Inference time per image (seconds). \textcolor{blue}{Size:} Storage size (GB). \textcolor{blue}{X$^*$:} Modified version of method ``X''. \textcolor{blue}{UIDM:} Unconditional image diffusion model \cite{dhariwal2021diffusion}. \textcolor{blue}{N/A:} Not applicable.}
\vspace{-10pt}
\label{tab:abla}
\setlength{\tabcolsep}{3pt}
\centering
\resizebox{1.0\textwidth}{!}{
\begin{tabular}{cl|c|ccccc|l|cccccc}
\shline
\rowcolor[HTML]{EFEFEF}
\multicolumn{2}{c|}{\cellcolor[HTML]{EFEFEF}Method} & Arch. of $\ep_\Th$ & E2E & Inv. & Reu. & Inj. & Pru. & \multicolumn{1}{|c|}{\cellcolor[HTML]{EFEFEF}Initialization} & PSNR ($\uparrow$) & Mem. ($\downarrow$) & Tra. $t$ ($\downarrow$) & NFEs ($\downarrow$) & Inf. $t$ ($\downarrow$) & Size ($\downarrow$) \\ \hline \hline
(1) & DDNM$^{*}$ & & \redcross & \redcross & \redcross & \redcross & \redcross & $\xhat_T\sim\mathcal{N}(\mathbf{0}_N,\mathbf{I}_N)$ & 22.86/25.13 & 22.3 & 417.3 & \secondbest{100} & 42.30 & 3.4 \\
(2) & DDNM$^{*}$ & & \redcross & \redcross & \greencheck & \redcross & \redcross & $\xhat_T\sim\mathcal{N}(\mathbf{0}_N,\mathbf{I}_N)$ & 24.28/26.25 & 22.3 & 40.6 & \secondbest{100} & 42.37 & 3.4 \\
(3) & IDM$^{*}$ & & \greencheck & \redcross & \redcross & \redcross & \redcross & $\xhat_T=\sqrt{\bar{\alpha}_T}\A^{\dagger}\y$ & 26.17/28.09 & 34.6 & 411.1 & \best{2} & 1.72 & 3.4 \\
(4) & IDM$^{*}$ & & \greencheck & \greencheck & \redcross & \redcross & \redcross & $\xhat_T=\sqrt{\bar{\alpha}_T}\A^{\dagger}\y$ & 26.21/28.14 & 13.4 & 820.7 & \best{2} & 1.69 & 3.4 \\
(5) & IDM$^{*}$ & & \greencheck & \redcross & \greencheck & \redcross & \redcross & $\xhat_T=\sqrt{\bar{\alpha}_T}\A^{\dagger}\y$ & 26.57/28.33 & 34.6 & 12.4 & \best{2} & 1.69 & 3.4 \\
(6) & IDM$^{*}$ & & \greencheck & \greencheck & \greencheck & \redcross & \redcross & $\xhat_T=\sqrt{\bar{\alpha}_T}\A^{\dagger}\y$ & 26.52/28.31 & 13.4 & 24.5 & \best{2} & 1.68 & 3.4 \\
(7) & IDM$^{*}$ & & \greencheck & \greencheck & \greencheck & \greencheck & \redcross & $\xhat_T=\sqrt{\bar{\alpha}_T}\A^{\dagger}\y$ & \best{29.53}/\best{30.41} & 13.4 & 26.4 & \best{2} & 1.77 & 3.4 \\
(8) & IDM$^{*}$ & & \greencheck & \greencheck & \greencheck & \greencheck & \greencheck & $\xhat_T\sim\mathcal{N}(\mathbf{0}_N,\mathbf{I}_N)$ & 27.21/29.12 & \secondbest{5.0}     & 8.3 & \best{2} & \secondbest{0.38} & \best{0.4} \\
\rowcolor[HTML]{FFEEED} (9) & \textbf{IDM (Ours)} & \multirow{-9}{*}{SD v1.5} & \greencheck & \greencheck & \greencheck & \greencheck & \greencheck & $\xhat_T=\sqrt{\bar{\alpha}_T}\A^{\dagger}\y$ & \secondbest{29.44}/\secondbest{30.37} & \best{4.9} & \secondbest{8.0} & \best{2} & \secondbest{0.38} & \best{0.4} \\ \hline \hline
(10) & DDNM & & \redcross & \redcross & \greencheck & \redcross     & \redcross & $\xhat_T\sim\mathcal{N}(\mathbf{0}_N,\mathbf{I}_N)$ & 20.76/22.18 & N/A & N/A & \secondbest{100} & 9.16 & 2.1 \\
(11) & IDM$^{*}$ & \multirow{-2}{*}{UIDM} & \greencheck & \greencheck & \greencheck & \redcross & \redcross & $\xhat_T=\sqrt{\bar{\alpha}_T}\A^{\dagger}\y$ & 24.45/26.52 & 18.4 & \best{4.2} & \best{2} & \best{0.33} & 2.1 \\ \hline \hline
(12) & PRL-PGD$^+$ & N/A & \greencheck & \redcross & \redcross     & \redcross & \redcross & $\Xhat^{(0)}=\text{Conv}(\A^{\dagger}\y)$ & 26.74/28.32 & 8.6 & 652.4 & N/A & 0.95 & \secondbest{0.8} \\ \shline
\end{tabular}}
\end{table*}

\textbf{All Learnable Parameters in IDM} are jointly fine-tuned end-to-end, including the shared, wired U-Net $\ep_\Th$ and its internal weighting factors (i.e., $v_i$s with $u_i\equiv 1-v_i$) across the $T$ sampling steps, the non-shared and step-specific diffusion parameters $\{\alpha_t\}$, weighting scalars $\{v_t\}$ with $u_t\equiv 1-v_t$ for our invertible diffusion sampling pipeline, the scaling factors $w_T$ and $w_0$, as well as the parameters of our designed $(20T)$ injectors. The values of scalars $\alpha_t$, $v_t$, $v_i$, $w_T$, and $w_0$ are initialized to 0.5, 0.5, 0.5, 1.0, and 0.0, respectively. All experiments employ the Adam \cite{kingma2014adam} optimizer for training.

\subsection{Comparison with State-of-the-Arts}
\textbf{Setup.} IDM is compared against twenty end-to-end learned and eight diffusion-based approaches for natural image CS tasks. Training data pairs $\{(\y,\x)\}$ are generated from the randomly cropped $256\times 256$ patches of the Waterloo exploration database (WED) \cite{ma2016waterloo,zhang2022plug,li2022d3c2,chen2023deep} using structurally random, orthonormalized i.i.d. block-based Gaussian matrices $\A$ of a fixed block size $32\times 32$ \cite{gan2007block,do2012fast,adler2016deep,chen2021deep,chen2024self}. Fine-tuning IDM with $T=3$ and a batch size of 32 for 50000 iterations on 4 NVIDIA A100 (80GB) GPUs and PyTorch \cite{paszke2019pytorch} takes three days, with a starting learning rate of 0.0001, halved every 10000 iterations. Four benchmarks: Set11 \cite{kulkarni2016reconnet}, CBSD68 \cite{martin2001database}, and $256\times 256$ center-cropped images from Urban100 \cite{huang2015single} and DIV2K \cite{agustsson2017ntire} are employed. For RGB images, we compressively sample each R, G, and B channel separately, and multiply the input and output channel numbers of the first and last convolutions in U-Net and injectors by 3. Peak signal-to-noise ratio (PSNR) and structural similarity index measure (SSIM) \cite{wang2004image} are employed as metrics. Results of existing methods are sourced from original publications, or obtained via careful hyperparameter tuning and re-training. In cases where replicable algorithms are not available, we indicate this in our tables using a ``-" symbol. Especially, for SR3, we adopt $\A^\dagger \y$ as its conditional input.

\textbf{Results.} Tab.~\ref{tab:compare_sota_e2e} shows that our IDM outperforms other end-to-end learned CS NNs, notably exceeding PRL-PGD$^+$ by average PSNR margins of 0.96dB, 1.22dB, 2.14dB, and 0.48dB on the four sets. Considering the test sample numbers, our IDM's PSNR improvements on CBSD68 and Urban100 are generally statistically significant, while the gains on Set11 and DIV2K are less statistically significant. As depicted in Fig.~\ref{fig:comp_cbsd68}, IDM recovers accurate and high-fidelity details, particularly on building edges and windows. In contrast, four other competing CS NNs tend to produce oversmoothed and blurry building textures. IDM effectively reduces artifacts and achieves the highest PSNR, surpassing the second-best end-to-end image CS approach PRL-PGD$^+$ by 2.91dB. The superiority of IDM is further visually shown in Fig.~\ref{fig:comp_set11}. Competing methods can either oversmooth details or produce incorrect patterns with noticeable artifacts. In contrast, IDM achieves accurate and vivid details on the garment, and sharp, artifact-free strips around parrot's eye.

\begin{figure}[!t]
\setlength{\tabcolsep}{0.5pt}
\centering
\resizebox{\linewidth}{!}{
\scriptsize
\begin{tabular}{cccc>{\columncolor[HTML]{FFEEED}}c}
Ground Truth & $\A^{\dagger}\y$ & (1) DDNM* & (5) w/o Inj. & \textbf{(9) Ours}\\
\includegraphics[width=0.09\textwidth]{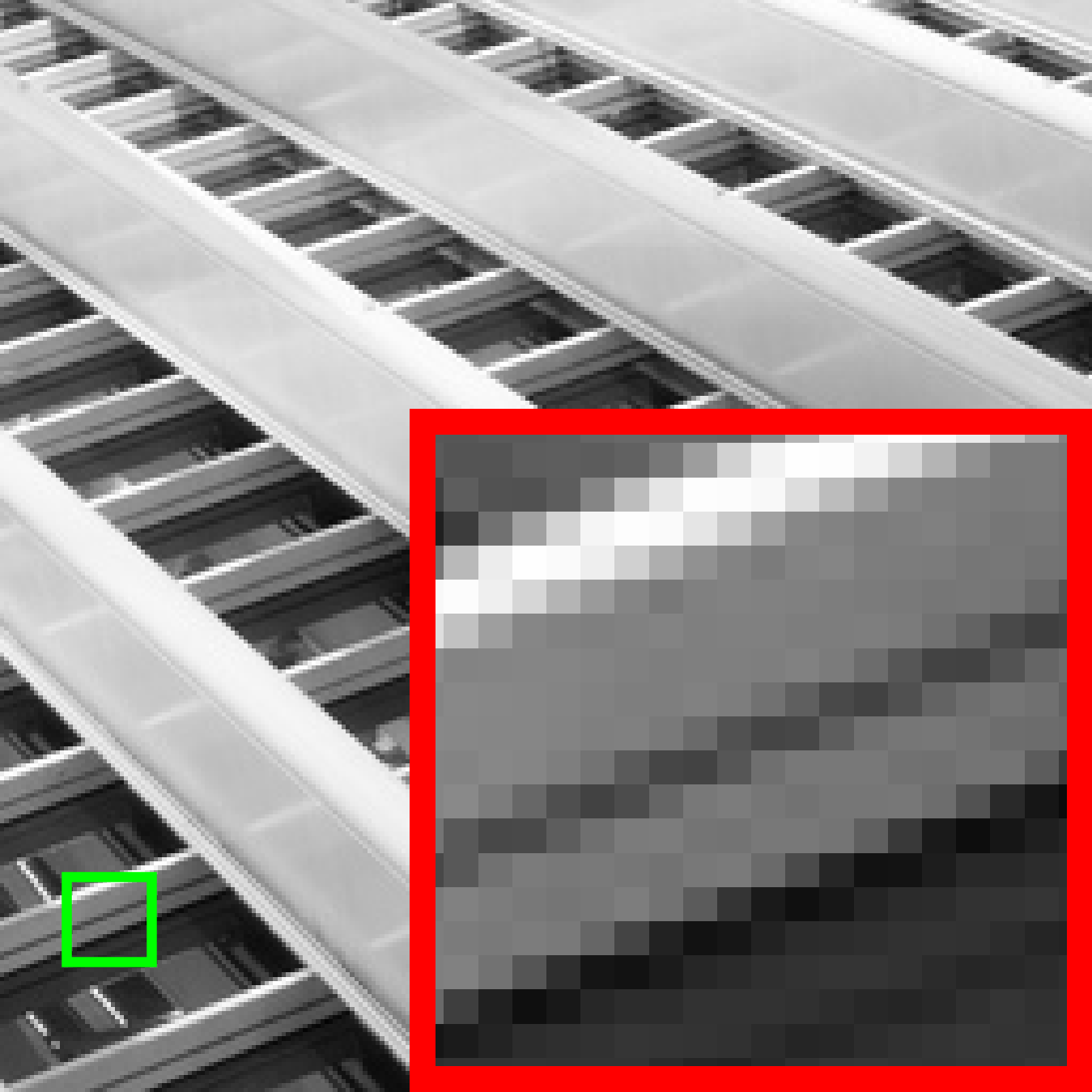}
&\includegraphics[width=0.09\textwidth]{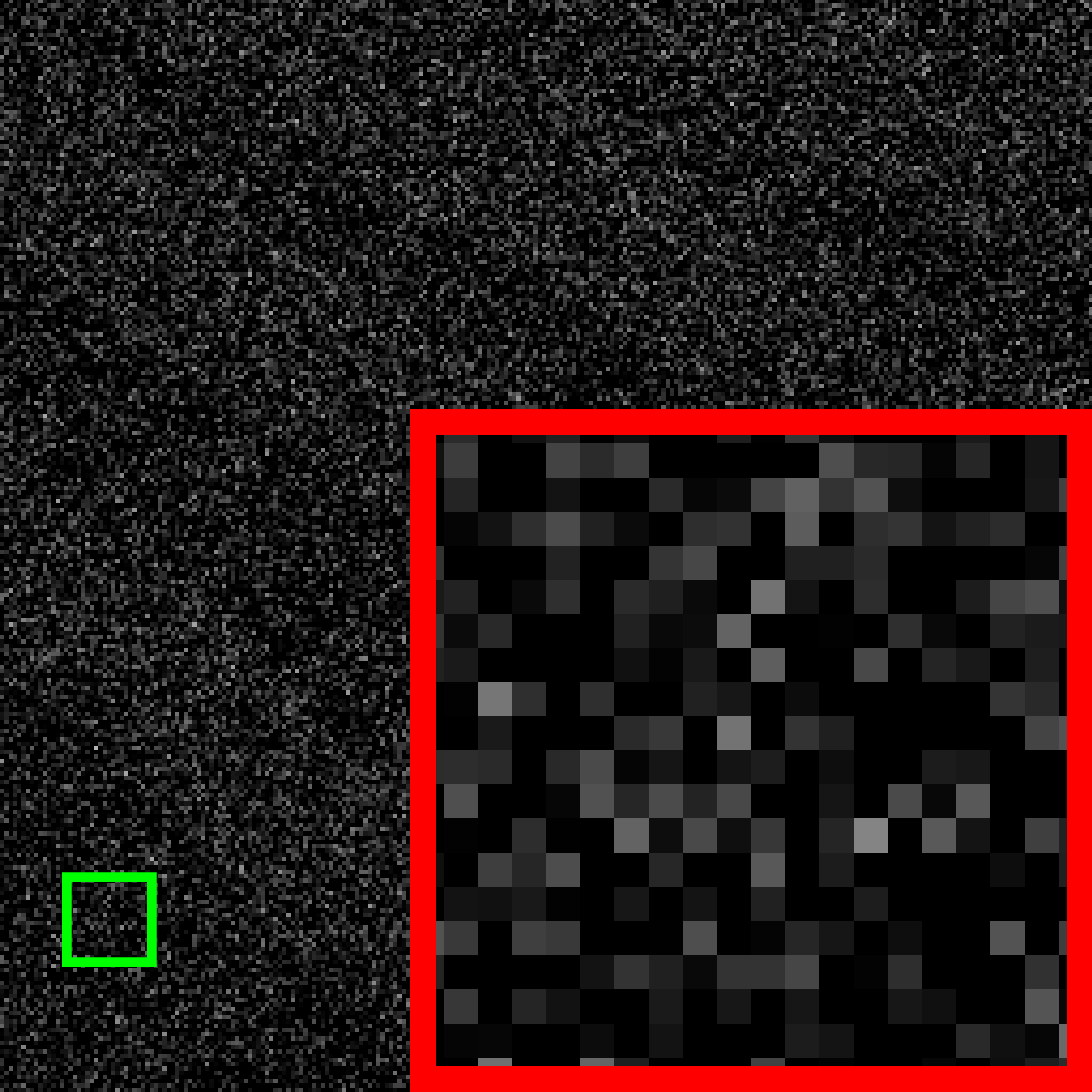}
&\includegraphics[width=0.09\textwidth]{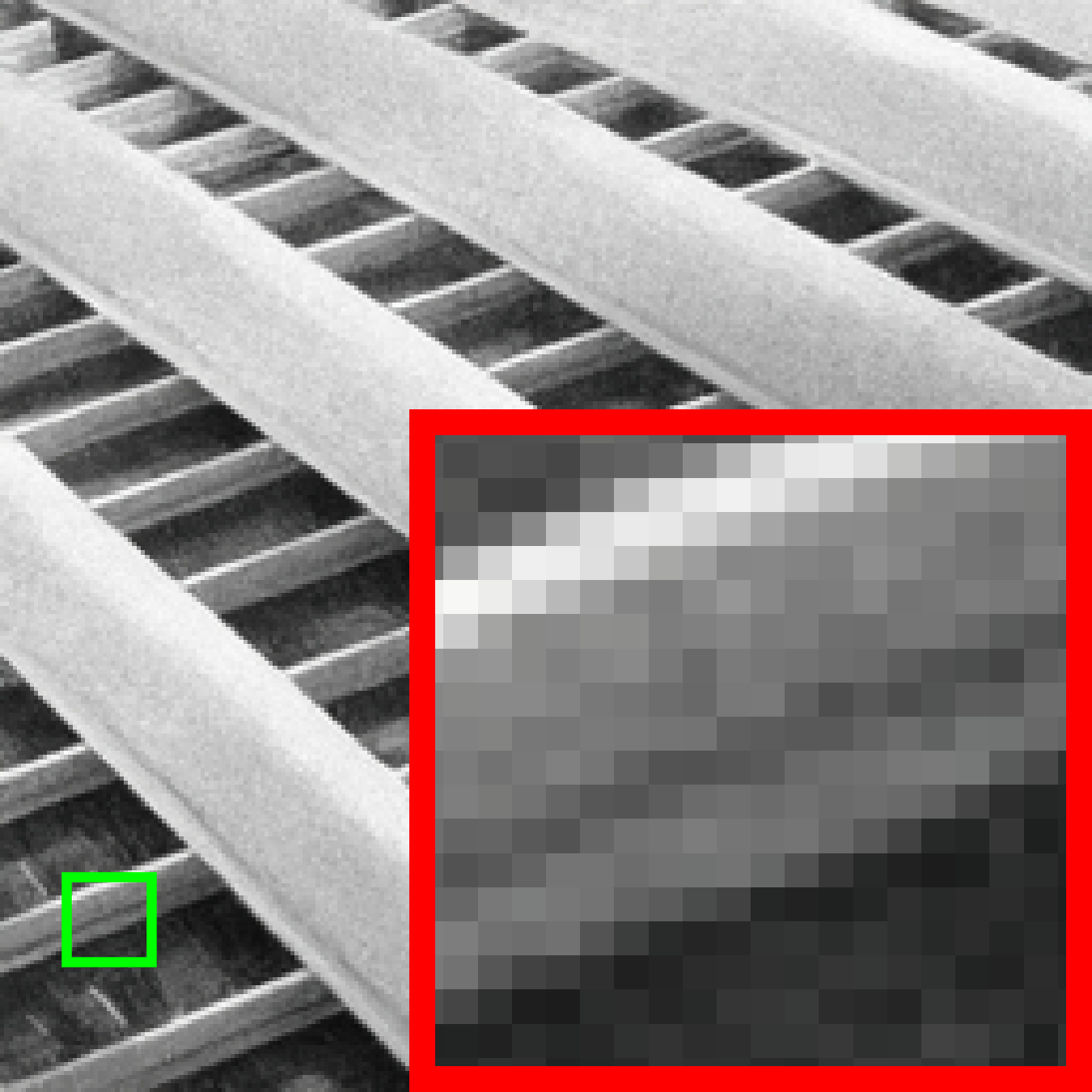}
&\includegraphics[width=0.09\textwidth]{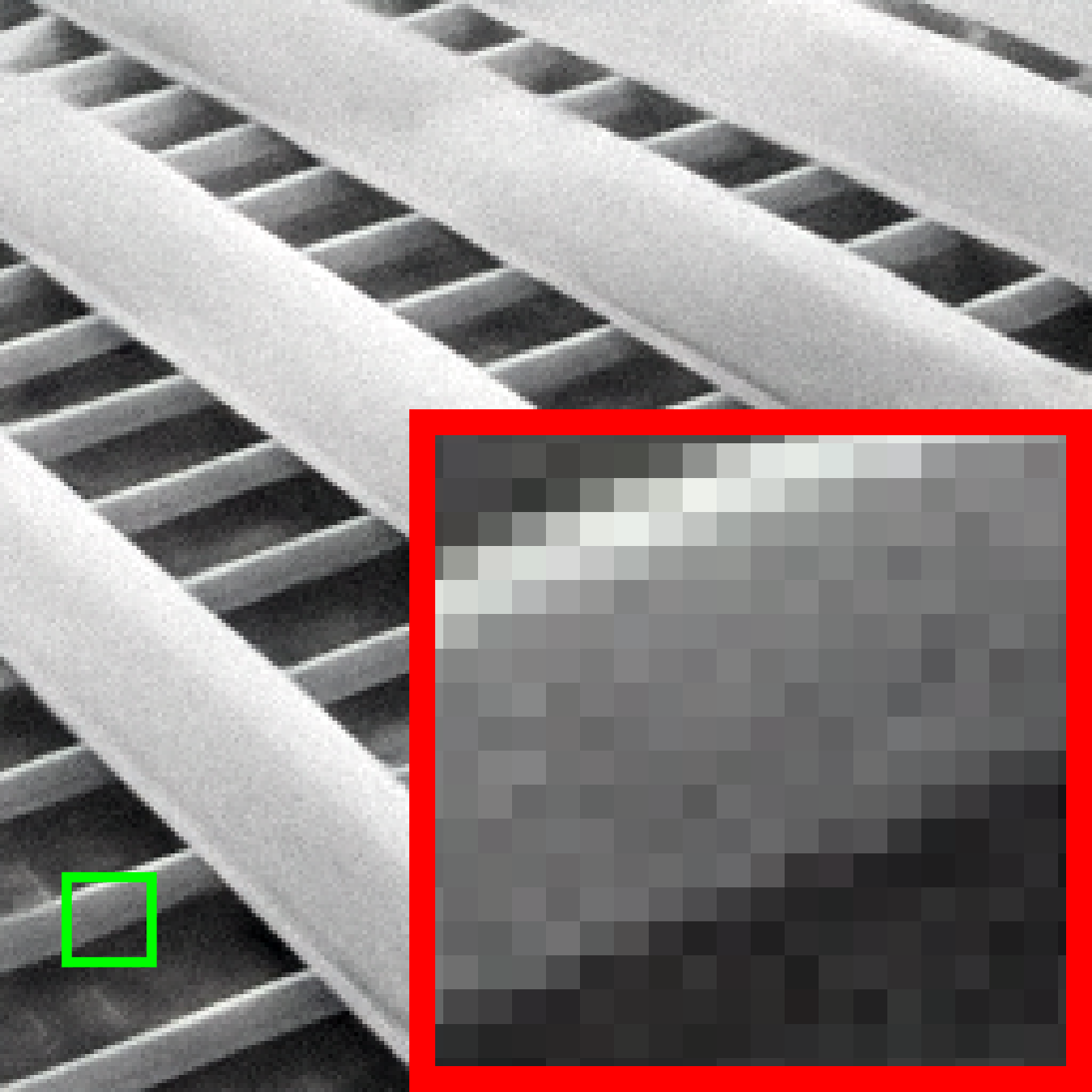}
&\includegraphics[width=0.09\textwidth]{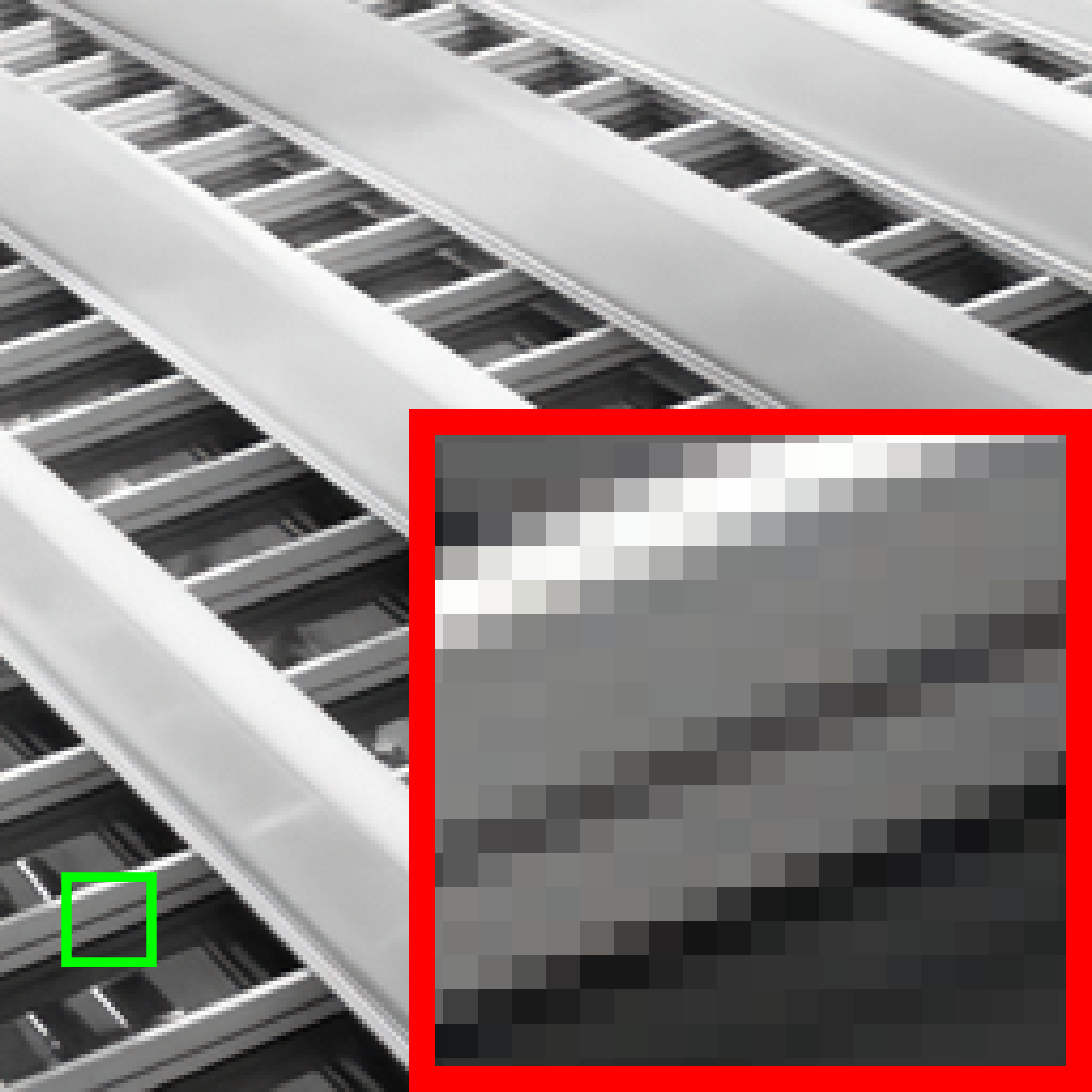}\\
PSNR/SSIM & 4.87/0.0293 & \secondbest{26.21}/\secondbest{0.7802} & 24.20/0.7446 & \best{34.18}/\best{0.9550}\\
\includegraphics[width=0.09\textwidth]{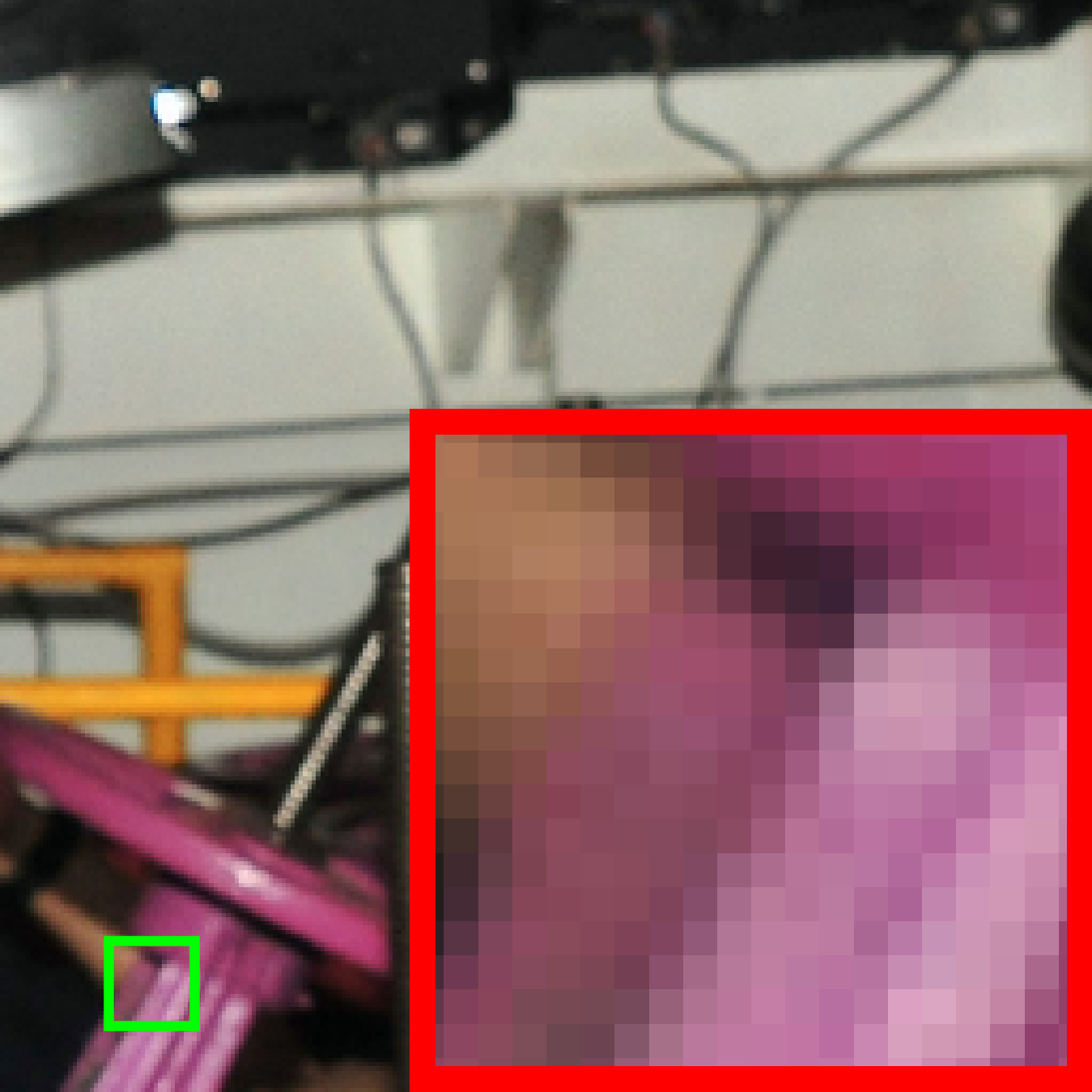}
&\includegraphics[width=0.09\textwidth]{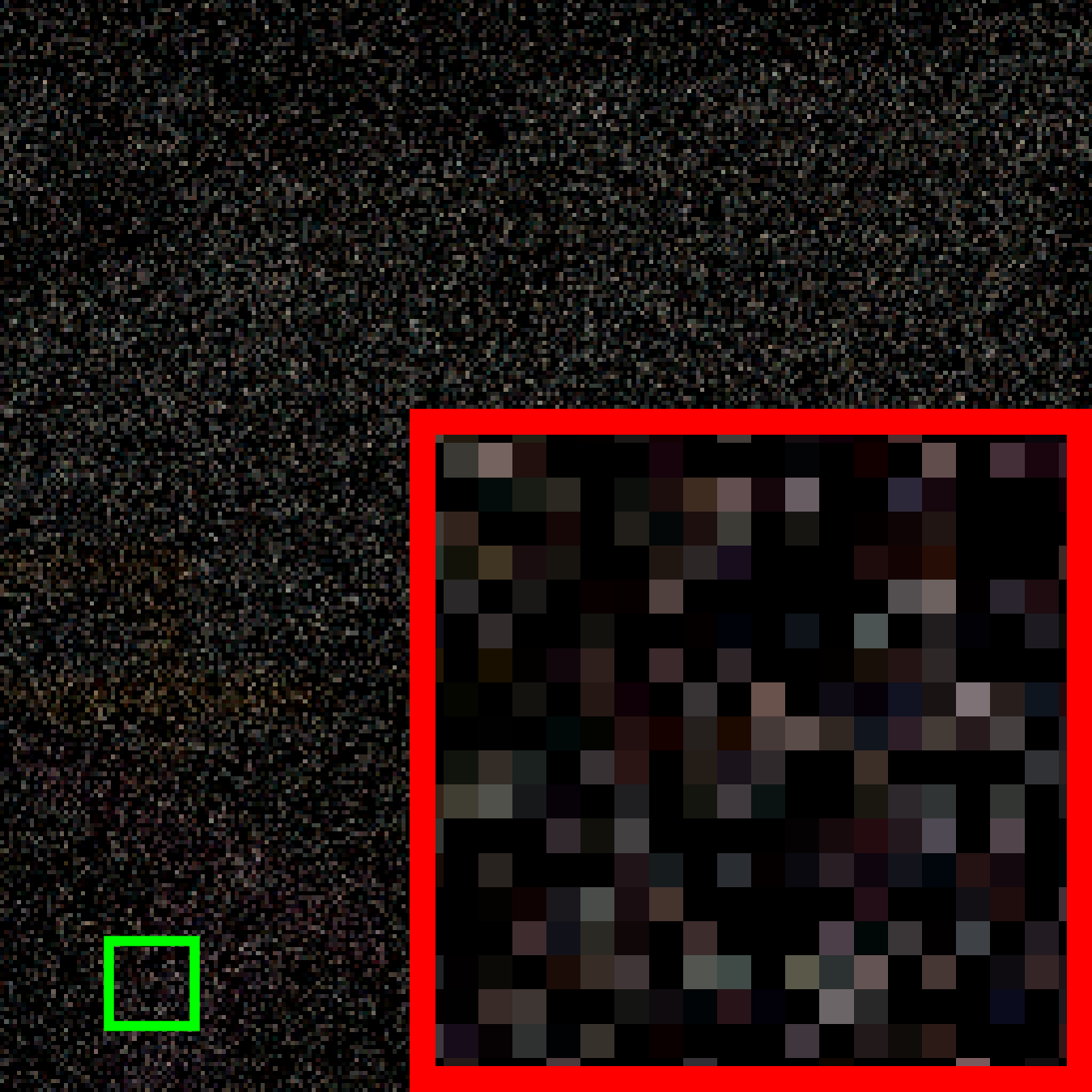}
&\includegraphics[width=0.09\textwidth]{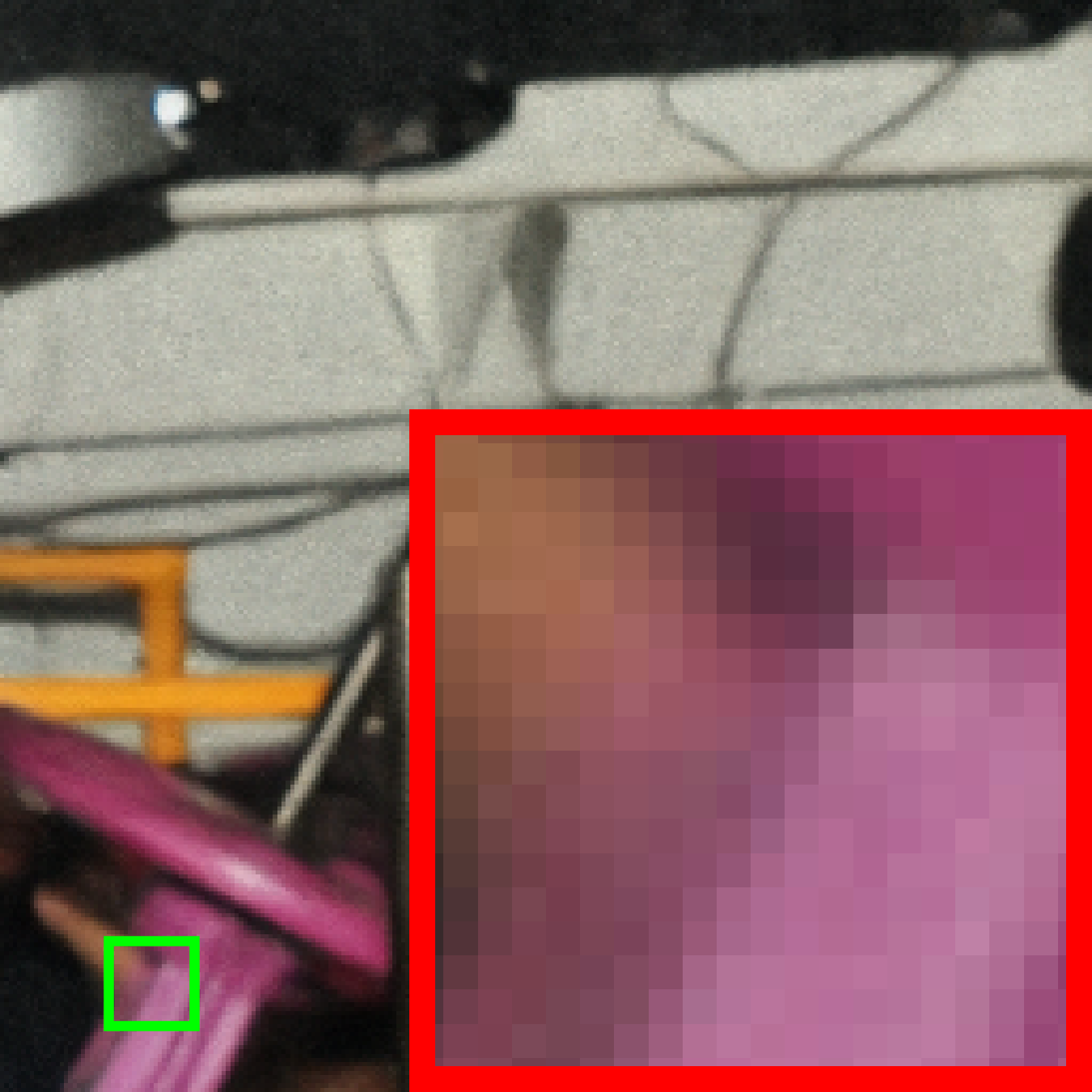}
&\includegraphics[width=0.09\textwidth]{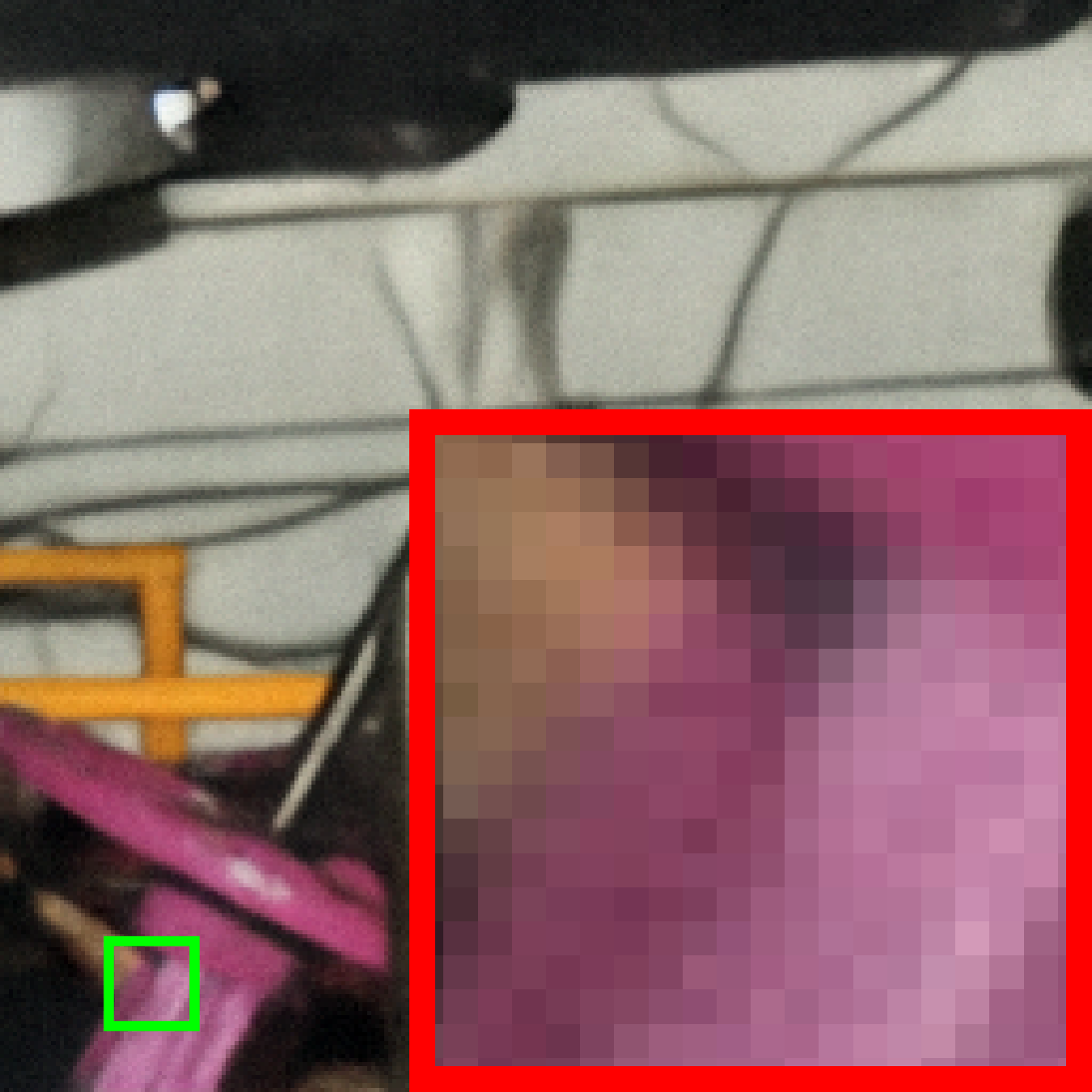}
&\includegraphics[width=0.09\textwidth]{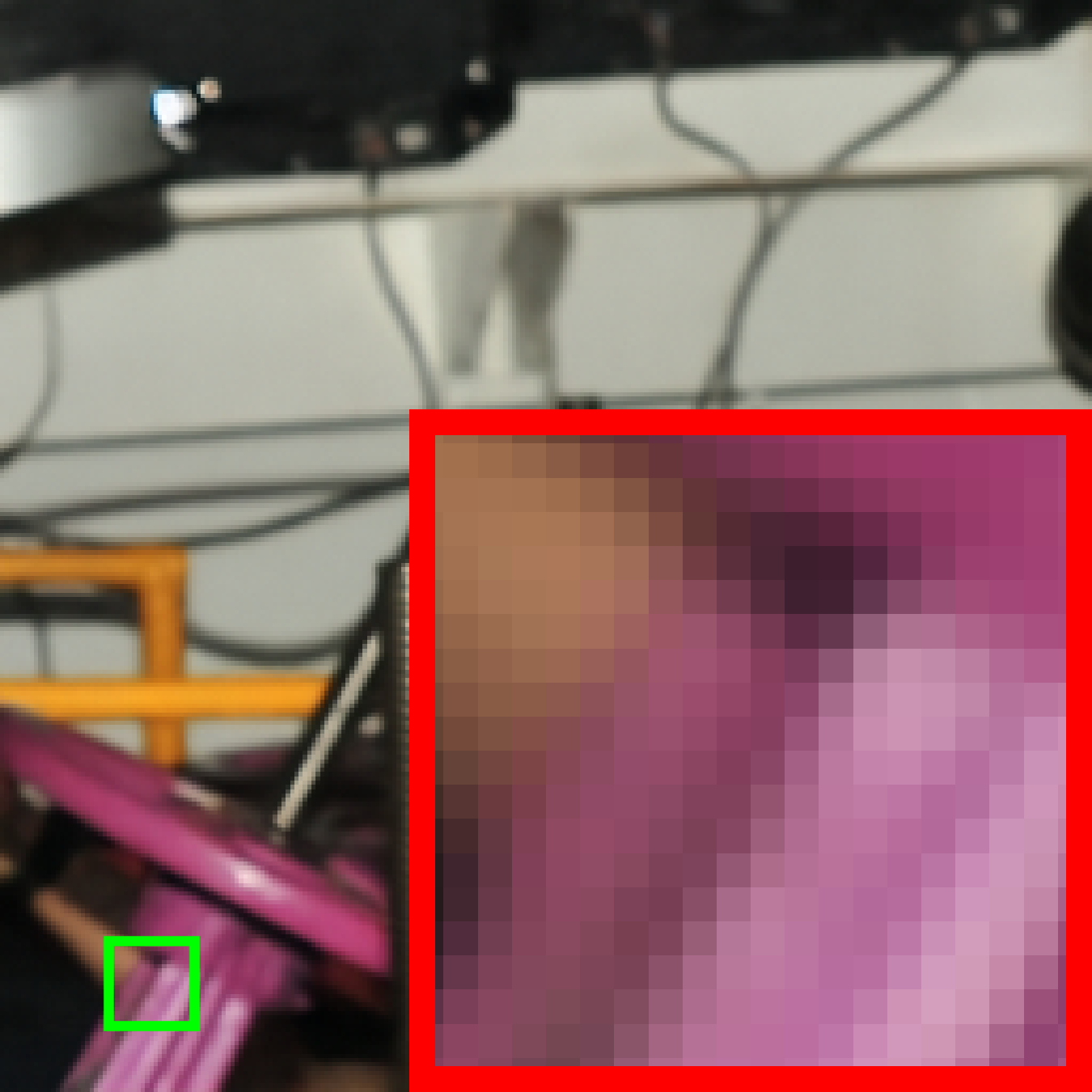}\\
PSNR/SSIM & 7.02/0.0446 & \secondbest{28.42}/0.8201 & 27.72/\secondbest{0.8315} & \best{37.20}/\best{0.9693}
\end{tabular}}
\vspace{-10pt}
\caption{\textbf{Comparison of CS recovery results from the method variants of (1), (5), and (9) in Tab.~\ref{tab:abla}} on the two images named ``img\_042" and ``0888" from Urban100 \textcolor{blue}{\textbf{(top)}} and DIV2K \textcolor{blue}{\textbf{(bottom)}} at CS ratio $\gamma =10\%$.}
\label{fig:comp_abla_CS}
\end{figure}

Tab.~\ref{tab:compare_sota_diff_quality} shows the advantage of IDM in PSNR and SSIM, outperforming the second-best contenders by a notable average margin of 7.27dB and 0.1652. Such an improvement is statistically significant and can be attributed to the ability of our method to address the shortcomings of generative priors in maintaining data fidelity and the insufficient adaptation to CS tasks of their noise estimation networks. Tab.~\ref{tab:compare_sota_diff_fid_lpips} compares the perceptual quality of images from diffusion-based approaches, using Fréchet inception distance (FID) \cite{heusel2017gans} and the learned perceptual image patch similarity (LPIPS) \cite{zhang2018unreasonable}. Our method achieves significant superiority, with 15-127 decreases in FID and 0.04-0.22 reductions in LPIPS. As depicted in Fig.~\ref{fig:comp_diff_urban100}, while DDNM and GDP can synthesize high-quality patterns in the areas of building windows, they also generate noise-like artifacts in the regions of sky and windows. Although $\Pi$GDM, DPS, PSLR, and SR3 reconstruct the basic shapes and structures of the original image, they struggle to recover detailed features of the building facade, which can be observed in the zoomed-in area. As Fig.~\ref{fig:comp_diff_urban100_and_div2k} shows, $\Pi$GDM, DPS, PSLD, and SR3 often result in oversmoothed and blurry outputs. DDNM and GDP can introduce noise-like artifacts despite capturing basic structures. Our IDM reliably produces high-quality and precise reconstructions of the intricate patterns of lines. Notably, SR3 requires 240 hours and 75GB of memory per GPU for training. Our IDM stands out by necessitating only 76 hours and 15GB memory per GPU, delivering high-quality, artifact-free images using merely 3 NFEs for the noise estimator---a stark contrast to the $\ge$100 NFEs required by competing methods. This results in an inference speed-up of approximately 15-370 times, as reported in Tab.~\ref{tab:compare_sota_diff_complexity}. These observations validate the effectiveness of our end-to-end fine-tuning framework for diffusion sampling learning, and wiring technique for two-level invertibility and reuse.

\subsection{Ablation Study and Analysis}
\label{sec:ablation_and_analysis}
\begin{figure}[!t]
\centering
\includegraphics[width=\linewidth]{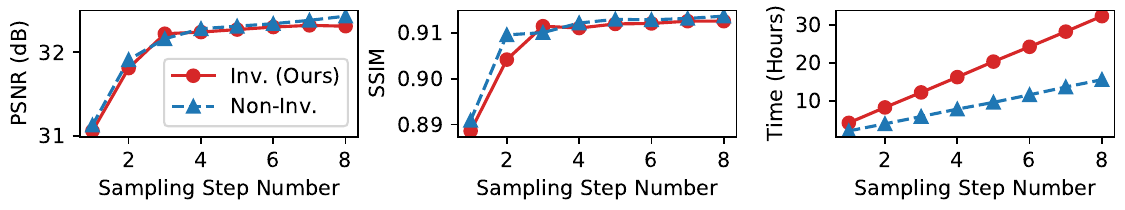}
\vspace{-25pt}
\caption{\textbf{Comparison of different IDM variants with $1\le T\le 8$.} We evaluate PSNR ($\uparrow$) and SSIM ($\uparrow$) for the luminance component image CS on Set11 at $\gamma =10\%$ \textbf{\textcolor{blue}{(left \& middle)}} and training time cost ($\downarrow$) \textbf{\textcolor{blue}{(right)}}. Parameter number of model is calculated as $(115.254 + 0.006T)$M.}
\label{fig:abla_psnr_ssim_time_T}
\end{figure}

\begin{figure}[!t]
\centering
\includegraphics[width=\linewidth]{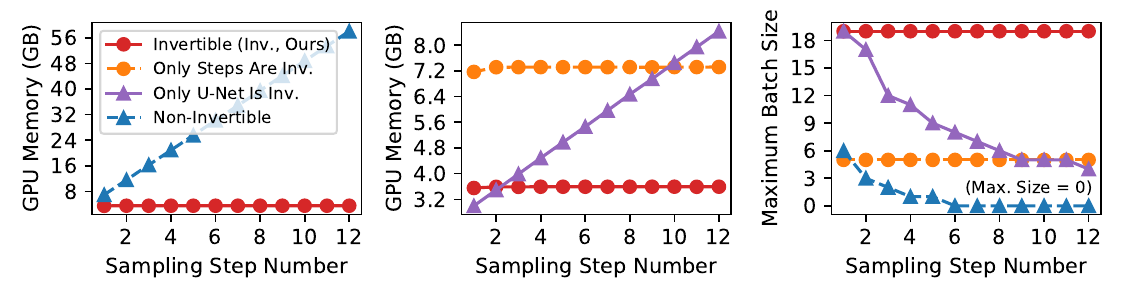}
\vspace{-25pt}
\caption{\textbf{Comparison of different IDM variants with $1\le T\le 12$.} We evaluate memory use ($\downarrow$) on A100 (80GB) GPU \textbf{\textcolor{blue}{(left)}}, and maximum batch size ($\uparrow$) for training on 1080Ti (11GB) GPU \textbf{\textcolor{blue}{(right)}}. Notably, our IDM reduces memory requirement by 93.8\% ($\approx$15/16) at $T=12$.}
\label{fig:abla_mem_max_bsz_T_zoomed}
\end{figure}

\textbf{Setup.} We evaluate scaled-down IDM variants trained on an NVIDIA A100 GPU with $T=2$, a batch size of 4, and a patch size of 128 in this section. Results are in Tab.~\ref{tab:abla} and Figs.~\ref{fig:comp_abla_CS}-\ref{fig:abla_mem_max_bsz_T_zoomed}. Two SOTAs: DDNM \cite{wang2023zero} and PRL-PGD$^+$ \cite{chen2023deep} re-trained for RGB image CS, are included as references in (10) and (12). We also train two measurement-conditioned SD-based DDNM variants in (1) and (2) \cite{saharia2023image,rombach2022high}.

\textbf{Effect of End-to-End DDNM Sampling Learning.} Comparisons (1) \vs (3), (2) \vs (5), and (10) \vs (11) in Tab.~\ref{tab:abla} exhibit that our end-to-end DDNM sampling learning provides substantial PSNR improvements of 2.06-4.34dB. It also decreases the number of required steps by 98\% and boosts inference speed by 27.55-29.50 times. Notably, the method variant (11) needs only 4.2 hours of fine-tuning, validating the efficiency and rapid deployment capability of our IDM.

\textbf{Effect of Two-Level Invertible Network Design, and Reuse of Pretrained Model Weights.} Comparisons of (3) \vs (4) and (5) \vs (6) in Tab.~\ref{tab:abla} show that our wired invertible models achieve a 61.3\% reduction in memory usage while effectively maintaining recovery performance. Fig.~\ref{fig:abla_mem_max_bsz_T_zoomed} demonstrates that our two-level wiring scheme consistently keeps memory usage within 3-4GB and scales the maximum batch size by a factor of 3, significantly improving overall training efficiency on modern GPUs \cite{mangalam2022reversible}.

As illustrated in Fig.~\ref{fig:abla_psnr_ssim_time_T} (right), our enabled invertibility (Inv.) can maintain competitive reconstruction quality but doubles training time due to recomputation during back-propagation. However, we mitigate this by reusing pre-trained models. Comparisons of (1) \vs (2), (3) \vs (5), and (4) \vs (6) in Tab.~\ref{tab:abla} reveal that weight reuse yields 0.17-1.42dB PSNR improvements and reduces training time by 90.3-97.0\%, enabling us to train an unpruned IDM within one day. These results underscore the critical importance of our ``wiring + reuse'' strategy for recovery. Furthermore, as evidenced by (6) \vs (11), reusing advanced models such as SD v1.5 further enhances CS performance.

In particular, we emphasize that scaling up our model back to the default settings of $T=3$, a batch size of 32, and a patch size of 256 makes fine-tuning a non-invertible IDM variant of (9) even impossible on 4 A100 (80GB) GPUs due to its excessive memory requirement. We also find that the maximum batch size on such a configuration is actually reduced from 184 (Inv.) to 20 (Non-Inv.), consuming a peak memory of 70.08GB per GPU and resulting in a decrease of 0.3-0.5dB in final PSNR. Notably, with our two-level invertible models, the peak memory use per GPU is suppressed to 14.64GB, making it manageable not only for A100 but also for other GPUs like 3090 and 4090 (24GB). This again verifies the effectiveness of our invertible design.

\textbf{Effect of Different Step Numbers $T$.} Fig. \ref{fig:abla_psnr_ssim_time_T} shows the relationship between $T$, reconstruction quality (in PSNR and SSIM), and computational cost (in time). As observed, PSNR and SSIM improve significantly when $T\leq 3$, with a rapid increase in reconstruction quality. However, when $T>3$, the improvements taper off, and the curves plateau, indicating diminishing returns in reconstruction performance. Meanwhile, the computational cost increases linearly with $T$, highlighting a trade-off between quality and efficiency. Based on these observations, we strategically select $T=3$ as the default value, as it achieves a near-optimal balance between image recovery quality and inference speed. This choice ensures the practicality of our method for real-world applications, where both accuracy and efficiency are critical.

\textbf{Effect of Physics Integration via Injectors.} Comparison (6) \vs (7) in Tab.~\ref{tab:abla} reveals that injectors improve PSNR by 2.10-3.01dB, requiring only an extra 1.9 hours of training and marginal $(0.005T)\%$ or $(0.006T)$M increase in parameters. Notably, our two-level wiring integrates these injectors seamlessly into the U-Net blocks, making them invertible with minimal extra memory usage. Fig.~\ref{fig:comp_abla_CS} demonstrates that our method (9) minimizes artifacts and decently recovers intricate textures of lines and corners, outperforming (5).

\begin{table*}[!t]
\centering
\caption{\textbf{Extension of our invertible design and injector modules to two end-to-end CS NNs: RK-CCSNet and CSNet+} on Set11 at $\gamma =50\%$.}
\vspace{-10pt}
\label{tab:extension_to_other_cs_networks}
\begin{tabular}{l|cccc}
\shline
\rowcolor[HTML]{EFEFEF} Method & PSNR (dB, $\uparrow$) & SSIM ($\uparrow$) & Memory (GB, $\downarrow$) & Parameter Number (M, $\downarrow$) \\ \hline \hline
RK-CCSNet (ECCV 2020) \cite{zheng2020sequential} & 38.03 & 0.9731 & \secondbest{1.3} & \best{0.631} \\
RK-CCSNet (w/ Invertible Design) & 38.07 & 0.9735 & \best{0.6} & \best{0.631} \\
RK-CCSNet (w/ Injectors) & \secondbest{38.91} & \secondbest{0.9762} & 1.5 & \secondbest{0.633} \\
RK-CCSNet (w/ Both Invertible Design and Injectors) & \best{38.95} & \best{0.9763} & \best{0.6} & \secondbest{0.633} \\ \shline
\end{tabular}
\\ ~\vspace{3pt} \\
\begin{tabular}{l|cccc}
\shline
\rowcolor[HTML]{EFEFEF} Method & PSNR (dB, $\uparrow$) & SSIM ($\uparrow$) & Memory (GB, $\downarrow$) & Parameter Number (M, $\downarrow$) \\ \hline \hline
CSNet$^+$ (TIP 2019) \cite{shi2019image} & 38.53 & 0.9750 & \secondbest{3.7} & \best{1.419} \\
CSNet$^+$ (w/ Invertible Design) & 38.55 & 0.9750 & \best{2.2} & \best{1.419} \\
CSNet$^+$ (w/ Injectors) & \secondbest{39.65} & \secondbest{0.9802} & 4.3 & \secondbest{1.421} \\
CSNet$^+$ (w/ Both Invertible Design and Injectors) & \best{39.77} & \best{0.9806} & \best{2.2} & \secondbest{1.421} \\ \shline
\end{tabular}
\end{table*}

\begin{table*}[!t]
\centering
\caption{\textbf{Average PSNR (dB, $\uparrow$) of IDM and state-of-the-art CS NN PRL-PGD$^+$ to different CS ratios} on Set11. The results correspond to Tab.~\ref{tab:compare_sota_e2e}.}
\vspace{-10pt}
\label{tab:extension_to_other_cs_ratios}
\begin{tabular}{l|ccc}
\shline
\rowcolor[HTML]{EFEFEF} Method & $\gamma =30\%$ (seen) & $\gamma =50\%$ (seen) & $\gamma =50\%$ (unseen) \\ \hline \hline
PRL-PGD$^+$ (IJCV 2023) \cite{chen2023deep} & \secondbest{37.89} (specific to 30\%) & \secondbest{41.78} (specific to 50\%) & \secondbest{28.85} (specific to 30\%, applied to unseen 50\%) \\
\rowcolor[HTML]{FFEEED} \textbf{IDM (Ours)} & \best{38.85} (specific to 30\%) & \best{42.48} (specific to 50\%) & \best{41.60} (specific to 30\%, applied to unseen 50\%) \\ \shline
\end{tabular}
\end{table*}

\textbf{Effect of $\xhat_T$ Initialization and Pruning Schemes.} Comparison of (8) \vs (9) in Tab.~\ref{tab:abla} confirms the superiority of our initialization $\xhat_T=\sqrt{\bar{\alpha}_T}\A^\dagger \y$ with a PSNR improvement of 1.74dB. Furthermore, we observe that the scheme does not significantly enhance the performance of DDNM in (1), (2), and (10), as there is no notable PSNR gain. Conversely, changing the initialization of (3) back to $\xhat_T\sim\mathcal{N}(\mathbf{0}_N,\mathbf{I}_N)$ results in reduced PSNR 25.01/27.16 with a drop of 1.05dB. This suggests that the initialization particularly benefits end-to-end IDM learning when using a very limited number of steps. Additionally, the comparison of (7) \vs (9) indicates that NN pruning \cite{kim2023bk} effectively complements our method by reducing 0.7 billion (B) NN parameters and 3.0GB storage size, with a minor 0.07dB decrease in PSNR.

\textbf{Summary.} As a progression (1)$\rightarrow$(3)$\rightarrow$(4)$\rightarrow$(6)$\rightarrow$(7)$\rightarrow$(9) shows, our method significantly enhances and extends previous diffusion-based image reconstruction paradigm for CS. Comparison (1) \vs (9) exhibits that our IDM delivers a 5.91dB PSNR gain, reduces training time by 98\%, accelerates inference by 110 times, and cuts storage size by 88\%. These benefits are achieved with an acceptable investment of 8.0 hours fine-tuning and 4.9GB memory use. Compared to the SOTA CS NN (12), our approach (9) offers advantages in accuracy, memory and time efficiencies, and model size.

\subsection{Extension to Other CS Networks}
To further validate the generalizability of our approach, we apply our (1) invertible NN design and (2) injectors to the two well-established models: RK-CCSNet \cite{zheng2020sequential} and CSNet$^+$ \cite{shi2019image}. Using the original training setting for each network and 400 images from BSDS500 \cite{arbelaez2010contour}, we evaluate their performance on Set11 at a sampling rate of 50\%. Quantitative results are summarized in Tab.~\ref{tab:extension_to_other_cs_networks}. For RK-CCSNet, our invertible design reduces training memory by 53.8\% (from 1.3GB to 0.6GB) without changing parameter number, while the injectors improve PSNR by 0.88dB and SSIM by 0.0031. Applying both two components jointly achieves the best results with a PSNR of 38.95dB and SSIM of 0.9763. Similarly, for CSNet$^+$, our invertible design effectively reduces memory usage by 40.5\%, and the injectors improve PSNR by 1.12dB and SSIM by 0.0052. Combining both yields a PSNR of 39.77dB and SSIM of 0.9806, demonstrating the effectiveness of our method across different CS NNs.

\subsection{Generalization to Different CS Ratios}
We evaluate the generalization ability of IDM to different CS ratios. While our current IDM implementation follows the common practice in the community of natural image CS—training a separate model for each specific task—we examine its performance when applied to CS ratios unseen during training. Specifically, we train an IDM on a 30\% CS ratio (seen) and evaluate its performance on a 50\% CS ratio (unseen). The results are compared against PRL-PGD$^+$ \cite{chen2023deep}, the previous best-performing end-to-end learned CS NN. For completeness, we also include the results of both methods trained and tested on the same seen ratios. As shown in Tab.~\ref{tab:extension_to_other_cs_ratios}, IDM achieves a PSNR of 41.60dB when a model trained at $\gamma=30\%$ is applied to $\gamma=50\%$. While this performance is slightly lower than that of IDM trained specifically for $\gamma=50\%$ (42.48dB), it is significantly higher than the performance of PRL-PGD$^+$ on the same unseen task (28.85dB), showing a remarkable 12.75dB improvement. This observation demonstrates that IDM retains a strong degree of generalization ability to unseen CS ratios, which is particularly advantageous for real-world scenarios where training for all possible ratios may not be feasible.

\section{Conclusion}
We propose Invertible Diffusion Model (IDM), a novel efficient, end-to-end diffusion-based image CS method, which converts a large-scale, pre-trained diffusion sampling process into a two-level invertible framework for end-to-end reconstruction learning. Our method provides three advantages. Firstly, it directly learns all diffusion model parameters utilizing the CS reconstruction objective, unlocking the full potential of diffusion networks in the recovery problem. Secondly, it improves the memory efficiency by making both (1) sampling steps and (2) noise estimation U-Net invertible. Thirdly, it reuses pre-trained diffusion models to minimize fine-tuning effort. Additionally, our introduced lightweight injectors further facilitate reconstruction performance. Experiments validate that IDM outperforms existing CS NNs and diffusion-based inverse problem solvers, achieving new state-of-the-art performance with only three sampling steps.

This work offers new possibilities for enhancing image CS and general image reconstruction using large diffusion models, particularly in GPU resource-limited scenarios. Future work includes extending IDM to real CS systems, for example, fluorescence microscopy \cite{lichtman2005fluorescence,liu2024deep} and interferometric imaging \cite{sun2021deep}, as well as exploring non-linear compression techniques like JPEG2000 and deep autoencoders.

\textbf{Limitations.} While we have not found explicit evidence of overlap between our adopted test sets (Set11, CBSD68, Urban100, DIV2K) and the LAION dataset used for training the SD v1.5 model, we can not definitively rule out the possibility of overlap due to the sheer scale of the LAION dataset. This remains a potential limitation of our experiments.

\bibliographystyle{IEEEtran}
\bibliography{ref}

\begin{IEEEbiography}[{\includegraphics[width=1in,height=1.25in,clip,keepaspectratio]{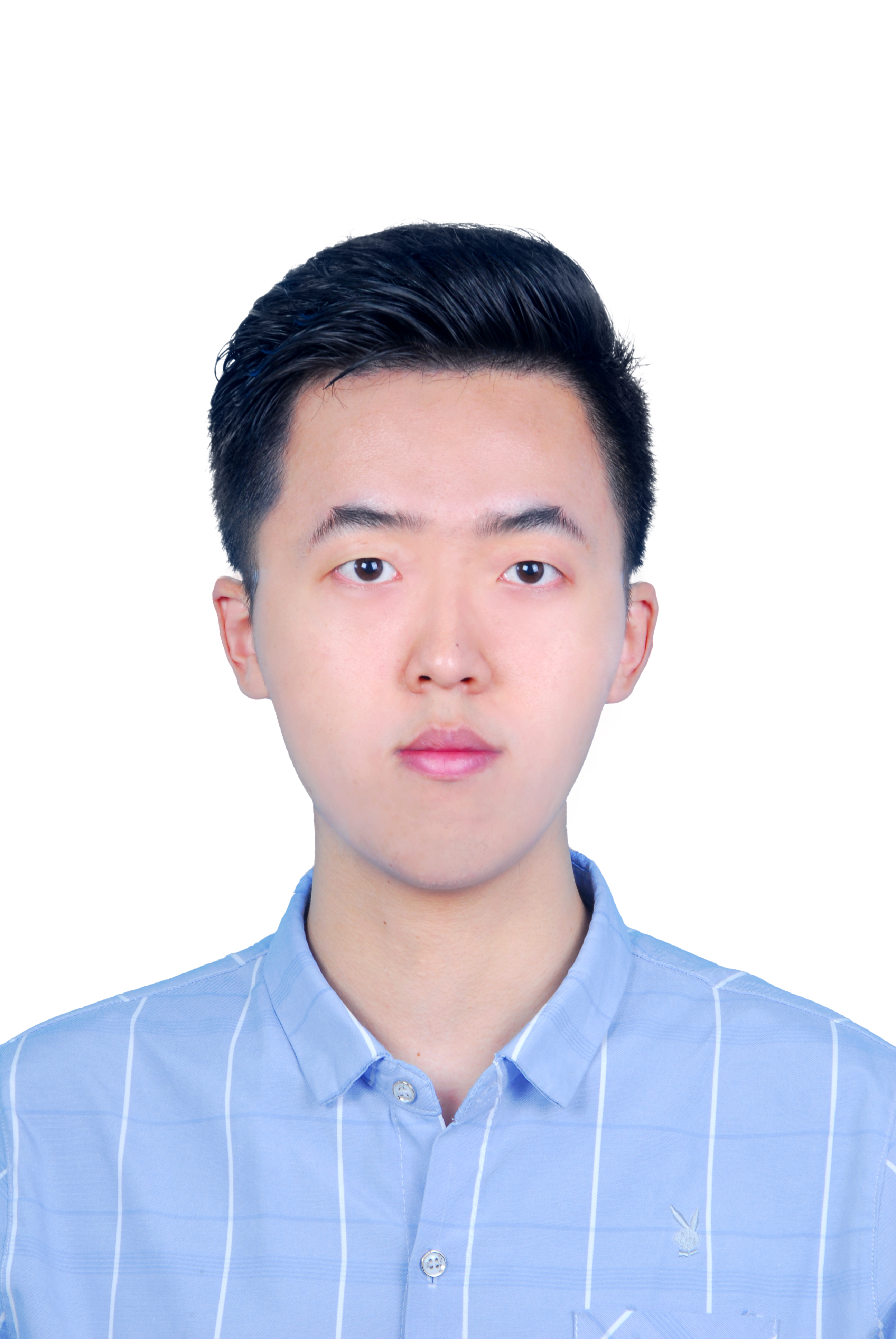}}]{Bin Chen} received the B.S. degree in the School of Computer Science, Beijing University of Posts and Telecommunications, Beijing, China, in 2021. He is currently working toward the Ph.D. degree in computer applications technology at the School of Electronic and Computer Engineering, Peking University, Shenzhen, China. His research interests include image compressive sensing and super-resolution. \end{IEEEbiography}

\begin{IEEEbiography}[{\includegraphics[width=1in,height=1.25in,clip,keepaspectratio]{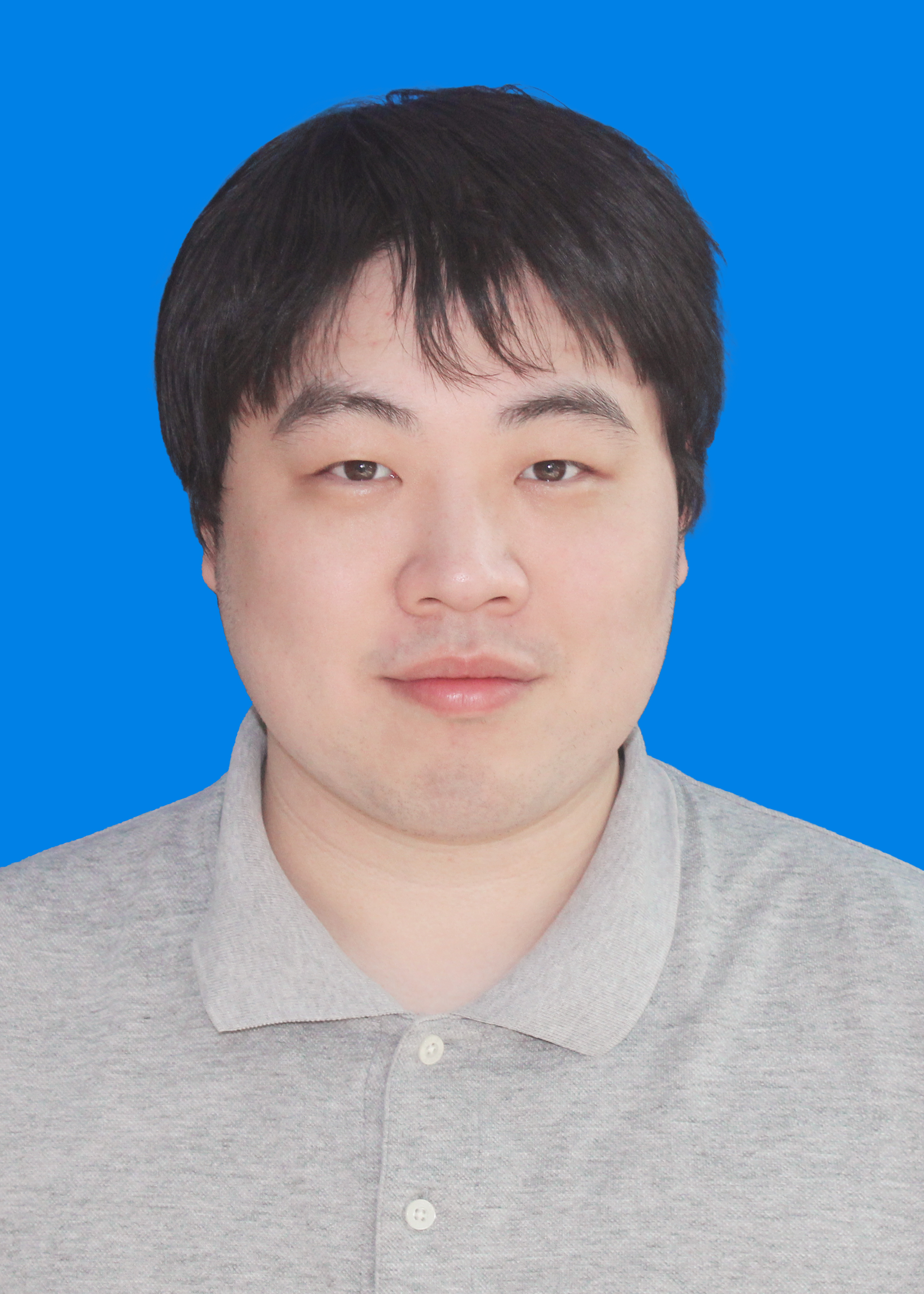}}]{Zhenyu Zhang} received the B.E. degree in software engineering from Beijing Jiaotong University, Beijing, China, in 2019. He is currently working toward the master’s degree in computer application technology with Peking University, Shenzhen, China. His research interests include image restoration, enhancement, and video processing.\end{IEEEbiography}

\begin{IEEEbiography}[{\includegraphics[width=1in,height=1.25in,clip,keepaspectratio]{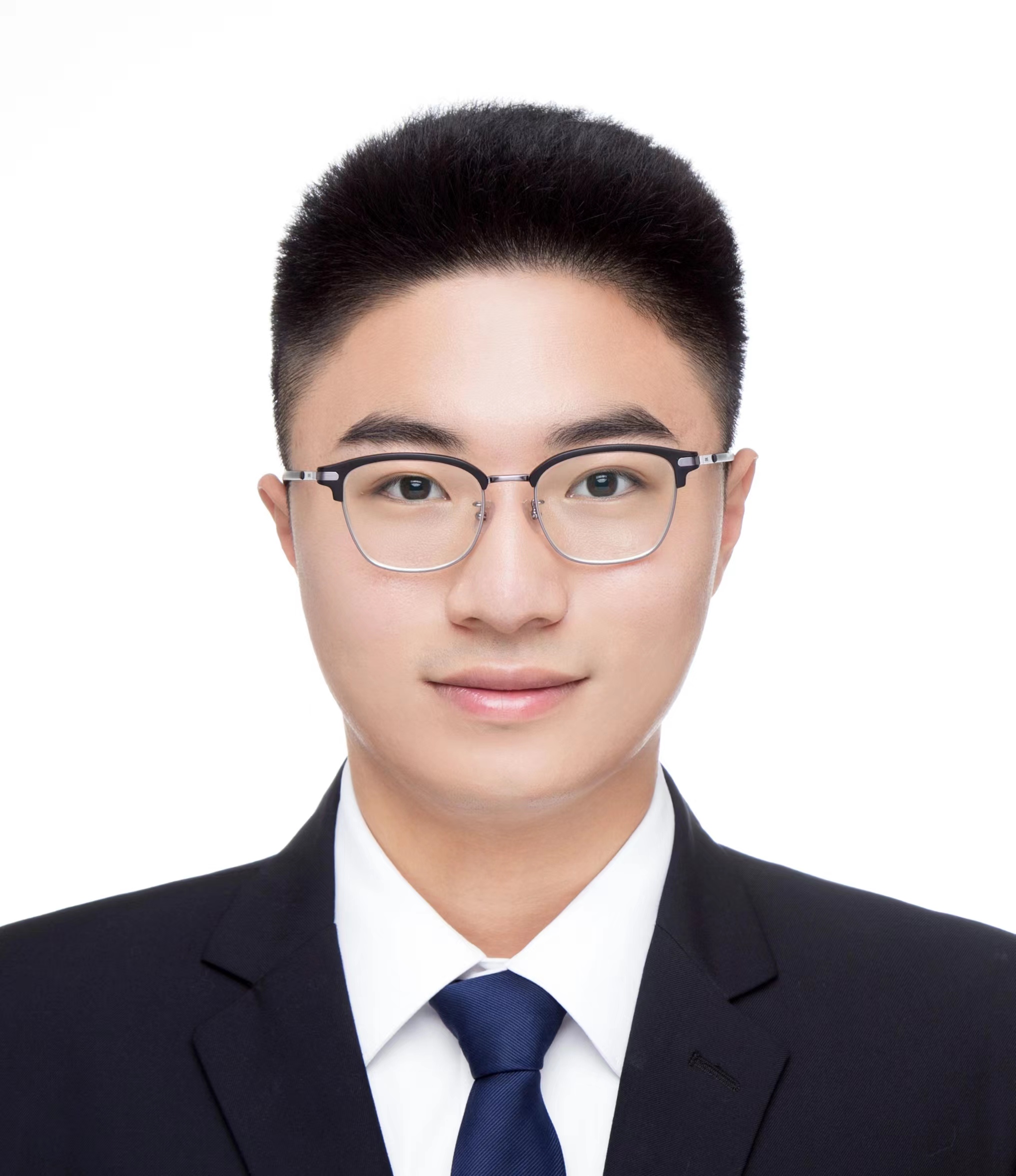}}]{Weiqi Li} received the B.E. degree from the school of Software Engineering, Tongji University, Shanghai, China, in 2022. He is currently working toward the master's degree in computer applications technology at Peking University, Shenzhen, China. His research interests include image super-resolution, compressive sensing, and computer vision.\end{IEEEbiography}

\begin{IEEEbiography}[{\includegraphics[width=1in,height=1.25in,clip,keepaspectratio]{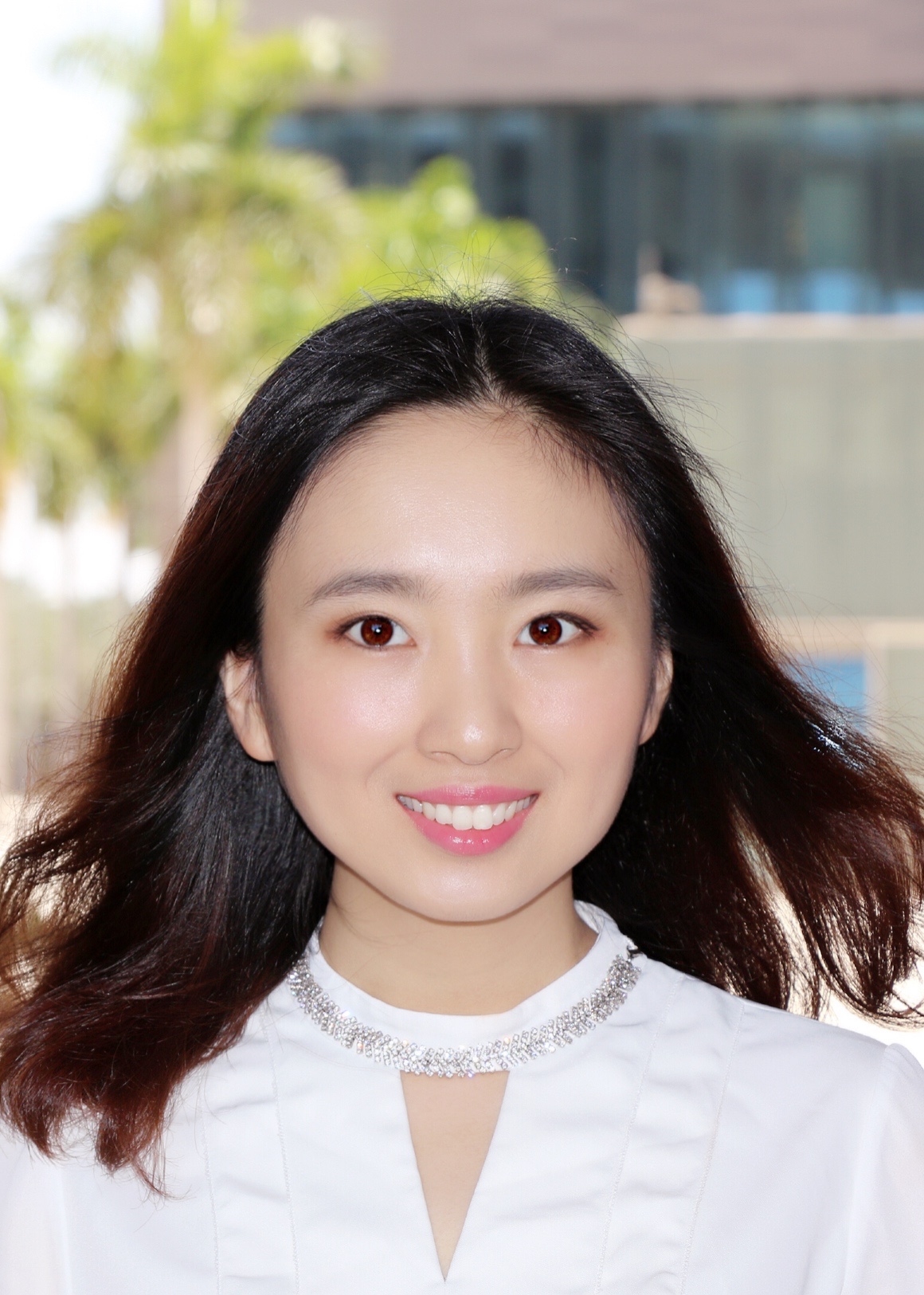}}]{Chen Zhao} is currently a Research Scientist at King Abdullah University of Science and Technology (KAUST), Saudi Arabia. She received her Ph.D. degree from Peking University (PKU), China in 2016, and studied in University of Washington (UW), US from 2012 to 2013. She did an internship at the National Institute of Informatics (NII), Japan and worked as a research assistant at Hong Kong University of Science and Technology (HKUST), China in 2016. Her research interests include image/video processing, image/video compression, image/video compressive sensing, and video understanding. On these topics, she has published over 50 papers in representative journals and conferences and has received over 3800 citations.   She has led the team to win the first‑place in 6 video understanding challenges in CVPR 2024, CVPR 2023 and ECCV 2022 respectively. She was the recipient of the Best Paper Award in CVPR workshop 2023, the Best Paper Nomination in CVPR 2022, and the Best Paper Award in NCMT 2015.
\end{IEEEbiography}

\begin{IEEEbiography}[{\includegraphics[width=1in,height=1.25in,clip,keepaspectratio]{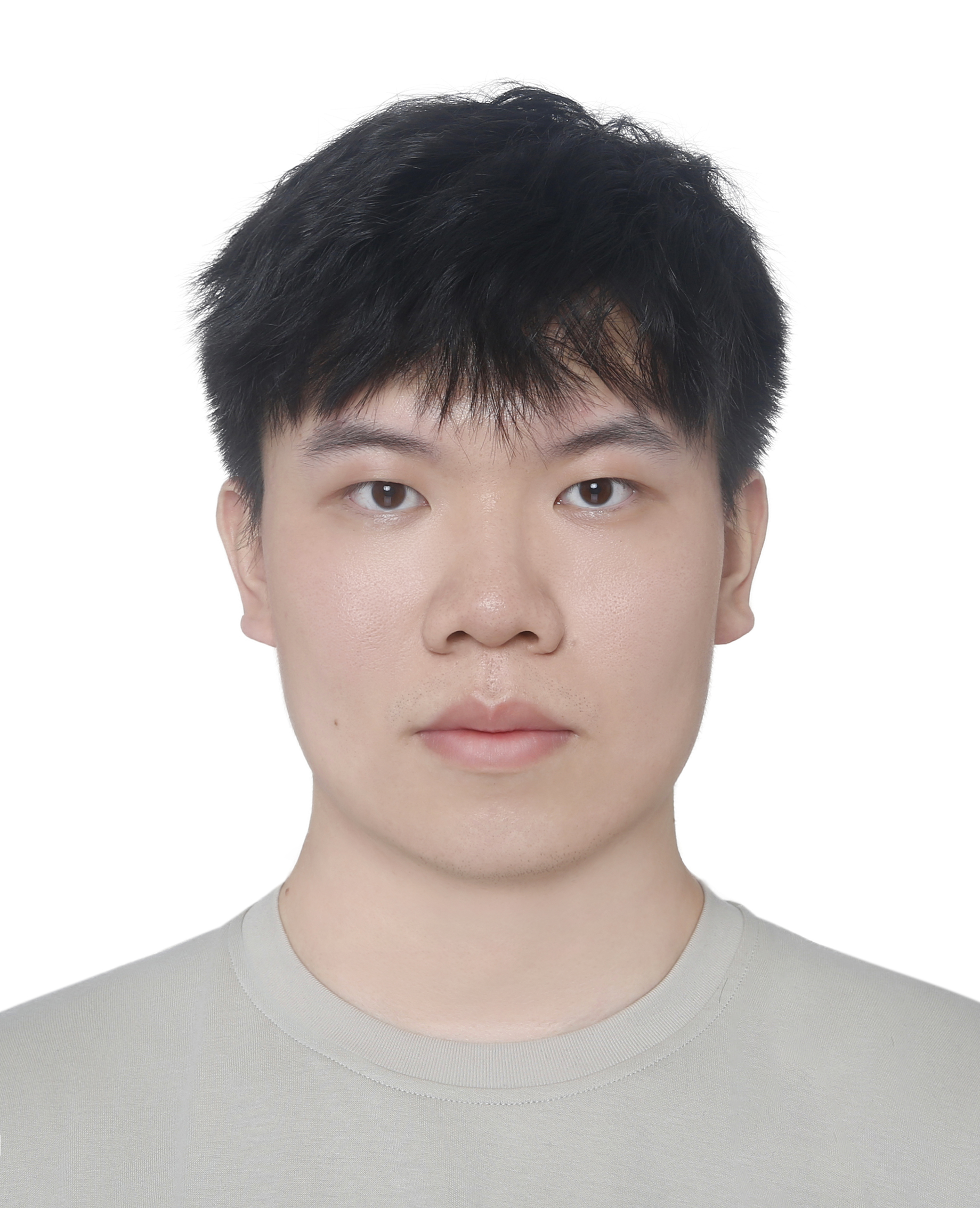}}]{Jiwen Yu} earned his bachelor's degree from the School of Computer Science at Northwestern Polytechnical University in Xi'an, China, in 2021. He is currently pursuing the master's degree in Computer Application Technology at the Peking University in Shenzhen, China. His research interests include image restoration, diffusion models, and computer vision.\end{IEEEbiography}

\begin{IEEEbiography}[{\includegraphics[width=1in,height=1.25in,clip,keepaspectratio]{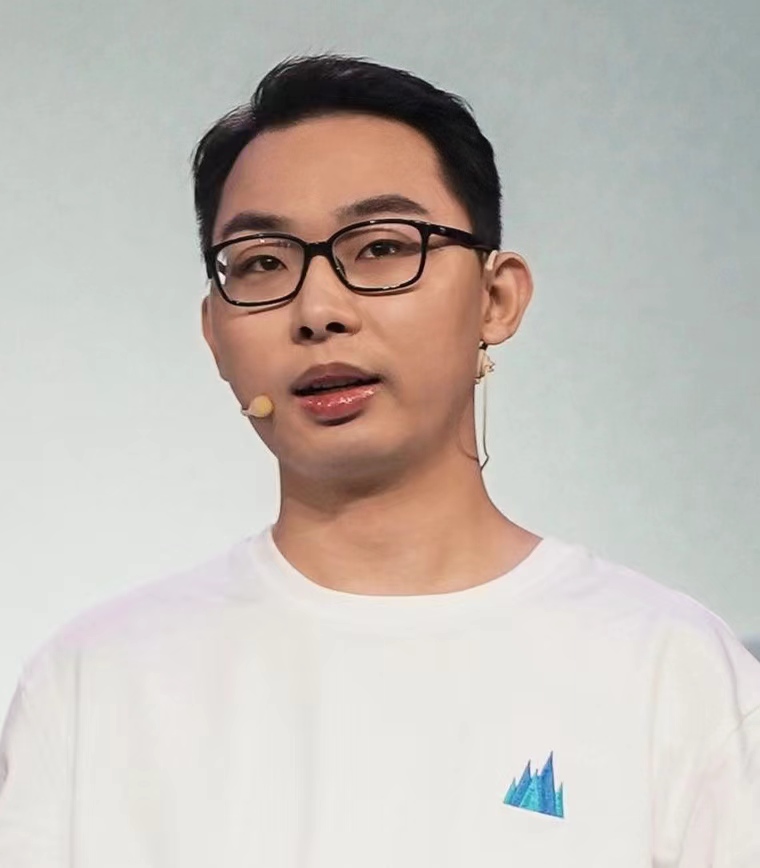}}]{Shijie Zhao} received his Bachelor's and Doctoral degrees from the School of Mathematics at Zhejiang University, in addition to a Master's degree from Imperial College London. Currently, he leads the Video Processing and Enhancement team at ByteDance's Multimedia Lab. His main areas of research include video enhancement, low-level vision, and video compression.
\end{IEEEbiography}

\begin{IEEEbiography}[{\includegraphics[width=1in,height=1.25in,clip,keepaspectratio]{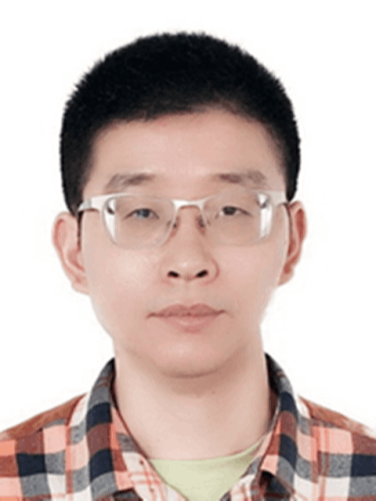}}]{Jie Chen} received the MSc. and Ph.D. degrees from the Harbin Institute of Technology, China, in 2002 and 2007, respectively. He joined as a Faculty Member with the Peking University, in 2019, where he is currently an Associate Professor with the School of Electronic and Computer Engineering. Since 2018, he has been working with the Peng Cheng Laboratory, China. From 2007 to 2018, he worked as a Senior Researcher with the Center for Machine Vision and Signal Analysis, University of Oulu, Finland. In 2012 and 2015, he visited the Computer Vision Laboratory, University of Maryland, and the School of Electrical and Computer Engineering, Duke University, respectively. His research interests include deep learning, computer vision, large language models, and Al4Science. He was the Co-Chair of International Workshops at ACCV, ACM MM, CVPR, ICCV, and ECCV. He was a Guest Editor of special issues Of IEEE Transactions on Pattern Analysis and Machine Intelligence, IJCV, and Neurocomputing. He has been selected as a Finalist of the 2022 Gordon Bell Special Prize for HPC-Based COVID-19 Research. He is an Associate Editor of the Visual Computer.
\end{IEEEbiography}

\begin{IEEEbiography}[{\includegraphics[width=1in,height=1.25in,clip,keepaspectratio]{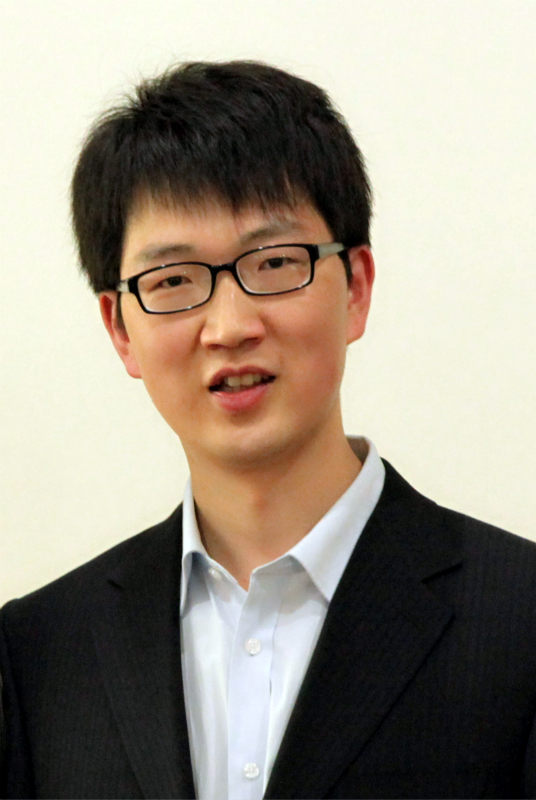}}]{Jian Zhang} (M'14) received Ph.D. degree from the School of Computer Science and Technology, Harbin Institute of Technology (HIT), Harbin, China, in 2014. He is currently an Associate Professor and heads the Visual-Information Intelligent Learning LAB (VILLA) at the School of Electronic and Computer Engineering, Peking University (PKU), Shenzhen, China. His research interest focuses on intelligent controllable image generation, encompassing three pivotal areas: efficient image reconstruction, controllable image generation, and precise image editing. He has published over 100 technical articles in refereed international journals and proceedings and has received over 10000 citations. He received several Best Paper Awards at international journals/conferences. He serves as an Associate Editor for the Journal of Visual Communication and Image Representation.
\end{IEEEbiography}

\end{document}